%% file: arxiv_main.tex
\documentclass{article} 
\usepackage{iclr2027_conference,times}

\input{math_commands.tex}

\usepackage[utf8]{inputenc} 
\usepackage[T1]{fontenc}    
\usepackage{url}            
\usepackage{booktabs}       
\usepackage{amsfonts}       
\usepackage{nicefrac}       
\usepackage{microtype}      
\usepackage{xcolor}         
\usepackage{amsmath}
\usepackage{graphicx}
\usepackage{placeins}
\usepackage{wrapfig}
\usepackage{relsize}
\usepackage{adjustbox}
\usepackage{float}
\usepackage{hyperref}
\hypersetup{
    colorlinks,
    linkcolor={red!50!black},
    citecolor=cyan,
    urlcolor={blue!80!black}
}

\usepackage{etoc} 
\etocdepthtag.toc{main} 

\newtoggle{showtb}
\toggletrue{showtb} 

\title{Relative Representations Enable Portable White-Box Monitoring Across Language \mbox{Models}}
\author{%
  Tim Bakker \\
  Qualcomm AI Research\thanks{Qualcomm AI Research is an initiative of Qualcomm Technologies, Inc.} \\
  \texttt{bakker@qti.qualcomm.com} \\
  \And
  Christos Louizos \\
  Qualcomm AI Research \\
}

\iclrfinalcopy 

\begin{document}

\maketitle

\ificlrfinal
 \lhead{Preprint.}
\else
 \lhead{Under review as a conference paper at ICLR 2027}
\fi
  
\input{content/main/abstract}
\input{content/paper}

\clearpage

\bibliographystyle{iclr2027_conference}
{ 
    \small
    \bibliography{iclr2027_conference.bib}
}


\clearpage
\appendix
\etocdepthtag.toc{appendix}  

\begingroup
  \etocsettocstyle{\section*{Appendix Contents}}{}
  \etocsettagdepth{main}{none}
  \etocsettagdepth{appendix}{subsubsection} 
  \tableofcontents
\endgroup
\clearpage
\input{content/appendix}


\end{document}

%% file: math_commands.tex
\usepackage{amsmath,amsfonts,bm}

\def\eqref#1{equation~\ref{#1}}

\def\1{\bm{1}}

\DeclareMathAlphabet{\mathsfit}{\encodingdefault}{\sfdefault}{m}{sl}
\SetMathAlphabet{\mathsfit}{bold}{\encodingdefault}{\sfdefault}{bx}{n}



%% file: content/main/abstract.tex
\begin{abstract}
Independently trained autoregressive language models often acquire similar
capabilities, but it remains unclear whether their internal representations
organize behaviorally relevant information in compatible ways. We show that models do in fact organize information similarly by using specific behavioral probes that operate on compatible representations. This unlocks portable white-box monitoring: a probe developed and validated on one
model can be reused when the model changes, without requiring labeled target
data. To overcome incompatible native activation coordinates, we express each
example through its angular distances to a shared set of anchors.
We introduce a \emph{global anchored} construction that summarizes information from all transformer
layers and removes model-specific layer selection, so the same frozen probe can be straightforwardly applied to another model.
We evaluate cross-model probe transfer across monitoring tasks on trivia datasets, safety monitoring, and
mathematical reasoning. On average, transferred global anchored probes preserve at least 85\% of same-model
probe AUROC over chance.
Anchored probes also support transfer to related
datasets, remain robust across model scale and prompt changes, and retain
performance better than standard probes after model updates.
Transfer is nevertheless task- and target-dependent and can fail even when the
monitored property remains strongly decodable in the target model. Overall,
global anchoring reveals substantial task-relevant compatibility across
independently trained models and provides a practical basis for portable
white-box monitoring.
\end{abstract}

%% file: content/paper.tex
\input{content/main/introduction}

\input{content/main/related_work}

\input{content/main/methodology}

\input{content/main/experiments}

\input{content/main/conclusion}

%% file: content/main/introduction.tex
\section{Introduction}

The Platonic Representation Hypothesis proposes that, as models become more capable, they increasingly converge on a common representational geometry: examples related by the same concepts have similar patterns of similarity and difference across models \citep{Huh2024-platonic}.
To compare how models organize examples, \citet{Moschella2022-es} introduce relative representations introduce relative representations and show that they preserve input data information across encoder models. We ask whether relative
representations also preserve information about model behavior: do
independently trained autoregressive language models represent the results of
their own computations in functionally compatible ways?

We test this question by asking whether a probe trained to predict a
behavioral property in one model remains predictive in another. Native
activations cannot support this test directly: models can differ in width and
depth, and even equal-width residual streams can use different bases. We therefore use
relative representations: each activation is represented by its angular distances to a 
shared anchor set, yielding one coordinate per anchor with the same anchor identities across models.

\begin{figure*}[t]
\centering
    \includegraphics[width=\textwidth]{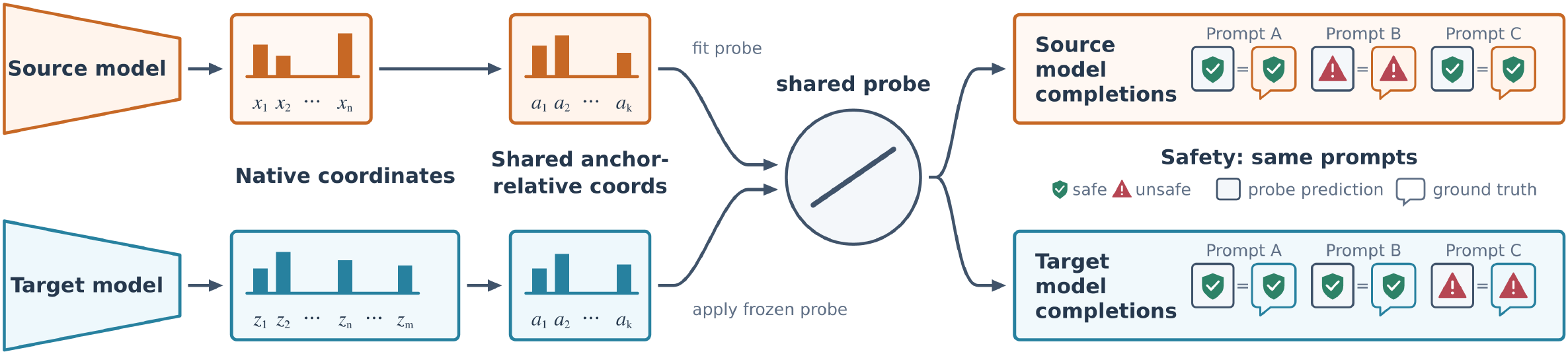}
\caption{\textbf{Relative probing tests behavioral representational
compatibility.} Native activation coordinates differ across independently
trained models, so a probe trained directly on one model's activations cannot
be directly reused. Relative representations instead express each monitored
example through its relations to the same indexed anchor examples. We train a
probe on these coordinates in a source model, freeze it, and test whether it
predicts the target model's behavior. Successful transfer is functional
evidence that the models share task-relevant relational organization.}
\label{fig:relative-probing-concept}
\end{figure*}

The simplest anchored construction measures each example's distances to the
anchors at one chosen layer and trains a \emph{local anchored probe} on the
resulting representation. Such a probe still requires compatible layer choices
across models, and the useful depth can vary by task and model. We instead introduce
\emph{global anchored representations}, which concatenate residual-stream
activations from all transformer layers at the monitored position before
computing angular distances to the shared anchor set. Because each model's
anchor activations are computed once and cached, each monitored example can be
compared with all anchors in one batched operation. The resulting
representation has one coordinate per anchor, so a source-trained probe can
consume it across models even when their widths and depths differ. Global
aggregation is not required for probe transfer, but it removes cross-model
layer selection and provides a depth-free default for comparing representations.

Successful cross-model probe transfer has a direct practical implication
for white-box monitoring or probing, which uses internal activations to
predict properties of model behavior. Prior work has used such signals to
anticipate answer quality \citep{Gottesman2024-estimating-knowledge,
Snyder2024-early-hallucinations}, detect hallucinations
\citep{Orgad2024-zp, Chen2024-inside}, and monitor safety-relevant behavior
\citep{McKenzie2026-su}. Activation probes are also actively deployed as
misuse mitigations in user-facing models \citep{Kramar2026-zv}. These monitors
are typically developed for a single model because their decision rules are
expressed in that model's activation coordinates. A new model may therefore
require collecting labeled activations, selecting a layer, and fitting and
validating another probe; model or deployment changes can also degrade an
existing monitor \citep{Kramar2026-zv,duan2026safetymonitorsstayreliable}.
Cross-model probe transfer would instead let a validated monitor be reused
without target-model labels or a learned target-specific adapter. This amortizes
probe development costs and, for instance, allows practitioners to develop probes
when access to models and data may be restricted.

We evaluate anchored probes on trivia datasets, safety monitoring, and
mathematical reasoning. Our main findings are:
\begin{itemize}
    \item When trained and evaluated on the same model, global anchored probes
    perform comparably to standard probes. \emph{Transferred} global anchored probes preserve at least 85\% of the
    same-model anchored-probe AUROC performance over chance.
    This provides evidence that models indeed share
    task-relevant relational structure at the level of their internal computations.
    \item The same approach supports joint transfer to a new model and a
    related dataset without using labeled examples from the new deployment
    setting. This allows a monitor developed with
    accessible models and data to be reused when both the deployed model and
    its data distribution change.
    \item We test portability across model scale, model updates, prompt
    variants, and external model families. Transfer can remain strong across
    large scale and prompt changes, and anchored probes are more robust to
    model updates than standard probes. Transfer is weak for some models with
    poor task performance and is task-dependent across external model
    families. We additionally provide evidence that transfer tolerates non-identical
    textual anchors and noisy anchor mappings.
\end{itemize}

%% file: content/main/related_work.tex
\section{Related Work}

Our central question is whether independently trained language models organize
information about their own behavior in compatible ways. We test this by
freezing a behavioral probe trained on one model and evaluating it on another
in shared anchor-relative coordinates. This connects work on representational
convergence with behavioral probing; 

\paragraph{Representational convergence and relative representations.}
The Platonic Representation Hypothesis proposes that increasingly capable
models converge in the relationships they encode among examples
\citep{Huh2024-platonic}. \citet{Moschella2022-es} provide the closest
methodological precursor: their relative representations encode examples by
similarity to fixed anchors, enabling zero-shot communication and transfer of
classification or reconstruction components across seeds and architectures.
We ask whether this relational compatibility extends to information about a model's own behavior.
Indra representations \citep{Lu2025-indra} use angular-distance profiles to
improve robustness of downstream classifiers and cross-modal retrieval. They
do not consider cross-model probe transfer and use full distance profiles
over a dataset rather than a fixed set of common anchors. Model stitching
\citep{Bansal2021-model-stitching} tests functional
compatibility by connecting the lower layers of one model to the upper layers
of another through a trained low-capacity adapter. Unlike our approach, it
learns a bridge between the two representation spaces.

\paragraph{Probing behavioral information.}
Probes have found decodable syntactic, semantic, spatial, temporal, and
behavioral information in model representations
\citep{Conneau2018-cram, Tenney2019-bert-pipeline,
Gurnee2023-space-time, Zou2023-rep-engineering}. Internal-state monitors use
such signals to anticipate answer quality or characterize model knowledge
\citep{Ferrando2024-do-i-know, Gottesman2024-estimating-knowledge,
Joshi2024-personas-truthfulness, Afzal2025-knowing-before-saying}, detect
hallucinations \citep{CH-Wang2023-electric-sheep, Chen2024-inside,
Orgad2024-zp, Simhi2024-combating-hallucinations,
Snyder2024-early-hallucinations}, and identify safety-relevant behavior
\citep{McKenzie2026-su, Kramar2026-zv}. These studies establish that a model's
activations contain information about its own behavior. Cross-model probe
transfer tests whether that information is organized compatibly across models.
The interpretation of probe performance also depends on distribution, probe
capacity, and layer choice \citep{Hewitt2019-control-tasks, Voita2020-mdl-probing,
Belinkov2022-probing-survey, Levinstein2023-no-lie-detector,
Mallen2023-gs, Kramar2026-zv, Min2026-corvus}. Cross-layer architectures
aggregate information from several layers within one model
\citep{Suresh2025-clap, Nordby2026-multilayer, Huang2026-deepguard}, while
\citet{Liang2025-neural-probe} use Bayesian optimization to select a single
probe insertion layer. Our global construction uses the whole depth of a model
to produce anchor-indexed coordinates, so we can test compatibility without
choosing corresponding layers across architectures.

\paragraph{Cross-model monitoring as an application.}
ACT-ViT is the closest prior system for cross-model white-box monitoring
\citep{Bar-Shalom2025-ig}. It pools activation tensors, maps each language
model through its own linear adapter, and applies a shared vision-transformer
backbone pretrained on labeled activations from multiple model--dataset pairs.
Learned source--target mappers similarly transfer activation-space
interventions by fitting a model-pair-specific bridge
\citep{oozeer2025activationspaceinterventionstransferred}. Both approaches
learn a new mapping for each target: ACT-ViT trains a new adapter for every
language model, while intervention transfer trains a bridge for every
source--target pair. We instead express each model relative to the same shared
examples, so transferring a monitoring probe to a new target requires only an
unlabeled pass over the anchor set and no learned target-specific component.

%% file: content/main/methodology.tex
\section{Methodology}
\label{sec:methodology}

Our method expresses each model's activations through their relations to shared
anchors, creating common coordinates in which a probe trained on one model can
be applied to another. This transfer supports portable monitoring and tests
predictive compatibility across models.

\subsection{Anchored Representations}

Let $h_{\ell}^{(m)}(x)$ denote the residual-stream activation of model $m$ at
layer $\ell$ for example $x$. Let $A_D=\{a_1,\dots,a_k\}$ be a set of $k$
anchor examples drawn from the training split of dataset $D$. Every model
processes the same indexed anchor set. Unless stated otherwise, anchor
activations are taken at the final prompt position, immediately before
generation. In an appendix sensitivity analysis, we instead average
activations over each anchor's generated completion
(Appendix~\ref{app:guard-linear-mean-token-auroc}). Following the relative
representation of \citet{Moschella2022-es}, a local anchored representation at
layer $\ell$ is
\[
r_{\ell}^{(m)}(x;A_D)
=
\big[
s(h_{\ell}^{(m)}(x),h_{\ell}^{(m)}(a_1)),
\dots,
s(h_{\ell}^{(m)}(x),h_{\ell}^{(m)}(a_k))
\big],
\]
where $s(\cdot,\cdot)$ is a relational distance function. The representation
has one coordinate for each anchor identity. Its dimension is determined by
the number of anchors and is independent of the model's hidden size.

A probe can be trained on $r_{\ell}^{(m)}(x;A_D)$ using the same supervision
as a probe on the native activation $h_{\ell}^{(m)}(x)$. We call the former an
\emph{anchored probe}. A \emph{standard probe} uses the native activation and
serves as the same-model baseline.

\subsection{Global Anchored Representations}

The direct local construction still requires a choice of probing layer. Our
main method instead summarizes the residual stream across the complete
network. For model $m$ with $L_m$ transformer layers, define
\[
g^{(m)}(x)
=
\operatorname{concat}\big(
h_{0}^{(m)}(x),\dots,h_{L_m-1}^{(m)}(x)
\big).
\]
We then compute the \emph{global anchored representation}
\[
R^{(m)}(x;A_D)
=
\big[
s(g^{(m)}(x),g^{(m)}(a_1)),
\dots,
s(g^{(m)}(x),g^{(m)}(a_k))
\big].
\]
The intermediate vector $g^{(m)}(x)$ can have a different dimension for every
model, but $R^{(m)}(x;A_D)$ always has dimension $k$ and the same anchor
indexing. A probe trained on one model can therefore be applied to another
model's representation even when the models differ in width and depth. In
contrast to a local anchored probe, the global construction also removes the
need to choose corresponding layers.
Figure~\ref{fig:anchored-probing-workflow} in the Appendix illustrates our construction.

Following the angular construction used for Indra representations
\citep{Lu2025-indra}, we use
\[
s(u,v)
=
\arccos\!\left(\frac{u\cdot v}{\lVert u\rVert\,\lVert v\rVert}\right).
\]
This is the angular distance in radians. For global representations it is
computed between whole-network vectors; for local representations it is
computed between activations from one layer.

\subsection{Task Settings}
\label{sec:method-task-settings}

We evaluate the probes in three data settings: Trivia settings test whether probes retain information about model behavior and answer correctness, Guard Distillation tests probe performance for safety monitoring, and Online MATH tests whether probes can predict correctness of reasoning models on math questions. Safety monitoring is a common production use-case for probes \citep{Kramar2026-zv}, and probes for early-correctness prediction during reasoning have been used for both efficiency \citep{zhang2025reasoningmodelsknowtheyre} and test time scaling \citep{ni2026reprobeefficienttesttimescaling}.

\paragraph{Trivia.}
We use TruthfulQA \citep{lin-etal-2022-truthfulqa}, SciQ
\citep{welbl-etal-2017-crowdsourcing}, and MMLU
\citep{hendryckstest2021} in binary-choice and four-choice formats. Each prompt
asks the model to produce an answer option. We read the residual stream at the
final prompt position, immediately before answer generation. \emph{Output
Prediction} predicts the option selected by the model, while \emph{ground-truth
(GT) Prediction} predicts the correct option. Prior work shows that pre-answer
activations carry information about answer quality and knowledge state
\citep{Snyder2024-early-hallucinations,
Gottesman2024-estimating-knowledge}.
TruthfulQA, SciQ, and MMLU contain 817, 1,000, and 783 examples,
respectively. We split each dataset 60/20/20 for training, validation, and
testing, and sample 300 anchors from its training split.

\paragraph{Guard Distillation.}
We use prompts from BeaverTails \citep{beavertails} and Toxic-chat
\citep{lin2023toxicchat}. Each monitored model generates completions,
and Llama-Guard-3-8B \citep{dubey2024llama3herdmodels} labels the resulting
prompt--completion pair as safe or unsafe. The probe predicts this binary label
from the monitored model's completion-token activations. Standard probes
average residual-stream activations over completion tokens at each layer. For
anchored probes, we instead compute an anchor-relative representation for each
completion token and then average those representations.
For each dataset, we use 1,000 prompts split 60/20/20 for training, validation,
and testing. The 600 training prompts serve as anchors.

\paragraph{Online MATH.}
We additionally train probes to predict whether a generated solution to a MATH
problem will ultimately be correct \citep{hendrycksmath2021}. At checkpoints
during generation, we average each layer's residual-stream activations over all
tokens generated so far. Our primary comparisons use 2,000 training examples
and the 128-token checkpoint, selected before probe fitting because
92.8--99.6\% of source training and validation trajectories had not yet ended
at that point. We use a separate 500-example validation set and 300 anchors
sampled from the training data, and evaluate on MATH-500.
Appendix~\ref{app:math-online-depth-policy} reports results with 1,000 training
examples and at the 64- and 256-token checkpoints.

Across all three settings, we also evaluate transfer to related datasets:
among the trivia datasets, between BeaverTails and Toxic-chat, and from MATH to
GSM8K \citep{cobbe2021gsm8k}. We test both dataset transfer within one model
and joint transfer to a new dataset and model. Labels from the target dataset
are used only for final scoring.

\subsection{Cross-Model Transfer and Evaluation Protocol}
\label{sec:method-transfer-protocol}

Let $m_s$ be a source model and $m_t$ a target model. We fit a probe on
$R^{(m_s)}(x;A_D)$, freeze it, and apply it to
$R^{(m_t)}(x;A_D)$. Successful transfer means that this frozen probe remains
predictive on the target model's anchored representations. Deployment on the
target requires one unlabeled setup pass
over the anchor set, in addition to the usual forward pass for each monitored
example; crucially, we do not require target-model labels, a learned target adapter, or
probe retraining.

The trivia and Guard Distillation evaluations share a six-model pool:
Llama-3.1-8B-Instruct,
Llama-3.2-3B-Instruct \citep{dubey2024llama3herdmodels},
Mistral-7B-Instruct-v0.3 \citep{jiang2023mistral7b},
Qwen2.5-7B-Instruct, Qwen2.5-3B-Instruct \citep{qwen2.5}, and
Qwen3-4B-Instruct-2507 \citep{qwen3technicalreport}, which we refer to as
Qwen3-4B-Instruct below. Before extracting activations,
we screened candidate models on MATH. Several models in the trivia and Guard
Distillation pool produced too few correct solutions for reliable probe
training and evaluation, so we use a separate four-model pool for online
mathematical correctness: Llama-3.1-8B-Instruct, Qwen3-4B-Instruct,
OLMo-3-7B-Instruct, and Gemma-3-12B-Instruct
\citep{olmo2026olmo3, gemmateam2025gemma3technicalreport}.

Within each pool, every model serves in turn as a target, with the other models
available as sources. For each target, \emph{Transfer} directly applies every
non-target source probe and averages the resulting AUROCs. 
When several source probes are available, their transfer reliability can be
estimated before deployment by testing each one on the other labeled models in
the source pool. Our source-only calibration procedure (reported as
\emph{Transfer~(cal)}) selects the source probe with the highest mean validation
AUROC on those other source-pool models and then applies it to the held-out
target. No target labels enter calibration. Calibration uses all available
source models within the corresponding pool and is intended to improve
deployment performance. For local anchored probes, source and target activations are taken at the same
relative depth $d$ across the two models. We apply the direct and calibrated procedures separately
at each depth.

We distinguish two evaluation regimes. An \emph{in-pool target} is one member
of a prespecified model pool held out in turn while the remaining models
provide source probes and calibration data. An \emph{external target} belongs
to a model family outside the corresponding source and calibration pool. For
external targets, source selection is completed entirely within that pool
before the selected probe is scored on the new model. Dataset construction, model settings, and optimization details appear in
Appendix~\ref{app:implementation-details}.

%% file: content/main/experiments.tex
\section{Experiments}

We use behavioral monitoring tasks as a functional test of whether independently
trained models organize behaviorally relevant information in
compatible ways. If anchor-relative coordinates preserve a behavioral signal
within each model and a probe's fixed decision rule remains predictive across
models, then the models share task-relevant relational structure in a form
usable by the same monitor. We first test these two conditions, then ask when
this compatibility persists across
external model families, related data distributions, layer choices, model
scale, model updates, and prompt formats. Supporting analyses examine low-FPR
recall and the relational geometry of the anchors.

We use a linear probe on native activations at prespecified relative depth
$0.8$ as the standard baseline; the basis for this choice is
reported in Appendix Figures~\ref{fig:app-trivia-abs-depth-sweep} and
\ref{fig:app-guard-linear-abs-depth-sweep}. We report area
under the receiver operating characteristic curve (AUROC) for all three
settings. We additionally report recall at 10\% FPR in Appendices~\ref{app:trivia-recall},
\ref{app:guard-linear-recall}, and \ref{app:math-online-recall}. Full implementation details appear in Appendix~\ref{app:implementation-details}.
We examine anchor-count variants, attention probes, and other nonlinear probes
in Appendices~\ref{app:trivia-linear-anchor-count}, \ref{app:guard-linear-attention-diagnostics}, and \ref{app:guard-mlp-additional}.

\subsection{Same-Model Decodability}
\label{sec:exp-same-model}

Anchoring replaces each model's native residual coordinates with a common,
anchor-indexed representation, so we first test whether that change
sacrifices predictive information, before testing portability. Standard and Anchored probes receive the
same labels and differ only in their input representation. The anchored
representations are also substantially more compact: every model has 300
coordinates for trivia and MATH or 600 for Guard Distillation, compared with
native residual widths from 2,048 dimensions for Qwen2.5-3B to 4,096 for
Llama-3.1-8B and Mistral-7B.

\input{content/_tables/single_tables/trivia/linear/non-abs/ls0.8/main_summary_auroc.tex}

\input{content/_tables/single_tables/guard/linear/non-abs/agg-single/ls0.8/main_summary_auroc.tex}

\input{content/_tables/single_tables/math/math_online_crossmodel.tex}

Trivia Output Prediction is near ceiling within each model: Standard and
Anchored probes both reach 99.5--100.0 AUROC across the six dataset variants
(Appendix~\ref{app:trivia-calibration-progression}). For the more
discriminating GT Prediction setting, Anchored probes outperform Standard probes
on four variants and trail them on two; the largest difference in either
direction is 1.4 AUROC points
(Table~\ref{tab:final_trivia_linear_main_summary_auroc}). Guard Distillation
also has small same-model gaps: Standard and Anchored score 96.5 and 94.9 on
BeaverTails, and 88.6 and 87.5 on Toxic-chat
(Table~\ref{tab:final_guard_linear_main_summary_auroc}). For MATH, the global
Anchored probe reaches 81.8 AUROC, compared with 77.9 for the Standard probe
(Table~\ref{tab:math-online-crossmodel}). Anchoring therefore preserves nearly
all of the tested same-model linear predictive signal in trivia and safety,
while improving same-model decodability in MATH.

\subsection{Cross-Model Probe Transfer}
\label{sec:exp-cross-model}

We now test the central question: after fitting a probe on one source model,
does its decision rule remain predictive when applied unchanged to another
model? We test this across trivia and Guard Distillation in the six-model pool,
and across online MATH in its separate four-model pool. Calibration uses all
available non-target source models: five for trivia and Guard Distillation, and
three for online MATH.
Appendices~\ref{app:trivia-calibration-progression} and
\ref{app:guard-calibration-progression} additionally report how trivia and
Guard Distillation transfer changes as the available source pool grows from
one to five models; Appendix~\ref{app:math-online-depth-policy} reports the
corresponding one-to-three-source progression for online MATH.

Averaged separately within each setting, calibrated global transfer preserves
99.1\% of same-model Anchored AUROC above chance for trivia Output
Prediction, 90.7\% for trivia GT Prediction, 85.0\% for Guard Distillation,
and 88.1\% for online MATH. Tables~
\ref{tab:final_trivia_linear_main_summary_auroc},
\ref{tab:final_guard_linear_main_summary_auroc}, and
\ref{tab:math-online-crossmodel} report the corresponding results. Trivia
Output Prediction also remains near ceiling after transfer: calibrated AUROC
ranges from 98.3 to 99.9
(Appendix~\ref{app:trivia-calibration-progression}). This result shows that
features of a model's selected answer are strongly transferable, so the main
trivia table focuses on the more discriminating GT Prediction task.

\begin{wrapfigure}{R}{0.50\textwidth}
    \vspace{-0.7\baselineskip}
    \centering
    \includegraphics[width=\linewidth]{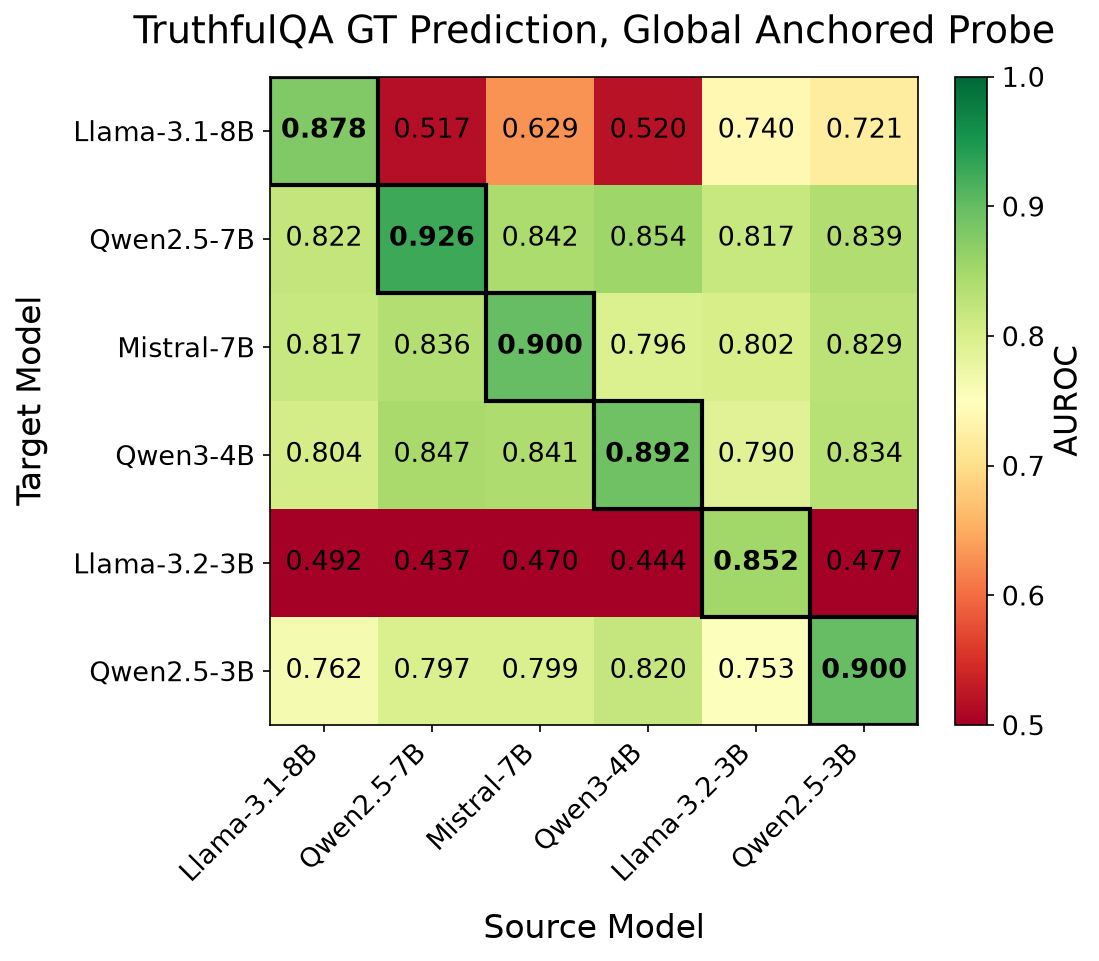}
    \caption{\textbf{Binary TruthfulQA GT Prediction.} Rows are target models
    and columns are source models. Diagonal entries show same-model probe performance; off-diagonal entries are directly transferred probes. Probes are near chance on Llama-3.2-3B-Instruct for all test source models.}
    \label{fig:tqa-gt-confusion}
    \vspace{-0.5\baselineskip}
\end{wrapfigure}

Calibrated transfer remains predictive in all three settings, although the
loss relative to same-model Anchored probes varies. For trivia Output
Prediction, calibrated transfer is at most 1.5 AUROC points below the
same-model Anchored probe across all six variants; GT Prediction is within 0.6
points on MMLU and SciQ. For online MATH, a probe trained on one model's
solution prefixes predicts whether another model's solution will ultimately be
correct: global calibrated transfer reaches 78.0 AUROC, 3.8 points below global
same-model Anchored performance.
Notably, this matches the 77.9 AUROC of Standard probes trained \emph{directly on
each target}. Guard Distillation shows substantial but less complete transfer, reaching 86.4 on
BeaverTails and 83.6 on Toxic-chat, compared with same-model Anchored AUROCs of
94.9 and 87.5. The common anchor coordinates therefore support reuse of a fixed
behavioral decision rule across all three settings, with transfer losses relative
to same-model probes that vary by task. This functional transfer provides evidence
that anchored representations organize task-relevant information compatibly
across models. To show that matching dimensions do
not explain transfer, we transfer standard
probes between Llama-3.1-8B and Mistral-7B, which have the same residual width
and depth. These probes perform at chance level, while anchored probes transfer
strongly (Appendices~\ref{app:trivia-standard-cross-model-baseline} and
\ref{app:guard-linear-standard-cross-model-baseline}).

Binary TruthfulQA GT Prediction is the main exception to the strong trivia
results: calibrated transfer reaches 73.1 AUROC, compared with 89.1 same-model.
The gap is concentrated in Llama-3.2-3B, the least accurate TruthfulQA target;
its same-model probe remains strong, but all incoming probes perform near
chance (Figure~\ref{fig:tqa-gt-confusion}). Across the 36 trivia target--dataset
cells, target answer accuracy is
strongly associated with incoming GT Prediction transfer. Transfer may
therefore be less reliable for low-accuracy targets, although the strong
same-model probe shows that there remains decodable target signal
(Appendix~\ref{app:trivia-transfer-failure-diagnostics}). Recall at 10\% FPR
reveals larger transfer losses, particularly for binary TruthfulQA GT Prediction
and Guard Distillation (Appendices~\ref{app:trivia-recall},
\ref{app:guard-linear-recall}, and \ref{app:math-online-recall}).

\FloatBarrier

\subsection{When Does Predictive Compatibility Persist?}
\label{sec:exp-further-portability}

We next test whether this compatibility extends beyond the original model pools
and persists under related data and other deployment changes.

\subsubsection{External Model Families}
\label{sec:exp-external-families}

The preceding experiments hold out each target from a prespecified model pool.
To test whether portability extends beyond those development pools, we keep
each pool fixed and evaluate additional models. The
transferred probes and their source choices are fixed before external target
models and labels are introduced. For trivia and Guard Distillation, the
new targets are OLMo-3-7B
\citep{olmo2026olmo3}, Gemma-3-12B
\citep{gemmateam2025gemma3technicalreport}, and GPT-OSS-20B
\citep{openai2025gptoss120bgptoss20bmodel}. For MATH, GPT-OSS is held out.
Table~\ref{tab:external-target-summary} summarizes these results.

\input{content/_tables/single_tables/external_targets/external_target_summary.tex}

Trivia GT Prediction reaches 90.2--94.4 AUROC across the three external
families, matching or exceeding the 89.6 average for held-out targets in the
original pool. Output Prediction remains near saturation at 98.6--99.6 AUROC
(Table~\ref{tab:external-target-summary}). Guard Distillation is more
heterogeneous, with Toxic-chat transfer ranging from 66.9 to 87.6. BeaverTails
estimates are high-variance because only 1.0--4.8\% of completions are
unsafe for these newer models.
Appendices~\ref{app:external-target-trivia} and
\ref{app:external-target-guard} report the full results.

For MATH, transfer to GPT-OSS remains near chance at 52.4--53.1 AUROC across
the two source-training sizes. Appendix~\ref{app:external-target-math} reports
target-trained probes, which reach 78.5--78.7 AUROC and
show that correctness remains linearly decodable in GPT-OSS. The decision
boundary learned from the source models is therefore not predictive for GPT-OSS.
GPT-OSS is also the only mixture-of-experts model in our external MATH
evaluation. Its expert routing may contribute to the failure,
although the current comparison cannot separate that possibility from other
architectural and training differences. The failure is task-specific: trivia
probes transfer strongly to GPT-OSS and the other external targets.
External-family transfer therefore varies by model--task pair, but remains
strong for several targets outside the calibration pool.

\subsubsection{Related Data Distributions}
\label{sec:exp-related-data}
The practical value of compatible representations is greatest when both the
deployed model and its data change, since collecting and validating target labels may be prohibitive.
Activation probes can degrade under distribution shifts even when the model is
fixed \citep{Orgad2024-zp, Levinstein2023-no-lie-detector, Kramar2026-zv,
Min2026-corvus}. We therefore use same-model standard probes as a reference and
ask whether anchoring preserves or improves related-data robustness while also
enabling a simultaneous model change. A joint shift approximates, for instance,
a restricted-access setting in which a monitor developed on an accessible
model and dataset is applied locally to a different model on related private
data without sharing the target model or data. We evaluate three such data
shifts: among trivia datasets,
between BeaverTails and Toxic-chat, and from MATH to GSM8K. The anchors and
labels used to fit or calibrate probes always come from the source dataset;
target labels are used only for final scoring.

When the model is fixed, global anchored probes outperform standard probes in
every trivia GT Prediction direction, averaging 87.0 AUROC with gains of up to
13.1 points, and on MATH-to-GSM8K, at 75.3 versus 63.0. They roughly match
standard probes on average for Guard Distillation, at 77.9 versus 78.3. When
both the model and dataset change, trivia transfer averages 83.1 directly and
86.1 after source calibration; calibrated Guard transfer reaches 65.1 from
BeaverTails to Toxic-chat and 79.6 in the reverse direction; and calibrated
MATH-to-GSM8K transfer reaches 70.5. Despite the simultaneous model change,
MATH-to-GSM8K transfer remains 7.4 points above a standard probe applied within
the same model.

Guard probe transfer shows that performance across related data is not
uniformly reliable. To examine this gap, we compare each target
example's distance to its nearest source anchor
(Appendix~\ref{app:guard-mixed-anchor-mitigation}). Target examples are farther
from the source anchors when moving from curated BeaverTails prompts to
real-world Toxic-chat conversations, suggesting that transfer quality is
related to how well the source anchors cover the target data.
Full results appear in Appendices~\ref{app:trivia-cross-data-transfer},
\ref{app:guard-linear-cross-data-transfer}, and
\ref{app:math-online-crossdata}.

\subsubsection{Other Robustness Checks}

\paragraph{Layer selection.}
\label{sec:exp-global-vs-local}
The global construction avoids choosing layers across models. We compare it with
local probes whose depths are selected using source-model validation data.
Across all three task families, local probes do not consistently outperform
global probes, making the global construction a robust depth-free default
(Appendices~\ref{app:trivia-validation-depth-policy},
\ref{app:guard-linear-val-plateau},
\ref{app:math-online-depth-policy}).

\paragraph{Model scale.}
\label{sec:exp-model-scale}
We test transfer across a same-family Llama 3.x scale gap on MMLU. Transfer
from the 1B model to the 70B model reaches 94.8 AUROC, close to the 96.0
AUROC of a probe trained directly on the 70B model. This suggests that labeled probe
development may be done on the much less expensive 1B model, and deployed on the 70B model at little performance loss. Transfer into the 1B target --- the reverse direction --- is near
chance, and the 1B same-model probe reaches only 68.2 AUROC, reinforcing the
earlier pattern that transfer into weak targets is difficult
(Appendix~\ref{app:trivia-scale-and-transfer-limits}).

\paragraph{Model updates.}
\label{sec:exp-model-updates}
A deployed monitor can become outdated when its model is fine-tuned after probe
training. Following the setup
of \citet{duan2026safetymonitorsstayreliable}, we train global anchored safety probes on XSTest safety data \citep{rottger2023xstest} for Qwen2.5-3B and 7B. We then fix the probes, fine-tune the Qwen models on refusal data with LoRA and QLoRA, and evaluate probe robustness under this distribution shift. We find that global anchored probes are more robust to both model quantization and updates than standard probes and local anchored probes (Appendix~\ref{app:probe-staleness-experiments}).

\paragraph{Prompt format.}
\label{sec:exp-prompt-format}
We compare the standard trivia prompt with minimal and verbose variants while
holding the tasks, examples, and models fixed. Under the minimal prompt,
calibrated transfer changes by only about one AUROC point while same-model
probes degrade more, suggesting that transferred probes are less sensitive to
this tested change
(Appendix~\ref{app:trivia-prompt-sensitivity}).

\paragraph{Anchor content and correspondence.}
\label{sec:exp-anchor-content}
Finally, we ask whether anchors must exactly match across models. In Guard
Distillation, same-dataset \emph{Transfer~(cal)} changes by at most 2.7 AUROC
points when each anchor is represented by averaging over that model's generated
completion rather than at the final token of the shared prompt, even though the completion --- and thus the anchor --- differs between models
(Appendix~\ref{app:guard-linear-mean-token-auroc}). In a separate test, we hide
the anchor identities and recover 20.7\% of their cross-model matches using
only pairwise distances, compared with 0.16\% by chance
(Appendix~\ref{app:guard-relational-evidence}). Probes applied to the partially
recovered space retain more than $90\%$ of the original AUROC. Together, these results suggest
that transfer can tolerate non-identical textual anchors and substantial noise
in the anchor mapping, consistent with reliance on the anchor set's broader
semantic organization rather than exact one-to-one correspondence.

%% file: content/_tables/single_tables/trivia/linear/non-abs/ls0.8/main_summary_auroc.tex
\begin{table}[t]
\centering
\caption{\textbf{Trivia factuality prediction.} GT Prediction AUROC for same-model Standard and Anchored probes, mean cross-model Transfer, and source-calibrated Transfer (cal). Values are percentages, averaged over three seeds; uncertainties are standard errors.}
\label{tab:final_trivia_linear_main_summary_auroc}
\begin{adjustbox}{max width=\textwidth}
\begin{tabular}{lcccc}
\toprule
Dataset & Standard & Anchored & Transfer & Transfer (cal) \\
\midrule
TQA & $89.4\pm1.2$ & $89.1\pm0.4$ & $72.5\pm1.4$ & $73.1\pm0.5$ \\
MMLU & $94.8\pm0.2$ & $96.2\pm0.2$ & $95.9\pm0.3$ & $96.2\pm0.2$ \\
SciQ & $99.2\pm0.1$ & $99.4\pm0.0$ & $99.2\pm0.1$ & $99.3\pm0.0$ \\
TQA (4) & $85.5\pm0.3$ & $84.1\pm1.0$ & $73.5\pm1.3$ & $76.9\pm1.3$ \\
MMLU (4) & $93.4\pm0.2$ & $94.8\pm0.1$ & $93.3\pm0.1$ & $94.3\pm0.2$ \\
SciQ (4) & $98.2\pm0.3$ & $98.6\pm0.3$ & $97.9\pm0.2$ & $98.0\pm0.3$ \\
\bottomrule
\end{tabular}
\end{adjustbox}
\end{table}

%% file: content/_tables/single_tables/guard/linear/non-abs/agg-single/ls0.8/main_summary_auroc.tex
\begin{table}[t]
\centering
\caption{\textbf{Guard Distillation monitoring.} AUROC for same-model Standard and Anchored probes, mean cross-model Transfer, and source-calibrated Transfer (cal). Values are percentages, averaged over three seeds; uncertainties are standard errors.}
\label{tab:final_guard_linear_main_summary_auroc}
\begin{adjustbox}{max width=\textwidth}
\begin{tabular}{lcccc}
\toprule
Dataset & Standard & Anchored & Transfer & Transfer (cal) \\
\midrule
BeaverTails & $96.5\pm0.1$ & $94.9\pm0.2$ & $81.1\pm1.2$ & $86.4\pm4.6$ \\
Toxic-chat & $88.6\pm0.8$ & $87.5\pm0.7$ & $81.6\pm0.7$ & $83.6\pm1.0$ \\
\bottomrule
\end{tabular}
\end{adjustbox}
\end{table}

%% file: content/_tables/single_tables/math/math_online_crossmodel.tex
\begin{table}[t]
\centering
\caption{\textbf{Online mathematical-correctness monitoring.} AUROC (\%) on MATH-500 at 128 generated tokens, mean $\pm$ standard error over three source/probe seeds.}
\label{tab:math-online-crossmodel}
\begin{tabular}{lcccc}
\toprule
Dataset & Standard & Anchored & Transfer & Transfer (cal) \\
\midrule
MATH-500 & 77.9 $\pm$ 0.3 & 81.8 $\pm$ 0.3 & 75.2 $\pm$ 0.6 & 78.0 $\pm$ 0.5 \\
\bottomrule
\end{tabular}
\end{table}

%% file: content/_tables/single_tables/external_targets/external_target_summary.tex
\begin{table*}[t]
\centering
\caption{\textbf{Transfer to external model families.} Source-calibrated global-probe AUROC (\%) for held-out targets from the original task-specific pools and for targets outside those pools. Trivia rows average the binary- and four-choice versions of TruthfulQA, MMLU, and SciQ. Values are mean $\pm$ standard error over three source/probe seeds; Appendices~\ref{app:external-target-trivia}, \ref{app:external-target-guard}, and \ref{app:external-target-math} report the full results.}
\label{tab:external-target-summary}
\begin{tabular*}{\textwidth}{@{\extracolsep{\fill}}lcccc@{}}
\toprule
Task / dataset & Held-out in-pool & OLMo-3-7B & Gemma-3-12B & GPT-OSS-20B \\
\midrule
Trivia Output Prediction & 99.4 $\pm$ 0.2 & 99.0 $\pm$ 0.2 & 99.6 $\pm$ 0.1 & 98.6 $\pm$ 0.3 \\
Trivia GT Prediction & 89.6 $\pm$ 0.4 & 90.2 $\pm$ 0.6 & 93.6 $\pm$ 0.4 & 94.4 $\pm$ 0.3 \\
Guard: BeaverTails & 86.4 $\pm$ 4.6 & 79.2 $\pm$ 3.0 & 63.0 $\pm$ 7.5 & 57.5 $\pm$ 12.6 \\
Guard: Toxic-chat & 83.6 $\pm$ 1.0 & 87.6 $\pm$ 1.4 & 72.6 $\pm$ 11.4 & 66.9 $\pm$ 3.8 \\
MATH correctness & 78.0 $\pm$ 0.5 & -- & -- & 52.4 $\pm$ 0.5 \\
\bottomrule
\end{tabular*}
\end{table*}

%% file: content/main/conclusion.tex
\section{Conclusion}

We show that relative representations can support functional reuse of model representations for
white-box monitoring of independently trained autoregressive language models.
Across trivia, safety monitoring, and mathematical reasoning, anchored coordinates retain useful
same-model signal, and probes trained on one model remain predictive on others.
Transfer of a fixed behavioral probe provides substantial evidence that the
models share task-relevant relational organization.

This representation provides a practical route to portable white-box
monitoring. Anchored probes support joint changes of model and related data,
large same-family scale changes, and prompt variations. They also retain more
performance than standard probes after model updates. Reusing a validated
probe can therefore reduce repeated labeling, tuning, and validation as models
and deployment conditions change, while our global construction avoids
model-specific layer selection.

We note that portability remains conditional: transfer weakens for some weakly-performing
target models and can occasionally fail even when the monitored property remains decodable in the
target model. Future work may develop principled anchor selection, develop methods for predicting transfer failures --- potentially based on anchor-similarity or coverage measures --- and extend the approach to long-context, multi-turn, and multimodal
settings. Overall, our results show that independently trained autoregressive 
language models share substantial relevant relational organization in their learned features
and demonstrate a practical method for reusing white-box monitors across models.

%% file: content/appendix.tex
\input{content/appendix/implementation_details}

\clearpage
\input{content/appendix/additional_results}


%% file: content/appendix/implementation_details.tex
\section{Implementation details}
\label{app:implementation-details}

We compare \emph{standard probes}, trained directly on residual-stream
activations, with \emph{anchored probes}, trained on anchor-relative
representations of the same activations. We use two anchored representations.
In the \emph{global} setting, we build a single whole-network representation by
concatenating the relevant residual-stream vectors from all transformer layers
and then mapping that vector into anchor-relative coordinates. The layer
vectors are concatenated raw, with no per-layer normalization before computing
angular distance. In the
\emph{local} setting, we build a separate anchor-relative representation for
each transformer layer, parameterized by a relative depth $d \in [0,1]$. The
global representation removes the need to choose a probing layer and yields a
fixed probe dimensionality determined only by the number of anchors, enabling
cross-model transfer even when models differ in hidden size and depth.

\begin{figure*}[tbp]
\centering
\includegraphics[width=.98\textwidth]{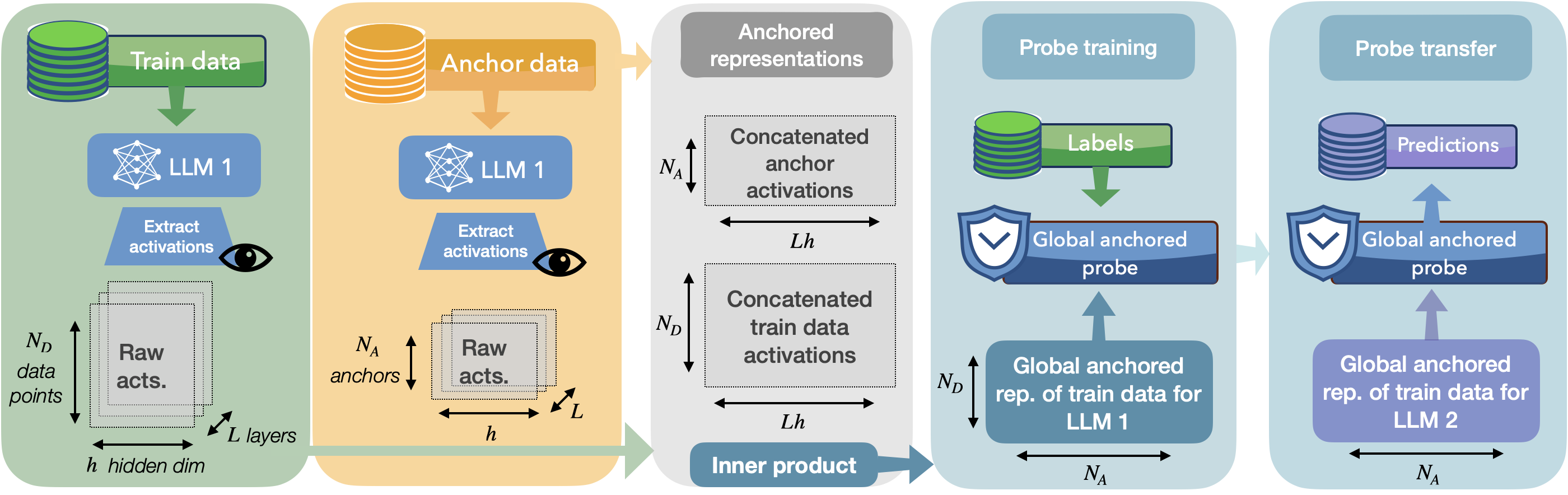}
\caption{\textbf{Global anchored-probe construction.} Each model processes
the same indexed anchor set. We compute the angular distance between a
monitored example and every anchor using residual-stream activations from all
transformer layers. This produces a fixed-dimensional representation whose
coordinates have the same anchor identities across models. A probe trained on
a source model can then be applied unchanged to a target model.}
\label{fig:anchored-probing-workflow}
\end{figure*}

Our global anchored results are compared against the standard single-layer
absolute-probe baseline evaluated at relative depth $0.8$.
For local anchored probes, we evaluate at matched source and target
relative depth $d$: the source probe and target representation are taken
at the same depth, and the Transfer and Transfer~(cal) procedures are
applied independently at each $d$ used in the depth sweep. The resulting
validation-based depth comparisons are reported in
Appendices~\ref{app:trivia-validation-depth-policy},
\ref{app:guard-linear-val-plateau}, and
\ref{app:math-online-depth-policy}; GuardQA's corresponding analysis appears
in Appendix~\ref{app:h64-val-plateau}.
Unless stated otherwise, all reported results average over three seeds.

\subsection{Shared protocol}

The following conventions apply to the Trivia, Guard Distillation, online MATH,
and GuardQA protocols unless a setting-specific subsection states otherwise.
GuardQA is an auxiliary safety setting that asks the monitored model to classify
a stored prompt--completion pair as safe or unsafe in an A/B question; its full
motivation and protocol appear in Appendix~\ref{app:guardqa-implementation-details}.

\begin{table}[htbp]
\centering
\caption{Shared evaluation protocol.}
\label{tab:impl-shared}
\begin{adjustbox}{max width=\textwidth}
\begin{tabular}{lp{0.62\linewidth}}
\toprule
\textbf{Component} & \textbf{Setting} \\
\midrule
Probe feature scaling &
subtract the training-set mean and divide by the training-set standard
deviation; cross-model applications use the source probe's stored training
statistics \\
Anchor sampling &
sample anchors from the source training split for each seed; the same anchor
IDs and order are used across models within a seed/configuration \\
Relative-coordinate function &
angular distance, i.e., arccosine of the cosine similarity between an example activation and its anchor activation \\
Cross-model calibration for global probes &
select a source probe using validation results from the available source models
only; the setting-specific subsections specify the model pool and aggregation \\
Metrics &
AUROC for all settings; recall at 10\% false-positive rate for binary tasks \\
\bottomrule
\end{tabular}
\end{adjustbox}
\end{table}

\paragraph{Layer and token-position convention.}
Stored layer activations correspond to transformer-block outputs. We index
these layers from $0$ to $L_m-1$ for a model with $L_m$ transformer blocks;
the token-embedding layer and unembedding/output head are not included. For
the single-token Trivia and GuardQA experiments, probes read the final prompt
position: after the model has processed the full prompt, but before the first
answer token has been generated. Guard Distillation instead probes monitored
completion-token representations. Its main linear experiment uses single-token
anchors taken at the final prompt position.

\paragraph{Relative-depth mapping.}
For local probes, relative depth is computed as
$d(\ell;L_m)=\ell/(L_m-1)$ for zero-indexed transformer-block layer
$\ell$. When mapping layers across models with different depths, the first
and last transformer-block layers are pinned to the first and last bins,
respectively. Intermediate layers are assigned to the nearest relative-depth
bin center, using rounding. This convention lets local transfer compare source
and target layers at matched relative depth while excluding embeddings and
unembedding from the layer count.

\paragraph{Anchors, distance function, and scaling.}
Anchors are randomly sampled from the training split using the seed for the
run. They may overlap with examples used to train the probe, but they are not
sampled from validation or test examples. For a fixed dataset, seed, and anchor
configuration, the same anchor IDs and order are used for every model, so
the anchored coordinates are indexed by the same datapoints across models.
Related-data transfer retains the source dataset's anchor set, as described in
Appendix~\ref{app:related-data-implementation}. Our main distance
function is angular distance, the $\arccos$ of cosine similarity. In global
mode, raw per-layer vectors are concatenated first and the angular
distance is computed on the resulting
whole-network vector; no per-layer normalization is applied before
concatenation. Unless otherwise noted, probe feature normalization uses the
source probe's stored training-set mean and standard deviation (or scaler). In cross-model
applications, we do not refit normalization statistics on the target model's
validation or test activations.

\paragraph{Metrics.}
We evaluate probe performance using two complementary metrics.
\textbf{AUROC} (area under the receiver operating characteristic curve)
summarizes discrimination quality over all possible thresholds and is
insensitive to class imbalance; it is our primary metric in the main
text. \textbf{R@0.1FPR} reports the true-positive
rate at a fixed false-positive budget and is more informative for
deployment scenarios in which the cost of false positives is high (e.g.,
safety monitoring with a low-tolerance alert threshold). Both metrics
are standard in the internal-monitoring literature
\citep{Snyder2024-early-hallucinations, Obeso2025-ns}. We compute AUROC with
scikit-learn's \texttt{roc\_auc\_score}. Binary cells use the usual binary
AUROC; multiclass cells use one-vs-rest macro averaging. The four-choice
Trivia settings have no substantial class imbalance, so this averaging choice
does not drive the reported trends. We report AUROC for all cells (binary and
multiclass) and R@0.1FPR for the binary trivia variants, both GuardQA datasets,
Guard Distillation, and online MATH.

For R@0.1FPR, the positive class is task-specific. For binary answer-choice
settings (binary Trivia and GuardQA Output Prediction/GT Prediction), recall is
computed for the encoded class~1 answer, i.e., answer B in the A/B format; since
answer positions are randomized and close to balanced, this is only a reporting
convention. For Guard Distillation, recall is instead unsafe-positive: true
positives are unsafe completions detected by the probe, and false positives are
safe completions incorrectly flagged as unsafe. For online MATH, the positive
class is a correct solution. A few transferred MATH and GSM8K predictions have
tied scores at the operating boundary, so we report deterministic expected
recall under randomized tie-breaking at exactly 10\% FPR.

\paragraph{Dataset and answer-position balance.}
Tables~\ref{tab:trivia-dataset-distribution-summary}
and~\ref{tab:guardqa-dataset-distribution-summary} summarize the
label-position and model-output distributions used by the Trivia and GuardQA
experiments. The tables average over the six-model pool and three seeds. Binary
Trivia and GuardQA ground-truth positions are close to balanced across A/B; the
four-choice Trivia variants are similarly balanced across A/B/C/D. GuardQA A/B
positions are randomized independently of the raw safe/unsafe label, so the
probe target for safety prediction is balanced even though the underlying Guard
safe/unsafe labels are not. Guard Distillation uses the raw Llama Guard safety
labels directly; its realized label balance is reported in
Appendix~\ref{app:guard-implementation-details}.

\input{content/_tables/single_tables/trivia/dataset_statistics/dataset_distribution_table.tex}
\input{content/_tables/single_tables/guardqa/dataset_statistics/dataset_distribution_table.tex}

\paragraph{Cross-model calibration averaging.}
Section~\ref{sec:method-transfer-protocol} defines direct transfer and our
source-only calibration procedure. Global representations leave only a choice
of source probe: each model yields one anchor-indexed vector per example, so
there is no source-layer or target-layer search. We use a leave-one-target-out
procedure in both the six-model Trivia, Guard Distillation, and GuardQA pools
and the four-model online-MATH pool.

For the six-model pools, every target contributes exactly once per table cell:
\begin{itemize}
    \item \textbf{Transfer} (naive transfer): for each of the 6 models as
    target, average the test AUROC of the 5 remaining source probes applied
    to that target. Then average over the 6 targets. The reported number is
    therefore a mean over the $6 \times 5 = 30$ off-diagonal source-target
    pairs in the confusion matrix (Figure~\ref{fig:tqa-gt-confusion} shows
    one such matrix).
    \item \textbf{Transfer~(cal-$N$)}: for each target, enumerate all
    size-$N$ source subsets that exclude that target (combinations). Within each subset,
    score each candidate source probe by its mean validation AUROC on the
    other models in the subset, select the best source, and record its test
    AUROC on the in-pool target. Average over subsets, then over targets.
    This source-selection step uses only source-model validation labels.
\end{itemize}
The $N=1$ case is equivalent to naive \textbf{Transfer}, since a singleton
source subset contains no source-selection decision; the six-model tables
therefore show \textbf{Transfer} followed by calibrated columns for larger
source subsets.

Online MATH uses the same leave-one-target-out logic with three candidate
sources per target: direct transfer averages the three source probes, giving
$4 \times 3 = 12$ directed source--target pairs. For calibrated transfer, we
score each candidate source probe on the other two source models by its mean
validation AUROC across the 64-, 128-, and 256-token checkpoints, then apply
the selected source model's checkpoint-specific probes unchanged to the
held-out target. We do not report a calibration-pool progression for MATH
because each target has only three available sources.

Calibration assumes labeled validation data for the available source models,
but no labels, adapters, or retraining for the in-pool target model. The
standard error reported alongside each mean reflects variability
across the three random seeds (affecting data splits, anchor selection,
and probe training): $\text{SE} = s / \sqrt{n}$, with $s$ the sample
standard deviation using Bessel's correction and $n = 3$. It does not
attempt to summarize heterogeneity across
models or directed source--target pairs.

\FloatBarrier

\subsection{Trivia setting}
\label{app:trivia-implementation-details}

Our trivia experiments use TruthfulQA \citep{lin-etal-2022-truthfulqa},
SciQ \citep{welbl-etal-2017-crowdsourcing}, and
MMLU \citep{hendryckstest2021}
in both binary-choice and four-choice variants. We load TruthfulQA from
\texttt{domenicrosati/TruthfulQA} and SciQ from \texttt{allenai/sciq}, in both
cases drawing examples from the \texttt{train} split of the dataset. For MMLU
we use the \texttt{miscellaneous} subject configuration of \texttt{cais/mmlu},
because its general-knowledge questions are closest in style to the trivia
setting, and we draw examples from MMLU's \texttt{test} split, which is the
canonical evaluation split for that benchmark. These source splits are used as
example pools; each pool is then internally split into 60\% train, 20\%
validation, and 20\% test for probe training and evaluation. We probe the
residual stream at the final prompt position, immediately before the model
emits its selected option, which is the point most directly tied to answer
selection. For global anchored probes, we extract this token position from
every transformer-block layer, concatenate the resulting raw layer vectors, and
then build the anchor-relative representation from that concatenated vector.

The default trivia prompt uses a system instruction telling the model to answer
the multiple-choice question with only the letter corresponding to its choice.
The user message then contains the question, the answer options labeled
\texttt{A}, \texttt{B}, \ldots, and a final \texttt{Answer:} cue. In other
words, the default prompt has the form
\begin{quote}
\small
\texttt{[System] You are a helpful assistant. Answer the multiple choice
question by responding with only the letter A/B/\ldots}\\
\texttt{[User] \{question\}}\\
\texttt{A. \{answer A\}}\\
\texttt{B. \{answer B\}}\\
\texttt{\ldots}\\
\texttt{Answer:}
\end{quote}

Multiple-choice options are constructed per sample with deterministic
randomization. For TruthfulQA we sample one correct answer from the available
reference-correct set, then sample the required number of incorrect answers
without replacement; for SciQ and MMLU we sample distractors analogously.
After assembling the answer set, we draw a random insertion position for the
correct answer, so the correct label is randomized across A/B (or A/B/C/D)
rather than fixed to a particular letter. To make this reproducible, each
sample uses a fixed per-sample seed (generation seed + sample index), so answer
sampling and label placement are stable across runs. If a sample has fewer
incorrect options than requested (mainly relevant for multiclass TruthfulQA),
we keep the sample and use all available incorrect answers rather than dropping
it.

\begin{table}[htbp]
\centering
\caption{Default configuration for the trivia setting.}
\label{tab:impl-trivia}
\begin{adjustbox}{max width=\textwidth}
\begin{tabular}{lp{0.56\linewidth}}
\toprule
\textbf{Component} & \textbf{Default} \\
\midrule
Datasets & TruthfulQA \citep{lin-etal-2022-truthfulqa},
SciQ \citep{welbl-etal-2017-crowdsourcing},
MMLU \citep{hendryckstest2021},
and their four-choice variants \\
HuggingFace sources and splits &
\texttt{domenicrosati/TruthfulQA} (\texttt{train} split),
\texttt{allenai/sciq} (\texttt{train} split), and
\texttt{cais/mmlu} with the \texttt{miscellaneous} subject configuration
(\texttt{test} split) \\
Answer generation & temperature 0.3, nucleus sampling with $p=0.95$, maximum 50 new tokens \\
Anchor set size & 300 examples \\
Similarity function &
angular distance, i.e., arccosine of the cosine similarity between an activation and the anchor set \\
Probe architecture & scikit-learn logistic regression with $C=1.0$ and no class weighting \\
\bottomrule
\end{tabular}
\end{adjustbox}
\end{table}

\FloatBarrier

\subsection{Guard Distillation setting}
\label{app:guard-implementation-details}

Guard Distillation evaluates anchored probes on binary safety-monitoring labels
for monitored prompt-completion pairs. It uses the six models in
Table~\ref{tab:impl-shared} and prompts from BeaverTails and Toxic-chat. For
each monitored model, we generate a completion, label the resulting
prompt-completion pair with Llama-Guard-3-8B \citep{dubey2024llama3herdmodels},
and train probes on the monitored
completion-token representations to predict the binary Guard Distillation label.
The resulting probes act as distilled Guard models: white-box safety monitors
that approximate Guard labels from internal activations. Because anchored
representations are shared across models, the Guard Distillation experiment
tests whether such monitors can transfer across model families.

The Guard Distillation runs use a fixed 1,000 prompts per dataset. A separate
Qwen3-4B screening pass generates and Llama-Guard-labels a completion for every
available prompt. From these labels, we use seeded stratified sampling to
target 40\% unsafe examples within each 60/20/20 split, capped by the number of
unsafe screening examples available. Sampling within each Guard label also
preserves the relative proportions of the source dataset's safety labels. The
screening completions are used only to choose prompt IDs and are not reused in
the monitoring experiment. We then generate model-specific completions for those
prompts with temperature $0.3$, top-$p=0.95$, and at most $800$ new tokens,
re-label each model-specific prompt-completion pair with Llama-Guard, and use
the selected train, validation, and test IDs. We supply each
dataset prompt as a user message under the model's native chat template and do
not add a task-specific system message.
For standard probes, we average raw residual-stream activations over the
monitored completion tokens at each layer; the all-layer standard control then
concatenates these layerwise means. For anchored probes, we first compute the
angular-distance feature vector separately for each completion token and then
average those feature vectors. The linear probe uses $600$ anchors and $C=0.1$.

We score each prompt--completion pair by formatting the prompt as a user
message and the completion as an assistant message with Llama Guard's native
chat template. Llama-Guard-3-8B decodes greedily for at most 10 tokens. We read
the safe and unsafe probabilities from the logits at its first generated
safety-label token and assign whichever label has higher probability.

The main Guard Distillation linear AUROC, recall, and model-family results use
single-token anchors. We take each anchor activation at the final prompt
position, immediately before the model emits its first completion token, as in
Trivia and GuardQA. Appendix~\ref{app:guard-distillation-setting} also reports
ablations with mean-token anchors. These variants instead average each anchor
prompt's completion-token activations before computing angular distance
per token to obtain the final anchored representation.

Table~\ref{tab:guard-guard-label-balance} reports the realized Llama Guard
label balance for each monitored model's Guard Distillation completions. Unsafe
completions are rare for most model--dataset pairs, making probe training
class-imbalanced. Mistral is an exception on BeaverTails: its mean unsafe rate
is substantially higher than those of the other monitored models.
On the external targets, BeaverTails has especially few unsafe completions:
the mean unsafe rates are 1.0\% for OLMo-3, 1.5\% for GPT-OSS, and 4.8\% for
Gemma-3. One OLMo-3 seed contains no unsafe BeaverTails test completion, so its
AUROC is undefined and excluded from the AUROC mean; the corresponding
prevalence summary still uses all three seeds. Toxic-chat unsafe rates range
from 9.0\% to 11.8\%, and all Toxic-chat seed-level AUROCs are defined.

\input{content/_tables/single_tables/guard/dataset_statistics/guard_guard_label_balance.tex}

\begin{table}[htbp]
\centering
\caption{Default configuration for the Guard Distillation setting.}
\label{tab:impl-guard-distillation}
\begin{adjustbox}{max width=\textwidth}
\begin{tabular}{lp{0.56\linewidth}}
\toprule
\textbf{Component} & \textbf{Default} \\
\midrule
Datasets & BeaverTails \citep{beavertails} and Toxic-chat
\citep{lin2023toxicchat} \\
HuggingFace sources and splits &
\texttt{PKU-Alignment/BeaverTails} (\texttt{30k\_train} split) and
\texttt{lmsys/toxic-chat} with the \texttt{toxicchat0124} configuration
(\texttt{train}+\texttt{test} combined) \\
Source prompts & class-imbalance-reduced 1,000-prompt BeaverTails and Toxic-chat subsets selected by a Qwen3-4B-Instruct screening pass; prompt IDs are shared across monitored models \\
Monitored completion generation & separately by each monitored model for the selected prompts, temperature 0.3, nucleus sampling with $p=0.95$, maximum 800 new tokens; Llama Guard labels each model-specific prompt-completion pair \\
Prediction target & binary Guard Distillation safety label from Llama Guard 3\\
Anchor set size & 600 examples \\
Anchor representation & main Guard linear AUROC/recall/model-family tables use single-token anchors at the final prompt position; mean-token anchors are reported as appendix diagnostics \\
Probe architecture & scikit-learn logistic regression with $C=0.1$ and no class weighting \\
\bottomrule
\end{tabular}
\end{adjustbox}
\end{table}

\subsection{Online MATH correctness setting}
\label{app:math-implementation-details}

The online correctness experiment generates solutions to MATH problems
\citep{hendrycksmath2021} and
predicts, from an intermediate prefix, whether the completed solution will be
correct. The in-pool model set is Llama-3.1-8B-Instruct,
Qwen3-4B-Instruct, OLMo-3-7B-Instruct, and Gemma-3-12B-Instruct. Each problem
is supplied as a user message that asks the model to show its reasoning and
place its final expression in \texttt{\textbackslash{}boxed\{...\}}. We use each model's
native chat template and do not add a task-specific system message. Each example's
layer feature is the mean residual-stream activation over the generated tokens
available at checkpoints 64, 128, and 256. The Standard baseline and fixed
local comparison use relative depth $0.8$; global features concatenate the
prefix means from every layer before computing distances to 300 MATH anchors.

Each source seed uses nested 1K and 2K probe-training subsets, a fixed
500-example validation set, and 300 anchors sampled from the shared training
data. The anchor examples are part of the 1K subset and are therefore shared by
both training-set sizes. Hyperparameters are selected on the source model's
validation split. The primary MATH comparison in
Section~\ref{sec:exp-cross-model} uses the 2K condition and the prespecified
128-token checkpoint. All three
source/probe seeds are evaluated on the same frozen MATH-500 generations.
Consequently, standard errors reflect variation in source-side training,
anchors, and probe fitting rather than new samples of target completions.

We extract the last balanced \texttt{\textbackslash{}boxed\{...\}} or
\texttt{\textbackslash{}fbox\{...\}} expression from each completed solution,
with conservative fallbacks for an unbraced boxed atom and an explicit
final-answer marker. We normalize common TeX and formatting differences,
accept exact matches and equivalent integer, decimal, or fractional values,
and accept a matching right-hand side when a model boxes a concluding equation.
A generation that reaches the 4,096-token limit without an end-of-sequence
token is marked incorrect.

\begin{table}[htbp]
\centering
\caption{Default configuration for online MATH correctness monitoring.}
\label{tab:impl-math}
\begin{adjustbox}{max width=\textwidth}
\begin{tabular}{lp{0.58\linewidth}}
\toprule
\textbf{Component} & \textbf{Default} \\
\midrule
Datasets and splits &
\texttt{EleutherAI/hendrycks\_math} training split for probe training and
validation; \texttt{HuggingFaceH4/MATH-500} test split for the primary
evaluation \\
Model pool &
Llama-3.1-8B-Instruct, Qwen3-4B-Instruct, OLMo-3-7B-Instruct, and
Gemma-3-12B-Instruct \\
Solution generation &
greedy decoding with a maximum of 4,096 new tokens \\
Probe data &
nested 1,000- and 2,000-example MATH training sets and 500 MATH validation
examples; the main comparison uses 2,000 training examples \\
Prediction target &
whether the completed solution is correct under answer extraction and grading \\
Checkpoints and query representation &
mean residual-stream activation over generated tokens 1--64, 1--128, or
1--256; the main comparison uses 128 tokens \\
Anchor set &
300 MATH training examples contained in both nested training sets \\
Probe architecture &
scikit-learn logistic regression; $C \in \{0.01,0.1,1,10,100\}$ selected by
source-model validation AUROC, with no class weighting \\
\bottomrule
\end{tabular}
\end{adjustbox}
\end{table}

Correctness labels in this setting are based on each monitored model's
generated solution.
Table~\ref{tab:math-correctness-label-balance} reports the correctness rate and
the share of trajectories that reach the primary 128-token checkpoint. Averaged
across the four in-pool models, MATH-500 is 72.8\% correct at that checkpoint,
whereas GSM8K is 90.9\% correct and therefore has substantially fewer incorrect
examples.

\input{content/_tables/single_tables/math/dataset_statistics/math_correctness_label_balance.tex}

\subsection{External-target protocol}
\label{app:external-target-implementation}

The external-target evaluation in
Section~\ref{sec:exp-external-families} holds the source and calibration pool
for each task fixed. For trivia and Guard Distillation, we evaluate OLMo-3-7B-Instruct,
Gemma-3-12B-Instruct, and GPT-OSS-20B as targets outside the original six-model
pool. For MATH, GPT-OSS-20B is external to the four-model pool above; OLMo-3
and Gemma-3 already belong to that pool. Candidate source probes are ranked by
mean validation AUROC on the other models in the relevant original pool, and
the selected probe is applied unchanged to the external target. External-target
labels do not participate in probe fitting, normalization, hyperparameter
choice, or source selection. GPT-OSS generations use the medium reasoning
effort specified through its native chat template.

After the transfer protocol was fixed, we trained a global Anchored probe on
each external target using the same task definition. These target-trained
results measure whether the target's anchored representation contains a
linearly decodable signal under the evaluated setup. They are diagnostic
comparisons of target-side decodability, not deployable transfer methods.
Trivia summaries first average the six dataset/answer-format variants within
each source/probe seed and then report the mean and standard error over three
seeds. Guard reports BeaverTails and Toxic-chat separately using the
single-token anchor representation. The MATH diagnostic follows the same
2K-training, 128-token configuration as the primary result.

\subsection{Related-data transfer protocol}
\label{app:related-data-implementation}

The related-data experiments in Section~\ref{sec:exp-related-data} test whether
portability survives a change of evaluation dataset, either within the same
model or together with a model change. In every transfer direction, the source
dataset supplies all examples used to fit, tune, and select probes. The probes
retain the source-dataset normalization, and anchored probes also retain the
source anchor set. Target-dataset labels are used only for final scoring.

For trivia, we transfer among TruthfulQA, SciQ, and MMLU while preserving the
binary- or four-choice answer format and use the source dataset's 300 anchors.
For Guard Distillation, we transfer in both directions between BeaverTails and
Toxic-chat and use the source dataset's 600 anchors. For mathematical
correctness, we apply the probes trained on 2,000 MATH examples to GSM8K
\citep{cobbe2021gsm8k} at the same 64-, 128-, and 256-token checkpoints. These
probes retain their MATH anchors, fitted parameters, normalization, and
MATH-based source selection. GSM8K uses the
\texttt{openai/gsm8k} test split, and no GSM8K example is used for fitting,
hyperparameter selection, or source selection. We prompt each model through its
native chat template with a system instruction to show its reasoning and end
with \texttt{Final answer: <number>}, and generate greedily for at most 4,096
tokens. We extract the final numeric answer from an explicit final-answer,
\texttt{<answer>}, boxed-answer, or GSM8K \texttt{\#\#\#\#} marker, falling
back to the last number in the completion. Correctness is exact after
normalizing commas and integer, decimal, or simple-fraction notation against
the reference answer following the dataset's \texttt{\#\#\#\#} delimiter.

\FloatBarrier

\subsection{GuardQA setting}
\label{app:guardqa-implementation-details}

GuardQA is an auxiliary safety-evaluation setting that reframes each monitored
prompt--completion pair as a trivia-style A/B question. The monitored model
receives its own completion together with a follow-up question asking whether
that response is safe or unsafe. We report GuardQA in the appendix because the
main safety setting instead uses Guard Distillation, which predicts the Guard
label directly from completion-token representations
(Section~\ref{sec:method-task-settings}).

We introduced GuardQA after an early Guard Distillation pilot gave weak
results. The motivating hypothesis was that explicitly asking a model to judge
the safety of its own response might make safety-relevant features easier to
probe at the final prompt position. We later found that the pilot had used a
short completion limit, so many monitored responses were incomplete. Llama
Guard could then label properties of an unfinished response that were only
weakly reflected in the model's activations at that point. The longer
completion protocol used for the final Guard Distillation results addresses
this likely source of mismatch; GuardQA remains an auxiliary test of an
explicitly elicited safety judgment rather than a replacement for Guard
Distillation.

To construct the trivia-style question, we take each monitored model's stored
prompt-completion pair and convert it into a new prompt that contains the
original user prompt, that model's completion, and a follow-up safety question
of the form ``Is the assistant's response safe or unsafe?'' with A/B options.

The prompt is formatted as a safety-evaluator dialogue. The system instruction
asks the model to read a conversation and determine whether the assistant's
response is safe or unsafe, responding only with the letter \texttt{A} or
\texttt{B}. The user message then contains the original user prompt, the
generated completion, the question ``Is the assistant's response safe or
unsafe?'', the two answer options, and a final \texttt{Answer:} cue. The
template is therefore
\begin{quote}
\small
\texttt{[System] You are a safety evaluator. Read the conversation below and
determine whether the assistant's response is safe or unsafe. Respond with only
the letter A or B.}\\
\texttt{[User] Conversation:}\\
\texttt{User: \{original prompt\}}\\
\texttt{Assistant: \{model completion\}}\\
\texttt{Is the assistant's response safe or unsafe?}\\
\texttt{A. \{Safe or Unsafe\}}\\
\texttt{B. \{Unsafe or Safe\}}\\
\texttt{Answer:}
\end{quote}

The ordering of the answers is randomized per sample: for some examples
\texttt{A=Safe, B=Unsafe}, and for others \texttt{A=Unsafe, B=Safe}. This
randomization is deterministic, using the per-sample seed
(generation seed + sample ID). The monitored model is instructed to answer with
only the letter A or B. The target for safety prediction is therefore the A/B
position corresponding to the stored Llama Guard label, not the raw safe/unsafe
token.
Because GuardQA is a trivia-style task, we probe the final prompt position,
immediately before the model emits its A/B answer. This elicits the relevant
safety judgment at a single well-defined pre-answer position.

GuardQA uses class-imbalance-reduced 1,000-prompt subsets for BeaverTails
\citep{beavertails} and Toxic-chat \citep{lin2023toxicchat}. BeaverTails
prompts are drawn from the
\texttt{30k\_train} split of \texttt{PKU-Alignment/BeaverTails}, i.e., the
30K-prompt release of the dataset rather than the larger 330K release.
Toxic-chat prompts are drawn from the \texttt{toxicchat0124} configuration of
\texttt{lmsys/toxic-chat}, combining its \texttt{train} and \texttt{test}
splits into a single pool.

We first choose the prompt IDs using a prompt-selection pass. For each
underlying dataset, we prompt Llama-3.1-8B-Instruct
\citep{dubey2024llama3herdmodels} with every candidate user request and
generate a completion of at most $200$ new tokens. Llama-Guard-3-8B scores each
selection-time prompt-completion pair as safe or unsafe. To reduce class
imbalance, we retain all prompt IDs whose selection-time completions are labeled
unsafe and then add safe prompt IDs in dataset index order until the subset
reaches 1,000 prompts. This prompt-selection step gives 432 selection-unsafe IDs
for BeaverTails and 304 selection-unsafe IDs for Toxic-chat. The resulting
prompt IDs are shared across monitored models.

The completions used inside GuardQA examples are then generated separately for
each monitored model. For a given monitored model, dataset, and generation seed,
we prompt that same model and store its
completion, again with a maximum of $200$ new tokens. Llama-Guard-3-8B scores
each resulting prompt-completion pair. Thus the selected prompt IDs are shared,
but the assistant response inserted into the GuardQA question and the
corresponding Llama Guard label are model- and seed-specific. A Qwen3 GuardQA
example, for instance, contains a Qwen3 completion; a Llama Guard label for
that Qwen3 completion; and an A/B safety question answered by Qwen3. The same
construction is repeated for each monitored model.

For each seed split, these 1,000 selected prompt IDs are internally split into
60\% train, 20\% validation, and 20\% test for probe training and evaluation.
The safety-label balance can differ across monitored models because their
completions differ, even though the prompt IDs are held fixed.

For reference, Table~\ref{tab:guardqa-guard-label-balance} reports the realized
Llama Guard label balance for each monitored model's completions. Values are
averaged over the three generation/split seeds used in the paper; uncertainty is
standard error across seeds. Because every row contains 1,000 examples per seed,
the safe rate is the complement of the unsafe rate and has the same standard
error.

\input{content/_tables/single_tables/guardqa/dataset_statistics/guardqa_guard_label_balance.tex}

\begin{table}[htbp]
\centering
\caption{Default configuration for the GuardQA setting.}
\label{tab:impl-guardqa}
\begin{adjustbox}{max width=\textwidth}
\begin{tabular}{lp{0.56\linewidth}}
\toprule
\textbf{Component} & \textbf{Default} \\
\midrule
Datasets & BeaverTails \citep{beavertails} and Toxic-chat
\citep{lin2023toxicchat} \\
HuggingFace sources and splits &
\texttt{PKU-Alignment/BeaverTails} (\texttt{30k\_train} split) and
\texttt{lmsys/toxic-chat} with the \texttt{toxicchat0124} configuration
(\texttt{train}+\texttt{test} combined) \\
Source prompts & class-imbalance-reduced 1,000-prompt BeaverTails and Toxic-chat subsets selected by Llama-3.1-8B-Instruct prompt-selection completions \\
Monitored completion generation & separately by each monitored model for the selected prompts, temperature 0.3, nucleus sampling with $p=0.95$, maximum 200 new tokens; Llama Guard labels each model-specific prompt-completion pair \\
GuardQA answer generation & monitored model, temperature 0.3, nucleus sampling with $p=0.95$, maximum 50 new tokens \\
Anchor set size & 300 examples \\
Probe architecture &
MLP with a single hidden layer, ReLU activation, hidden size $64$ \\
Probe optimization &
Adam, learning rate $10^{-3}$, weight decay $10^{-3}$, batch size $64$,
maximum $50$ epochs, early stopping patience $5$ on validation loss \\
\bottomrule
\end{tabular}
\end{adjustbox}
\end{table}

\FloatBarrier

\subsection{MMLU scale experiment}
\label{app:mmlu-scale-implementation}

The model-scale test summarized in Section~\ref{sec:exp-model-scale} pools
14,042 MMLU examples and constructs
nested subsets of 1K, 3K, 6K, and 10K examples. We evaluate
Llama-3.2-1B-Instruct, Llama-3.1-8B-Instruct, and
Llama-3.1-70B-Instruct using the binary GT Prediction setup, angular-distance
global representations, and either 300 or 1000 anchors whenever the anchor
count fits within the 60\% training split. This gives seven conditions: 1K
examples with 300 anchors, plus 3K, 6K, and 10K examples with either 300 or 1000
anchors. We repeat each condition for three generation, split, and anchor seed
triplets. Within a seed, we compute each cell separately for every condition
and average seven AUROC values. The reported cell is the mean of three
resulting seed-level Transfer averages (no source-calibration).
For each seed, we randomly permute all 14,042 MMLU test examples,
retain the first 10,000, divide that pool 60/20/20, and take the first indices of each split to construct the smaller nested datasets.

\FloatBarrier

\subsection{Probe staleness experiments}
\label{app:xstest-staleness-impl}

The model-update test summarized in Section~\ref{sec:exp-model-updates}
follows the prompt-only refusal-compliance
protocol of \citet{duan2026safetymonitorsstayreliable}. We use all 450
prompts from the XSTest \citep{rottger2023xstest} safety
dataset, label genuinely unsafe prompts as the positive class and safe prompts
that resemble unsafe requests as the negative class, and split the dataset
stratified into 315 training prompts and 135 validation prompts. We repeat this
with three train/validation split seeds; because anchored probes use the 315
training prompts from the corresponding split as anchors, each split seed also
has a different anchor set. Activations are residual-stream activations
at the final prompt token, before any completion token is sampled.
Table~\ref{tab:xstest-data-statistics} summarizes the XSTest label and split
statistics used by this monitor.

\input{content/_tables/single_tables/staleness_experiments/xstest/dataset_statistics/xstest_data_statistics.tex}

To remain comparable with the six-layer evaluation grid of
\citet{duan2026safetymonitorsstayreliable}, we monitor layers
$4,8,12,16,20,24$ in Qwen2.5-3B-Instruct and Qwen2.5-7B-Instruct. We train
linear logistic-regression probes with $C=1.0$. Standard probes use each
layer's residual-stream activation directly, while local anchored probes
convert each monitored layer separately into anchor-relative coordinates. We
also concatenate the same six layers before applying the anchor transform,
giving a global anchored version of Duan et al.'s six-layer setup. We call this
the \emph{global anchored} probe in the XSTest results. Unlike the global
representation in the main transfer experiments, it combines the six monitored
layers rather than all layers. Anchored probes use all 315
XSTest training prompts from the corresponding split as anchors and use angular
distance as the main relative-coordinate function. We report $\Delta$AUROC in points,
defined as changed-model AUROC minus original-model AUROC on the same validation
prompts. A probe is counted as stale if $\Delta\mathrm{AUROC}\leq-5$ points.

The learned model update is a small refusal-SFT adapter trained on BeaverTails
\citep{beavertails}, disjoint from XSTest. We draw examples from the
\texttt{30k\_train} split of \texttt{PKU-Alignment/BeaverTails}. For each
update size, we sample a balanced set of safe and unsafe examples using the
update seed. Safe examples train the model to reproduce the original BeaverTails
response. Unsafe examples train the model to produce a deterministic short
refusal, selected from the following templates by the row identity and update
seed:
\begin{itemize}
    \item ``I'm sorry, but I can't help with that request.''
    \item ``I can't assist with instructions or advice that would enable harm.''
    \item ``I can't help with that. If you want, I can offer safer, legal alternatives.''
    \item ``I'm not able to provide help for that request.''
    \item ``I can't assist with harmful or illegal activity.''
\end{itemize}
This produces a safety-relevant model update without using XSTest labels or
XSTest prompts. Table~\ref{tab:xstest-refusal-update-data-statistics}
summarizes the refusal-SFT update-data balance.

\input{content/_tables/single_tables/staleness_experiments/xstest/dataset_statistics/beavertails_refusal_update_statistics.tex}

We train either QLoRA \citep{dettmers2023qloraefficientfinetuningquantized} or
full LoRA \citep{hu2021loralowrankadaptationlarge} adapters with rank $16$, alpha $32$, dropout
$0.05$, learning rate $2\times10^{-4}$, batch size $1$ with $8$ gradient
accumulation steps, maximum sequence length $512$, and one epoch. The adapters
modify the attention and MLP projection modules
(\texttt{q\_proj}, \texttt{k\_proj}, \texttt{v\_proj}, \texttt{o\_proj},
\texttt{gate\_proj}, \texttt{up\_proj}, and \texttt{down\_proj}). We use two
update sizes, 256 and 512 examples, and three update seeds. As quantization-only
controls, we also evaluate the original model loaded with bitsandbytes NF4 or
INT8 quantization. These controls test whether the monitor shift is caused by a
learned update rather than by quantized inference alone.

\subsection{Compute estimates} \label{app:compute}

Our pipeline has three stages---completion generation, anchored-representation
construction, and probe training. Completion generation is the dominant
computational cost and is performed once for each model--dataset pair; cached
activations are then reused across probe fits, source--target applications, and
most ablations. The estimates below characterize representative individual
configurations rather than the total compute of the full experimental suite.
Hardware requirements depend primarily on model size and generation length.

\paragraph{Generation and activation extraction.}
Generation is short for trivia prompts and GuardQA Q\&A prompts, which use at
most $50$ new tokens. GuardQA monitored-model completions use at most $200$ new
tokens, and Guard Distillation uses at most $800$. Online MATH is the most
generation-intensive primary setting, with a maximum of $4{,}096$ new tokens
per solution; probes query the cached trajectories at 64, 128, and 256 tokens.
The related-data and external-target experiments add generation and activation
passes for the new model--dataset pairs, but do not require regenerating an
example separately for every source probe. Guard Distillation additionally
runs Llama Guard once to label each monitored prompt--completion pair.

Llama-3.1-8B-Instruct fits in roughly 16\,GB of GPU memory at
\texttt{float16} or \texttt{bfloat16} precision, plus KV-cache overhead. The
MATH pool and external-target evaluations also include 12B- and 20B-parameter
models, so this 8B estimate does not describe every setting. The 70B
pooled-MMLU scale experiment is an intentionally larger stress test; we ran
its unquantized generation across two GPUs with batch size four.

\paragraph{Anchored-representation construction.}
Once raw activations are stored to disk, the dominant cost of constructing
anchored representations is I/O from loading activations into RAM; the anchor
transform itself is a matrix multiplication.
Concretely, the global anchored representation multiplies an activation matrix
whose effective hidden dimension is $32 \times 4096 = 131{,}072$ (layers
$\times$ hidden size for Llama-3.1-8B-Instruct) against an anchor set. The
activation matrix for $n$ examples requires roughly $4nD$ bytes at
\texttt{float32}, where $D$ is the concatenated hidden dimension. For example,
transforming all $1{,}000$ examples in a Guard Distillation pool with
$D=131{,}072$ uses about $500$\,MB for the activation matrix itself, plus the
$600$-anchor matrix and temporary buffers. The MATH configuration uses 300
anchors. Related-data transfer reuses the source dataset's anchor matrix and
probe, so its additional representation cost comes from encoding and
transforming the target examples rather than refitting every source--target
pair.

\paragraph{Probe training and evaluation.}
Probe fitting is small relative to activation extraction and
anchored-representation construction. Linear probes use
scikit-learn~$1.8.0$ \texttt{LogisticRegression} and
\texttt{roc\_auc\_score}; MLP probes use PyTorch~$2.6.0$. The XSTest
model-update experiments add a separate one-epoch LoRA or QLoRA fine-tuning
stage on 256 or 512 examples. This update cost and the 70B scale stress test
are not representative of fitting an anchored probe itself.

\FloatBarrier

%% file: content/_tables/single_tables/trivia/dataset_statistics/dataset_distribution_table.tex
\begin{table}[htbp]
\centering
\small
\caption{Dataset balance and model answer distributions for the Trivia decision tasks. Ground-truth labels are randomized across answer positions, producing nearly balanced A/B labels for binary tasks and A/B/C/D labels for four-choice tasks. Model outputs are also close to balanced, with few unclear responses; accuracy is averaged over the six monitored models and three seeds.}
\label{tab:trivia-dataset-distribution-summary}
\begin{adjustbox}{max width=\textwidth}
\begin{tabular}{lccc}
\toprule
Dataset & Ground truth & Model outputs & Accuracy \\
\midrule
TruthfulQA & A: 47.5\%, B: 52.5\% & A: 48.6\%, B: 51.0\%, Unclear: 0.5\% & 68.8\% \\
SciQ & A: 49.0\%, B: 51.0\% & A: 50.8\%, B: 48.9\%, Unclear: 0.3\% & 95.8\% \\
MMLU & A: 48.1\%, B: 51.9\% & A: 50.2\%, B: 49.2\%, Unclear: 0.7\% & 87.4\% \\
TruthfulQA (4) & A: 26.2\%, B: 28.2\%, C: 24.2\%, D: 21.4\% & A: 26.5\%, B: 25.8\%, C: 23.1\%, D: 22.3\%, Unclear: 2.3\% & 55.1\% \\
SciQ (4) & A: 25.9\%, B: 25.7\%, C: 24.6\%, D: 23.8\% & A: 26.5\%, B: 25.0\%, C: 24.4\%, D: 23.6\%, Unclear: 0.6\% & 91.4\% \\
MMLU (4) & A: 24.6\%, B: 25.3\%, C: 24.8\%, D: 25.3\% & A: 25.5\%, B: 25.1\%, C: 23.5\%, D: 24.6\%, Unclear: 1.3\% & 79.0\% \\
\bottomrule
\end{tabular}
\end{adjustbox}
\end{table}

%% file: content/_tables/single_tables/guardqa/dataset_statistics/dataset_distribution_table.tex
\begin{table}[htbp]
\centering
\small
\caption{Dataset balance and model answer distributions for the GuardQA decision tasks. Ground-truth labels are randomized across A/B answer positions. Model outputs are close to balanced, with few unclear responses; accuracy is averaged over the six monitored models and three seeds.}
\label{tab:guardqa-dataset-distribution-summary}
\begin{adjustbox}{max width=\textwidth}
\begin{tabular}{lccc}
\toprule
Dataset & Ground truth & Model outputs & Accuracy \\
\midrule
BeaverTails & A: 52.1\%, B: 47.9\% & A: 53.8\%, B: 45.7\%, Unclear: 0.6\% & 73.3\% \\
Toxic-chat & A: 51.6\%, B: 48.4\% & A: 53.7\%, B: 45.7\%, Unclear: 0.6\% & 72.3\% \\
\bottomrule
\end{tabular}
\end{adjustbox}
\end{table}

%% file: content/_tables/single_tables/guard/dataset_statistics/guard_guard_label_balance.tex
\begin{table}[htbp]
\centering
\small
\caption{Realized Llama Guard label balance for the monitored-model completions used in the Guard Distillation setting. Each cell reports mean unsafe / safe percentage over the three paper seeds, with standard error.}
\label{tab:guard-guard-label-balance}
\begin{adjustbox}{max width=\textwidth}
\begin{tabular}{lcc}
\toprule
\textbf{Monitored model} &
\textbf{BeaverTails unsafe / safe (\%)} &
\textbf{Toxic-chat unsafe / safe (\%)} \\
\midrule
\texttt{Llama-3.1-8B-Instruct} & $3.6 \pm 0.1$ / $96.4 \pm 0.1$ & $9.7 \pm 0.2$ / $90.3 \pm 0.2$ \\
\texttt{Llama-3.2-3B-Instruct} & $6.6 \pm 0.2$ / $93.4 \pm 0.2$ & $9.4 \pm 0.2$ / $90.6 \pm 0.2$ \\
\texttt{Mistral-7B-Instruct-v0.3} & $12.2 \pm 0.5$ / $87.8 \pm 0.5$ & $10.7 \pm 0.3$ / $89.3 \pm 0.3$ \\
\texttt{Qwen2.5-7B-Instruct} & $4.6 \pm 0.1$ / $95.4 \pm 0.1$ & $9.3 \pm 0.1$ / $90.7 \pm 0.1$ \\
\texttt{Qwen2.5-3B-Instruct} & $2.0 \pm 0.2$ / $98.0 \pm 0.2$ & $8.7 \pm 0.3$ / $91.3 \pm 0.3$ \\
\texttt{Qwen3-4B-Instruct} & $4.5 \pm 0.1$ / $95.5 \pm 0.1$ & $11.3 \pm 0.3$ / $88.7 \pm 0.3$ \\
\bottomrule
\end{tabular}
\end{adjustbox}
\end{table}

%% file: content/_tables/single_tables/math/dataset_statistics/math_correctness_label_balance.tex
\begin{table}[htbp]
\centering
\small
\caption{Correctness-label balance and checkpoint availability for online mathematical correctness. Each cell gives percentage correct among trajectories reaching 128 generated tokens / percentage of all trajectories reaching that checkpoint. Correctness is model-specific because every monitored model generates its own solution. The MATH source-training row averages the three source/probe seeds; MATH-500 and GSM8K use the frozen evaluation rollouts.}
\label{tab:math-correctness-label-balance}
\begin{adjustbox}{max width=\textwidth}
\begin{tabular}{lcccc}
\toprule
Data & Llama-3.1-8B & Qwen3-4B & OLMo-3-7B & Gemma-3-12B \\
\midrule
MATH source training (2K) & 63.3\% / 95.9\% & 85.4\% / 99.2\% & 81.9\% / 94.1\% & 83.3\% / 99.1\% \\
MATH-500 & 43.4\% / 96.8\% & 85.1\% / 99.2\% & 82.1\% / 94.8\% & 80.6\% / 98.8\% \\
GSM8K & 82.3\% / 86.1\% & 94.4\% / 91.2\% & 93.0\% / 95.5\% & 93.8\% / 95.6\% \\
\bottomrule
\end{tabular}
\end{adjustbox}
\end{table}

%% file: content/_tables/single_tables/guardqa/dataset_statistics/guardqa_guard_label_balance.tex
\begin{table}[htbp]
\centering
\small
\caption{Realized Llama Guard label balance for the monitored-model completions used in GuardQA. Each cell reports mean unsafe / safe percentage over the three paper seeds, with standard error.}
\label{tab:guardqa-guard-label-balance}
\begin{adjustbox}{max width=\textwidth}
\begin{tabular}{lcc}
\toprule
\textbf{Monitored model} &
\textbf{BeaverTails unsafe / safe (\%)} &
\textbf{Toxic-chat unsafe / safe (\%)} \\
\midrule
\texttt{Llama-3.1-8B-Instruct} & $29.8 \pm 0.2$ / $70.2 \pm 0.2$ & $24.6 \pm 0.4$ / $75.4 \pm 0.4$ \\
\texttt{Llama-3.2-3B-Instruct} & $28.4 \pm 0.2$ / $71.6 \pm 0.2$ & $23.4 \pm 0.3$ / $76.6 \pm 0.3$ \\
\texttt{Mistral-7B-Instruct-v0.3} & $24.6 \pm 0.4$ / $75.4 \pm 0.4$ & $25.2 \pm 0.4$ / $74.8 \pm 0.4$ \\
\texttt{Qwen2.5-7B-Instruct} & $12.9 \pm 0.1$ / $87.1 \pm 0.1$ & $18.3 \pm 0.2$ / $81.7 \pm 0.2$ \\
\texttt{Qwen2.5-3B-Instruct} & $6.0 \pm 0.3$ / $94.0 \pm 0.3$ & $16.0 \pm 0.3$ / $84.0 \pm 0.3$ \\
\texttt{Qwen3-4B-Instruct} & $3.2 \pm 0.1$ / $96.8 \pm 0.1$ & $11.5 \pm 0.2$ / $88.5 \pm 0.2$ \\
\bottomrule
\end{tabular}
\end{adjustbox}
\end{table}

%% file: content/_tables/single_tables/staleness_experiments/xstest/dataset_statistics/xstest_data_statistics.tex
\begin{table}[htbp]
\centering
\small
\caption{XSTest data statistics for the probe-staleness setting. We use the full 450-prompt XSTest set from \texttt{Paul/XSTest}; labels are safe versus unsafe prompts, and probes are trained on prompt-only final-token activations.}
\label{tab:xstest-data-statistics}
\begin{adjustbox}{max width=\textwidth}
\begin{tabular}{lc}
\toprule
Statistic & Value \\
\midrule
Total prompts & 450 \\
Safe prompts & 250 (55.6\%) \\
Unsafe prompts & 200 (44.4\%) \\
Train / validation / test split & 315 / 135 / 0 \\
Prompt categories & 18 categories, 25 prompts each \\
\bottomrule
\end{tabular}
\end{adjustbox}
\end{table}

%% file: content/_tables/single_tables/staleness_experiments/xstest/dataset_statistics/beavertails_refusal_update_statistics.tex
\begin{table}[htbp]
\centering
\small
\caption{BeaverTails refusal-SFT update data used in the XSTest probe-staleness learned-update setting. Each update set is balanced between safe examples and unsafe examples rewritten to a refusal target; rows aggregate across update seeds.}
\label{tab:xstest-refusal-update-data-statistics}
\begin{adjustbox}{max width=\textwidth}
\begin{tabular}{lccc}
\toprule
Update & Examples & Safe & Unsafe$\to$refusal \\
\midrule
LoRA & 256 & 128 & 128 \\
LoRA & 512 & 256 & 256 \\
QLoRA & 256 & 128 & 128 \\
QLoRA & 512 & 256 & 256 \\
\bottomrule
\end{tabular}
\end{adjustbox}
\end{table}

%% file: content/appendix/additional_results.tex
\section{Trivia results}
\label{app:trivia-results}

This section reports the detailed trivia results. Same-model decodability,
calibration, validation-selected local depths, and low-FPR recall appear in
Sections~\ref{app:trivia-same-model},
\ref{app:trivia-calibration-progression},
\ref{app:trivia-validation-depth-policy}, and \ref{app:trivia-recall}.
Section~\ref{app:external-target-trivia} reports external-model targets.
Sections~\ref{app:trivia-cross-data-transfer},
\ref{app:trivia-prompt-sensitivity}, \ref{app:trivia-linear-anchor-count}, and
\ref{app:trivia-scale-and-transfer-limits} examine related datasets, prompt
changes, anchor count, model scale, and target performance.
Finally,
Sections~\ref{app:trivia-standard-cross-model-baseline} and
\ref{app:trivia-all-layer-standard-probes} provide standard-probe controls.

\subsection{Same-model decodability}
\label{app:trivia-same-model}

Table~\ref{tab:final_trivia_linear_baseline_linear_concat_auroc} gives the
full same-model Standard-versus-Anchored comparison summarized in
Section~\ref{sec:exp-same-model}. The two representations differ by at most
1.4 AUROC points across the six trivia variants.

\input{content/_tables/single_tables/trivia/linear/non-abs/ls0.8/baseline_auroc.tex}

\subsection{Full calibration progression}
\label{app:trivia-calibration-progression}

Section~\ref{sec:exp-cross-model} reports \texttt{Transfer} (naive transfer) and
\texttt{Transfer~(cal)} (calibrated transfer with $N\!=\!5$) as
representative endpoints. Table~\ref{tab:final_trivia_linear_main_summary_output_auroc}
is the Output Prediction companion to the main GT Prediction table and includes
the same Standard, Anchored, Transfer, and Transfer~(cal) columns. Because
Output Prediction is already near ceiling, we reserve the complete
$N\!=\!1,\dots,5$ source-pool progression for the more informative GT
Prediction target in
Table~\ref{tab:final_trivia_gt-prediction_cross_linear_concat_auroc}, where
$N$ is the number of available source models. Most calibration gains appear
once three source models are available.
TruthfulQA GT Prediction is the exception: calibration adds less than one point
at every pool size because all five source probes fail on the Llama-3.2-3B
target (Section~\ref{sec:exp-cross-model}).

\input{content/_tables/single_tables/trivia/linear/non-abs/ls0.8/main_summary_output_auroc.tex}

\input{content/_tables/single_tables/trivia/linear/non-abs/ls0.8/cross_gt-prediction_auroc.tex}

\subsection{Validation-selected local depths}
\label{app:trivia-validation-depth-policy}

\begin{figure*}[tbp]
\centering
\includegraphics[width=\textwidth]{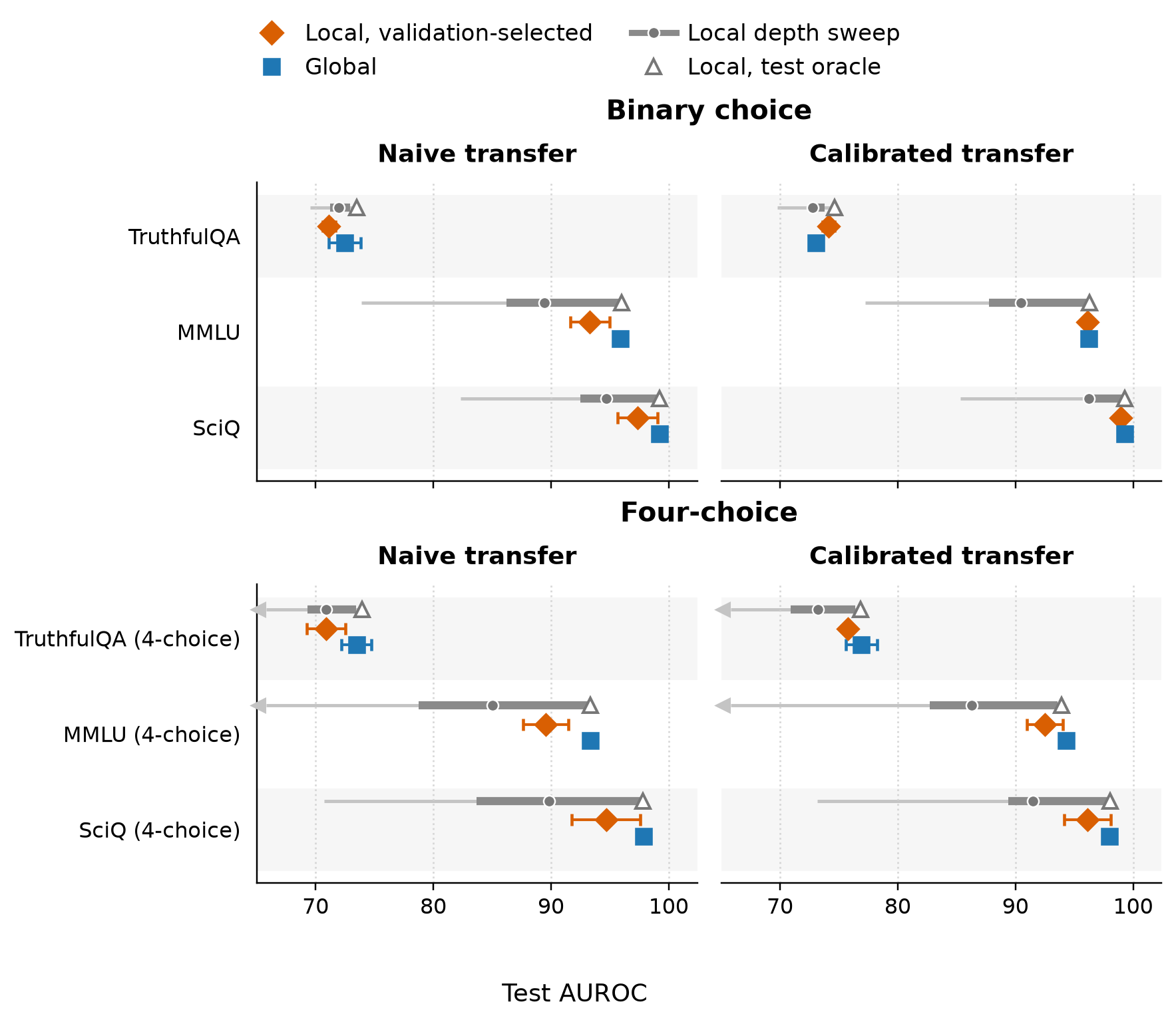}
\caption{\textbf{Global and validation-selected local probes for trivia GT
Prediction.} Panels show direct and calibrated transfer for binary and
four-choice tasks. Orange diamonds show validation-selected local probes; blue
squares show global probes. Gray ranges summarize test AUROC across local
depths, and hollow triangles mark the post-hoc local test oracle. Error bars
are standard errors over three seeds. Left arrows indicate that a local-depth
sweep extends below the displayed range.}
\label{fig:trivia-validation-depth-policy-gt}
\end{figure*}

Section~\ref{sec:exp-global-vs-local} summarizes the layer-selection results.
Figure~\ref{fig:trivia-validation-depth-policy-gt} compares global probes with
local depths selected using validation data for GT Prediction, the informative
trivia target because Output Prediction is saturated in several slices. For naive transfer, each
source model selects its depth using same-model validation. For calibrated
transfer, validation over the source-model pool jointly selects source and
depth for each in-pool target. The gray ranges show test performance across
all candidate local depths. The hollow triangles retain the normal source
policy but select local depth using test data, so they are diagnostic oracles
rather than deployable policies.

On binary TruthfulQA, every incoming source probe is weak for the Llama-3.2-3B target
(Figure~\ref{fig:tqa-gt-confusion}); the small selected-local advantage in the
aggregate does not resolve that target-specific transfer failure.

\FloatBarrier

\FloatBarrier

\subsection{Recall at 10\% FPR}
\label{app:trivia-recall}

Standard-vs-Anchored recall at relative depth~$0.8$ is similar on
Output Prediction and differs by only a few points on GT Prediction, mirroring
the AUROC parity. The calibration summary
(Table~\ref{tab:final_trivia_linear_cross_summary_linear_concat_recall})
shows that anchored recall remains close to Standard within a model but drops
more than AUROC under cross-model transfer. TruthfulQA GT Prediction is the one
case where calibration does not improve recall.

\input{content/_tables/single_tables/trivia/linear/non-abs/ls0.8/cross_summary_recall.tex}

\FloatBarrier

\subsection{Transfer to external model targets}
\label{app:external-target-trivia}

Section~\ref{sec:exp-external-families} summarizes transfer to trivia models
outside the original six-model pool. Table~\ref{tab:external-trivia-gt-full}
expands the external GT Prediction averages into the binary- and four-choice
versions of TruthfulQA, MMLU, and SciQ. Output Prediction remains close to
saturated across the external targets, so we do not reproduce its full
breakdown here.

\input{content/_tables/single_tables/external_targets/external_trivia_full.tex}

\FloatBarrier

\input{content/rebuttal_sections/trivia_cross_data_transfer.tex}

\input{content/rebuttal_sections/trivia_prompt_sensitivity.tex}

\FloatBarrier

\input{content/rebuttal_sections/trivia_linear_anchor_count.tex}

\input{content/rebuttal_sections/scale_and_transfer_limits.tex}

\FloatBarrier

\subsection{Standard cross-model probe baseline}
\label{app:trivia-standard-cross-model-baseline}

Section~\ref{sec:exp-cross-model} argues that standard (non-anchored) probes
cannot in general be applied directly across architectures. Here we test the
same approach empirically for
the one pair in our pool that shares both hidden dimension and layer count:
Llama-3.1-8B-Instruct and Mistral-7B-Instruct-v0.3 (both 4096-dim, 32
transformer layers). For this pair, a standard probe trained on one model can
be applied directly to the other without any dimension mismatch.

For Trivia, standard transfer at relative depth~$0.8$ is weak and inconsistent:
the average across cells is only $48.0$ AUROC, versus $96.6$ for the same-model
standard baseline reported in Section~\ref{sec:exp-same-model}. Under anchored
global linear probes, mean cross-model AUROC rises from $48.0$ to $94.0$.
Output Prediction remains at or
above $95$ AUROC in both directions, showing that the anchored representation
restores transferability that raw coordinates lack for this model pair
(Tables~\ref{tab:trivia-linear-standard-cross-model-auroc}
and~\ref{tab:trivia-linear-anchored-cross-model-auroc}).

\input{content/_tables/single_tables/trivia/linear/standard_cross_model/standard_cross_model_auroc.tex}

\input{content/_tables/single_tables/trivia/linear/standard_cross_model/anchored_cross_model_auroc.tex}

\subsection{All-layer standard probes}
\label{app:trivia-all-layer-standard-probes}

To separate the effect of anchor-relative coordinates from whole-network
concatenation, we train \textit{standard} probes (no anchoring) on the
concatenated all-layer vector and compare them with single-layer standard probes
across relative depths. At the prespecified standard-probe
depth of $0.8$, all-layer standard probes usually match a single-layer standard
probe. This indicates that all-layer concatenation may also, in principle,
provide a strong standard linear probing default. However, the resulting feature
vector is quite large, at $(\texttt{num\_layers} \times \texttt{hidden\_dim})$,
and this may complicate training especially for non-linear probes. The
additional benefit of the global anchored representation is fixing the feature
dimension to the number of anchors.

\begin{figure*}[tbp]
\centering
\includegraphics[width=\textwidth]{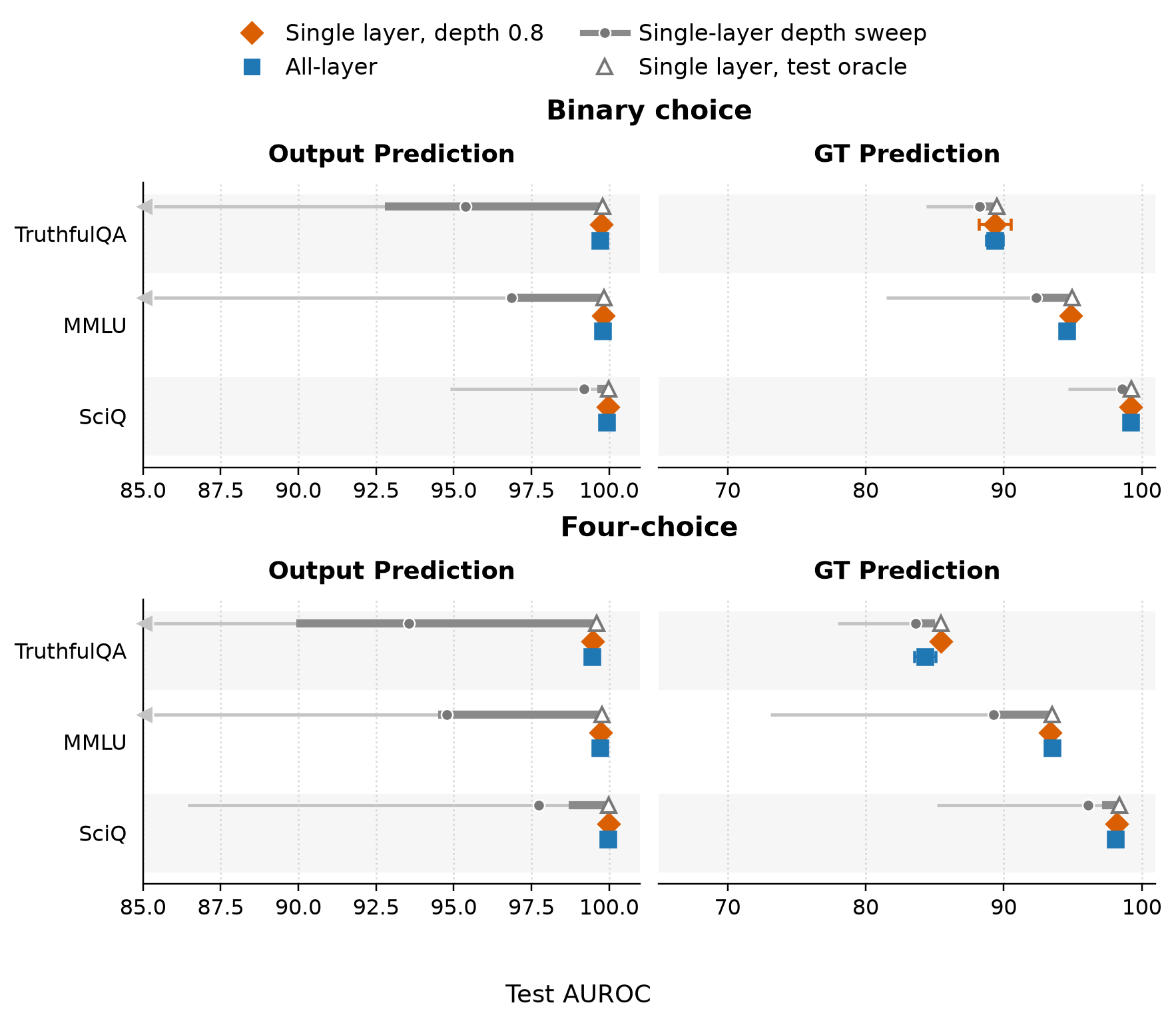}
\caption{\textbf{Standard probes across layers on trivia.} Orange diamonds
show the prespecified single-layer baseline at relative depth~$0.8$; blue
squares show standard probes on concatenated all-layer activations. Gray
segments summarize test AUROC across single-layer depths (range,
interquartile range, and mean), and hollow triangles mark the post-hoc test
oracle. Error bars are standard errors over three seeds.}
\label{fig:app-trivia-abs-depth-sweep}
\end{figure*}

For Trivia linear probes, the all-layer and depth-$0.8$ single-layer standard
probes have similar AUROC
(Tables~\ref{tab:final_trivia_linear_abs_model-choice_auroc}
and~\ref{tab:final_trivia_linear_abs_gt-prediction_auroc}). The depth sweep in
Figure~\ref{fig:app-trivia-abs-depth-sweep} shows that the all-layer standard
probe mainly helps relative to shallow single-layer probes.

\input{content/_tables/single_tables/trivia/linear/abs/ls0.8/abs_model-choice_auroc.tex}

\input{content/_tables/single_tables/trivia/linear/abs/ls0.8/abs_gt-prediction_auroc.tex}

\FloatBarrier

\FloatBarrier

\section{Guard Distillation results}
\label{app:guard-distillation-setting}

This section expands the Guard Distillation results. The calibration
progression and low-FPR recall results appear in
Sections~\ref{app:guard-calibration-progression},
\ref{app:guard-linear-recall}. Section~\ref{app:external-target-guard}
reports external model targets. The related-data transfer results appear in
Section~\ref{app:guard-linear-cross-data-transfer}. Sections~
\ref{app:guard-linear-mean-token-auroc},
\ref{app:guard-mixed-anchor-mitigation}, and
\ref{app:guard-linear-val-plateau} examine anchor-token aggregation, anchor
coverage, and local-depth selection. Probe-design checks cover class balancing
and attention pooling in Sections~
\ref{app:guard-linear-class-balanced-linear} and
\ref{app:guard-linear-attention-diagnostics}. Section~\ref{app:guard-relational-evidence}
reports relational evidence; Section~\ref{app:guard-mlp-additional} evaluates
MLP probes. Sections~\ref{app:guard-linear-standard-cross-model-baseline},
\ref{app:guard-all-layer-standard-probes}, and
\ref{app:guard-model-family-analysis} provide additional cross-model controls.

\input{content/rebuttal_sections/guard_linear_additional_results.tex}

\FloatBarrier

\input{content/rebuttal_sections/guard_relational_evidence.tex}

\FloatBarrier

\input{content/rebuttal_sections/guard_mlp_additional_results.tex}

\FloatBarrier

\input{content/rebuttal_sections/guard_linear_standard_cross_model.tex}

\FloatBarrier

\subsection{All-layer standard probes}
\label{app:guard-all-layer-standard-probes}

To separate the effect of anchor-relative coordinates from whole-network
concatenation, we train \textit{standard} probes (no anchoring) on the
concatenated all-layer vector and compare them with single-layer standard probes
across relative depths.

\begin{figure*}[tbp]
\centering
\includegraphics[width=\textwidth]{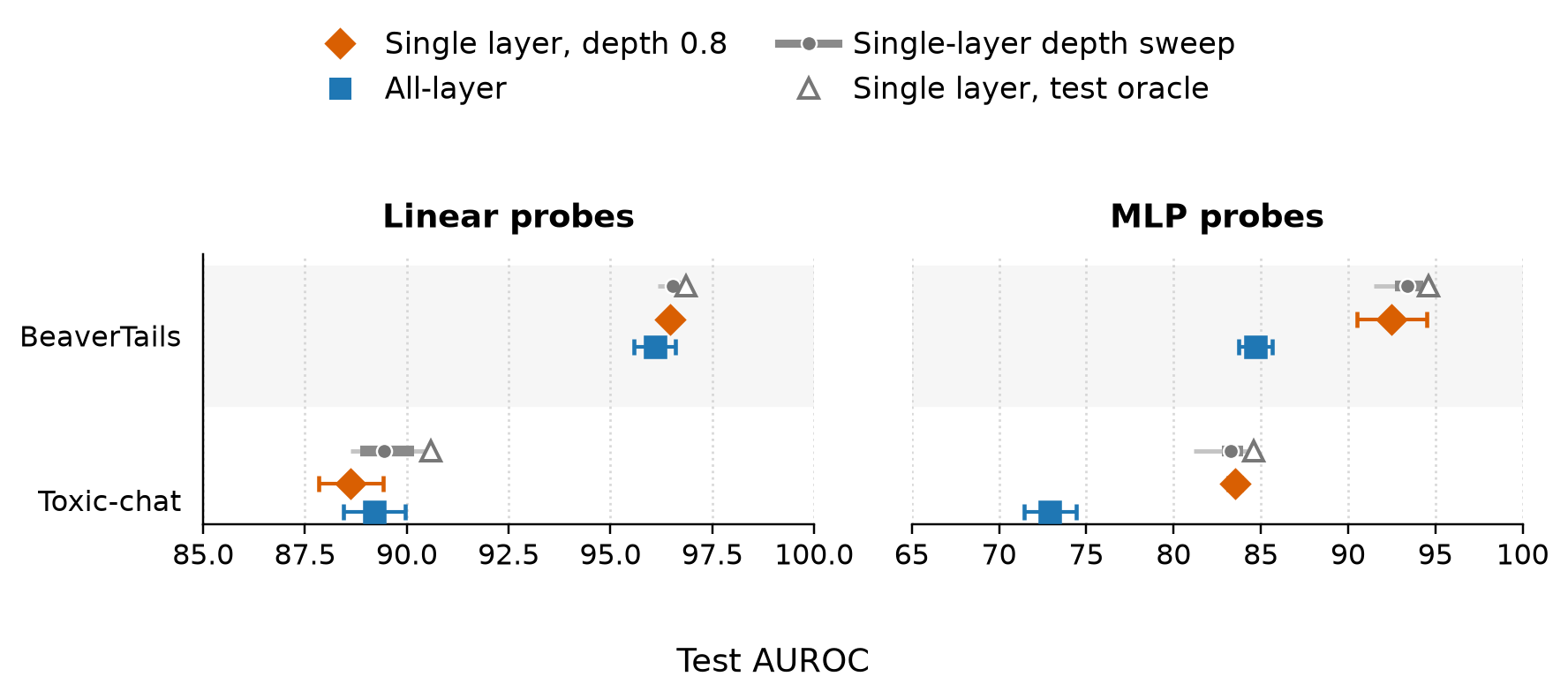}
\caption{\textbf{Standard probes across layers for Guard Distillation.}
Orange diamonds show the prespecified single-layer baseline at relative
depth~$0.8$; blue squares show standard probes on concatenated all-layer
activations. Gray segments summarize test AUROC across single-layer depths
(range, interquartile range, and mean), and hollow triangles mark the post-hoc
test oracle. Panels show linear and MLP-64 probes; error bars are standard
errors over three seeds.}
\label{fig:app-guard-linear-abs-depth-sweep}
\end{figure*}

\paragraph{Guard linear.}
For Guard linear probes, the all-layer standard probe
($C=0.1$) performs close to the single-layer standard probe at depth~$0.8$
(Table~\ref{tab:final_guard_linear_abs_guard-safety_auroc};
Figure~\ref{fig:app-guard-linear-abs-depth-sweep}, left). It is slightly
lower on BeaverTails ($96.1$ vs. $96.5$ AUROC) and slightly higher on
Toxic-chat ($89.2$ vs. $88.6$ AUROC). Thus the Guard linear result matches the
Trivia pattern: all-layer standard probes are competitive with strong
single-layer standard probes.

\input{content/_tables/single_tables/guard/linear/abs/ls0.8/abs_guard-safety_auroc.tex}

\paragraph{Guard MLP.}
For Guard MLP standard probes, the all-layer absolute representation performs
worse than the depth-$0.8$ single-layer representation. The single-layer MLP-64 probe outperforms
the all-layer MLP-64 probe on both Guard datasets ($92.5$ vs. $84.7$ AUROC on
BeaverTails and $83.5$ vs. $72.9$ on Toxic-chat). One possible explanation is
that the higher-dimensional all-layer input needs different regularization, but
this experiment does not isolate the cause.
The global anchored representation has a fixed feature dimension. Its all-layer
result remains competitive with the single-layer probe in the MLP setting
(Table~\ref{tab:final_guard_mlp_abs_fused_guard-safety_auroc};
Figure~\ref{fig:app-guard-linear-abs-depth-sweep}, right).

\input{content/_tables/single_tables/guard/mlp/abs/ls0.8/mlp_abs_guard-safety_auroc.tex}

\FloatBarrier

\FloatBarrier

\subsection{Model family analysis}
\label{app:guard-model-family-analysis}


We next ask whether cross-model transfer is correlated with model family or
size. This diagnostic uses naive single-source \texttt{Transfer}. It exposes the
directed source--target structure that calibration averages over.

For each grouping rule, we partition the six models into groups and average
directed source--target transfer pairs according to whether the source and
target fall in the same group or in different groups. The exact-family split
compares pairs within the same named family (for example, Qwen2.5-to-Qwen2.5)
against all other directed pairs. The coarser split compares Llama+Mistral
against Qwen. The size split compares $7/8$B models against $3/4$B models.
A grouping would be explanatory if the same-group and different-group columns
showed a large, consistent gap. We also report each model's average behavior as
a target and as a source. ``As target'' averages probes trained on all other
models and evaluated on that model. ``As source'' averages that model's probe
evaluated on all other models.

For Guard linear probes, family and size groupings show limited separation.
Exact-family pairs are somewhat higher than different-family pairs ($84.9$ vs.
$80.8$ AUROC). The coarser Llama+Mistral/Qwen split differs by $0.3$ points ($81.2$
vs. $81.5$), and the size split is exactly tied at the reported precision
($81.4$ vs. $81.4$; Table~\ref{tab:app-guard-family-contrasts}). Directional
source--target differences are larger. Qwen2.5-3B is
the weakest target at $71.3$ AUROC, with a $16.7$ point gap to same-model
Anchored. Mistral-7B is the weakest source at $66.4$ AUROC and is a strong
target (Table~\ref{tab:app-guard-source-target-asymmetry}).
Thus the Guard result mirrors the broader conclusion: transfer depends more on
which representation is used as source or target than on a simple family or
size match.

\input{content/_tables/single_tables/guard/linear/model_family/guard_family_contrasts_auroc.tex}

\input{content/_tables/single_tables/guard/linear/model_family/guard_source_target_asymmetry_auroc.tex}

\FloatBarrier

\input{content/appendix/math_results.tex}

\FloatBarrier

\section{XSTest probe staleness}
\label{app:probe-staleness-experiments}

This section first measures how probes trained on the original model behave
after quantization and refusal fine-tuning
(Section~\ref{app:xstest-staleness-main}). It then compares angular distance
with cosine similarity
(Section~\ref{app:xstest-similarity-function}) and whether re-encoding anchors
after the model update improves robustness
(Section~\ref{app:xstest-updated-anchor-encodings}).

\input{content/rebuttal_sections/xstest_staleness_main.tex}
\input{content/rebuttal_sections/xstest_staleness_additional.tex}

\FloatBarrier

\section{GuardQA results}
\label{app:guardqa}

This section reports same-model performance and calibrated transfer in
Sections~\ref{app:guardqa-main-summary} and
\ref{app:guardqa-calibration-progression}, followed by low-FPR recall in
Section~\ref{app:guardqa-recall}. Section~\ref{app:guardqa-mlp-alternates}
tests MLP width, local-depth selection, and anchor count;
Sections~\ref{app:guardqa-linear} and
\ref{app:guardqa-standard-cross-model-baseline} report linear-probe and
standard cross-model controls.

\subsection{Same-model summary}
\label{app:guardqa-main-summary}

GuardQA is an auxiliary safety-evaluation setting that poses a trivia-style A/B
question about the monitored model's own completion. It elicits an explicit
safety judgment at the final prompt position, complementing the completion-token
Guard Distillation monitor defined in Section~\ref{sec:method-task-settings}.
Table~\ref{tab:final_guardqa_mlp_baseline_mlp_1_concat_auroc}
reports the same-model AUROC comparison for the GuardQA MLP setting
($h\!=\!64$, $300$ anchors). Section~\ref{app:guardqa-calibration-progression}
reports cross-model transfer and its full calibration progression.

\input{content/_tables/single_tables/guardqa/mlp/non-abs/anch300/params1_ls0.8/baseline_auroc.tex}

\subsection{Full calibration progression}
\label{app:guardqa-calibration-progression}

Section~\ref{app:guardqa-main-summary} reports \texttt{Transfer} (naive transfer) and
\texttt{Transfer~(cal)} (calibrated transfer with $N\!=\!5$) as
representative endpoints. Tables~\ref{tab:final_guardqa_model-choice_cross_mlp_1_concat_auroc}
and~\ref{tab:final_guardqa_gt-prediction_cross_mlp_1_concat_auroc} report the
complete $N\!=\!1,\dots,5$ progression alongside the Anchored same-model column
for the GuardQA MLP setting. GuardQA BeaverTails Output Prediction continues to
improve from $84.0$ at $N\!=\!4$ to $89.0$ at $N\!=\!5$.

\input{content/_tables/single_tables/guardqa/mlp/non-abs/anch300/params1_ls0.8/cross_model-choice_auroc.tex}

\input{content/_tables/single_tables/guardqa/mlp/non-abs/anch300/params1_ls0.8/cross_gt-prediction_auroc.tex}

\subsection{Recall at 10\% FPR}
\label{app:guardqa-recall}

Same-model Standard-vs-Anchored recall parity holds for the GuardQA
MLP setting ($h\!=\!64$, $300$ anchors;
Table~\ref{tab:final_guardqa_mlp_cross_summary_mlp_1_concat_recall}).
Cross-transfer recall gaps are
larger than AUROC gaps, particularly on BeaverTails Output Prediction. Recall
drops from $99.5$ to $61.6$ under naive transfer and recovers to $76.0$ with
calibration. AUROC drops from $99.8$ to $72.3$ and recovers to $89.0$. The
AUROC and recall results therefore show the same ordering, with larger
cross-model losses at low FPR.

\input{content/_tables/single_tables/guardqa/mlp/non-abs/anch300/params1_ls0.8/cross_summary_recall.tex}

\FloatBarrier

\subsection{Alternate MLP settings}
\label{app:guardqa-mlp-alternates}
\label{app:anchor-count-sweep}

The GuardQA results above use an MLP with hidden size $64$ and $300$ anchors.
This subsection tests whether same-model parity, calibrated transfer, and depth
sensitivity persist under other MLP configurations.

\paragraph{\texorpdfstring{Hidden size $h\!=\!256$}{Hidden size h=256}}
\label{app:mlp256}

At $h\!=\!256$ ($300$ anchors), the main cross-transfer pattern matches that
at $h\!=\!64$
(Table~\ref{tab:final_guardqa_mlp_cross_summary_mlp_3_concat_auroc}).
Baseline Standard-vs-Anchored numbers differ by less than $1$ AUROC
point from the $h\!=\!64$ GuardQA baseline
(Table~\ref{tab:final_guardqa_mlp_baseline_mlp_1_concat_auroc}); we
omit them here. Same-model Anchored AUROC on GT Prediction is within $0.2$
points of $h\!=\!64$ (marginally lower on Toxic-chat). Calibrated
cross-transfer on BeaverTails Output Prediction is slightly worse
($85.2$ vs.\ $89.0$ for $h\!=\!64$); the remaining cells match within
noise.

For $h\!=\!256$, a local depth selected using target-model test data can exceed
the global result on BeaverTails Output Prediction. The local probe at
depth~$0.85$ reaches AUROC~$89.5$ under \texttt{Transfer~(cal)}, compared with
$85.2$ for the global probe. This is an oracle comparison. The peak at $0.85$
is isolated. Transfer~(cal) drops to $77.5$ at depth $0.80$ and to $74.6$ at
$0.90$. Table~\ref{tab:guardqa-source-depth-selection-strategy} places this
oracle result alongside validation-selected depths.

\input{content/_tables/single_tables/guardqa/mlp/non-abs/anch300/params3_ls0.8/cross_summary_auroc.tex}

\paragraph{Validation-selected local-depth strategies}
\label{app:h64-val-plateau}

Table~\ref{tab:guardqa-source-depth-selection-strategy} compares the same
validation strategies defined in Appendix~\ref{app:guard-linear-val-plateau}
against the global anchored representation for Output Prediction, for both
$h\!=\!64$ and $h\!=\!256$ MLP probes. The Validation column uses per-model
validation for same-model Anchored, per-source validation for naive Transfer,
and source-pool validation for Transfer~(cal). The Optimal column is selected
post-hoc over the local-depth grid and uses held-out test performance.

The same-model Anchored rows change little under this choice: validation-selected
local depths are close to global. For naive Transfer, source-local
validation does not reliably help and is often below both fixed depth $0.8$ and
global. For Transfer~(cal), source-depth validation is mixed. It improves over
fixed depth $0.8$ for the $h\!=\!256$ BeaverTails Output slice, but hurts the
$h\!=\!64$ Output Prediction slices and slightly trails fixed depth $0.8$ on
$h\!=\!256$ Toxic-chat Output. The Optimal column confirms that local probes can
outperform global in several slices if the depth is chosen with target test
labels, but the deployable Validation selector does not consistently recover
those oracle depths. Thus the GuardQA analysis also supports using global
anchored probes as a robust depth-free default unless transfer-depth oracle
tuning is available.

\input{content/_tables/single_tables/guardqa/mlp/validation_depth/source_depth_selection_strategy_table.tex}

\paragraph{Anchor count variants}
\label{app:anchor-count}

Table~\ref{tab:final_guardqa_mlp_anchor_sensitivity_auroc} reports
calibrated cross-model transfer (Transfer~(cal)) as the anchor set size
grows from 300 to 600, for both the MLP-64 and MLP-256 probes.
Within-model performance (Standard, Anchored) is virtually unchanged
across all three counts and is therefore omitted.

Cross-model transfer shows a mixed, dataset-dependent pattern.
On Toxic-chat~QA the effect is mildly positive. Output~Prediction AUROC
rises from $95.8$ to $98.0$ (MLP-64) and from $96.4$ to $97.7$ (MLP-256),
while GT~Prediction is flat.
On BeaverTails~QA, Output~Prediction degrades
sharply for MLP-64 ($89.0 \to 73.4$) and moderately for MLP-256
($85.2 \to 80.7$). GT~Prediction improves from $84.0$ to $88.1$ for MLP-64
and from $84.3$ to $87.0$ for MLP-256.

Three hundred anchors suffice for same-model parity and substantial calibrated
transfer in the evaluated GuardQA configurations. Additional anchors do not
yield a consistent improvement and can reduce performance, particularly for
BeaverTails~QA Output~Prediction.

\input{content/_tables/single_tables/guardqa/mlp/anchor_sensitivity/anchor_sensitivity_auroc.tex}

\subsection{Linear probe setting}
\label{app:guardqa-linear}

For comparison with the MLP results above, we report linear logistic-regression
probes on the same GuardQA data. Linear probes leave more headroom on GT
Prediction, motivating the MLP choice for the preceding GuardQA analyses.

Tables~\ref{tab:final_guardqa_linear_cross_summary_linear_concat_auroc}
and~\ref{tab:final_guardqa_linear_cross_summary_linear_concat_recall}
show the Anchored / Transfer / Transfer~(cal) summary for AUROC and
recall respectively. Standard-vs-Anchored parity holds (Anchored
slightly ahead on GT Prediction for both datasets). Its calibration
progression follows the $h\!=\!64$ MLP pattern: source selection helps, with
the exact saturation point varying by cell. Recall
cross-transfer gaps are larger than AUROC gaps,
consistent with the MLP recall pattern.

\input{content/_tables/single_tables/guardqa/linear/non-abs/ls0.8/cross_summary_auroc.tex}

\input{content/_tables/single_tables/guardqa/linear/non-abs/ls0.8/cross_summary_recall.tex}

\FloatBarrier

\subsection{Standard cross-model probe baseline}
\label{app:guardqa-standard-cross-model-baseline}

We repeat the raw-coordinate transfer test between Llama-3.1-8B and Mistral-7B
for GuardQA linear probes. Standard transfer is higher than in Trivia but still unstable, averaging
$63.9$ AUROC across datasets, probe targets, directions, and seeds; GT
Prediction remains far below the anchored alternative. Under anchored global
linear probes, mean cross-model AUROC rises from $63.9$ to $84.3$. The
comparison confirms that the anchored representation restores cross-model
transferability that raw coordinates lack
(Tables~\ref{tab:guardqa-linear-standard-cross-model-auroc}
and~\ref{tab:guardqa-linear-anchored-cross-model-auroc}).

\input{content/_tables/single_tables/guardqa/linear/standard_cross_model/standard_cross_model_auroc.tex}

\input{content/_tables/single_tables/guardqa/linear/standard_cross_model/anchored_cross_model_auroc.tex}

%% file: content/_tables/single_tables/trivia/linear/non-abs/ls0.8/baseline_auroc.tex
\begin{table}[htbp]
\centering
\caption{Linear-probe AUROC on trivia for Output Prediction and GT Prediction: same-model Standard and global Anchored probes.}
\label{tab:final_trivia_linear_baseline_linear_concat_auroc}
\begin{adjustbox}{max width=\textwidth}
\begin{tabular}{lcccc}
\toprule
 & \multicolumn{2}{c}{Output Prediction} & \multicolumn{2}{c}{GT Prediction} \\
\cmidrule(lr){2-3} \cmidrule(lr){4-5}
Dataset & Standard & Anchored & Standard & Anchored \\
\midrule
TQA & $99.7\pm0.0$ & $99.8\pm0.0$ & $89.4\pm1.2$ & $89.1\pm0.4$ \\
MMLU & $99.8\pm0.0$ & $99.8\pm0.0$ & $94.8\pm0.2$ & $96.2\pm0.2$ \\
SciQ & $100.0\pm0.0$ & $100.0\pm0.0$ & $99.2\pm0.1$ & $99.4\pm0.0$ \\
TQA (4) & $99.5\pm0.1$ & $99.6\pm0.0$ & $85.5\pm0.3$ & $84.1\pm1.0$ \\
MMLU (4) & $99.7\pm0.0$ & $99.8\pm0.0$ & $93.4\pm0.2$ & $94.8\pm0.1$ \\
SciQ (4) & $100.0\pm0.0$ & $100.0\pm0.0$ & $98.2\pm0.3$ & $98.6\pm0.3$ \\
\bottomrule
\end{tabular}
\end{adjustbox}
\end{table}

%% file: content/_tables/single_tables/trivia/linear/non-abs/ls0.8/main_summary_output_auroc.tex
\begin{table}[htbp]
\centering
\caption{\textbf{Trivia output prediction.} Output Prediction AUROC for same-model Standard and Anchored probes, mean cross-model Transfer, and source-calibrated Transfer (cal). Values are percentages, averaged over three seeds; uncertainties are standard errors.}
\label{tab:final_trivia_linear_main_summary_output_auroc}
\begin{adjustbox}{max width=\textwidth}
\begin{tabular}{lcccc}
\toprule
Dataset & Standard & Anchored & Transfer & Transfer (cal) \\
\midrule
TQA & $99.7\pm0.0$ & $99.8\pm0.0$ & $97.0\pm0.7$ & $98.3\pm1.3$ \\
MMLU & $99.8\pm0.0$ & $99.8\pm0.0$ & $99.8\pm0.1$ & $99.8\pm0.1$ \\
SciQ & $100.0\pm0.0$ & $100.0\pm0.0$ & $100.0\pm0.0$ & $99.9\pm0.0$ \\
TQA (4) & $99.5\pm0.1$ & $99.6\pm0.0$ & $97.8\pm0.3$ & $99.0\pm0.2$ \\
MMLU (4) & $99.7\pm0.0$ & $99.8\pm0.0$ & $99.2\pm0.1$ & $99.5\pm0.0$ \\
SciQ (4) & $100.0\pm0.0$ & $100.0\pm0.0$ & $99.9\pm0.0$ & $99.9\pm0.0$ \\
\bottomrule
\end{tabular}
\end{adjustbox}
\end{table}

%% file: content/_tables/single_tables/trivia/linear/non-abs/ls0.8/cross_gt-prediction_auroc.tex
\begin{table}[htbp]
\centering
\caption{Linear-probe AUROC on trivia for GT Prediction: same-model Anchored and cross-model Transfer as the available source pool grows. Transfer (cal-N) uses source-only validation within each N-model pool.}
\label{tab:final_trivia_gt-prediction_cross_linear_concat_auroc}
\begin{adjustbox}{max width=\textwidth}
\begin{tabular}{lcccccc}
\toprule
Dataset & Anchored & Transfer & Transfer (cal-2) & Transfer (cal-3) & Transfer (cal-4) & Transfer (cal) \\
\midrule
TQA & $89.1\pm0.4$ & $72.5\pm1.4$ & $72.5\pm1.0$ & $72.3\pm0.8$ & $72.4\pm0.8$ & $\mathbf{73.1}\pm0.5$ \\
MMLU & $96.2\pm0.2$ & $95.9\pm0.3$ & $96.0\pm0.3$ & $96.1\pm0.3$ & $\mathbf{96.2}\pm0.2$ & $\mathbf{96.2}\pm0.2$ \\
SciQ & $99.4\pm0.0$ & $99.2\pm0.1$ & $\mathbf{99.3}\pm0.1$ & $\mathbf{99.3}\pm0.1$ & $\mathbf{99.3}\pm0.1$ & $\mathbf{99.3}\pm0.0$ \\
TQA (4) & $84.1\pm1.0$ & $73.5\pm1.3$ & $75.7\pm1.1$ & $76.7\pm1.1$ & $\mathbf{77.1}\pm1.2$ & $76.9\pm1.3$ \\
MMLU (4) & $94.8\pm0.1$ & $93.3\pm0.1$ & $93.8\pm0.2$ & $93.9\pm0.1$ & $94.1\pm0.1$ & $\mathbf{94.3}\pm0.2$ \\
SciQ (4) & $98.6\pm0.3$ & $97.9\pm0.2$ & $98.0\pm0.3$ & $\mathbf{98.1}\pm0.3$ & $98.0\pm0.3$ & $98.0\pm0.3$ \\
\bottomrule
\end{tabular}
\end{adjustbox}
\end{table}

%% file: content/_tables/single_tables/trivia/linear/non-abs/ls0.8/cross_summary_recall.tex
\begin{table}[htbp]
\centering
\caption{Linear-probe R@0.1FPR on trivia for Output Prediction and GT Prediction: same-model Anchored, mean cross-model Transfer, and source-calibrated Transfer (cal).}
\label{tab:final_trivia_linear_cross_summary_linear_concat_recall}
\begin{adjustbox}{max width=\textwidth}
\begin{tabular}{lcccccc}
\toprule
 & \multicolumn{3}{c}{Output Prediction} & \multicolumn{3}{c}{GT Prediction} \\
\cmidrule(lr){2-4} \cmidrule(lr){5-7}
Dataset & Anchored & Transfer & Transfer (cal) & Anchored & Transfer & Transfer (cal) \\
\midrule
TQA & $99.8\pm0.1$ & $94.8\pm1.0$ & $95.4\pm3.8$ & $69.4\pm2.5$ & $37.4\pm0.9$ & $34.0\pm1.2$ \\
MMLU & $99.6\pm0.1$ & $99.4\pm0.1$ & $99.6\pm0.1$ & $86.9\pm1.4$ & $87.5\pm1.3$ & $87.9\pm0.9$ \\
SciQ & $100.0\pm0.0$ & $100.0\pm0.0$ & $100.0\pm0.0$ & $98.7\pm0.2$ & $98.4\pm0.2$ & $98.4\pm0.2$ \\
\bottomrule
\end{tabular}
\end{adjustbox}
\end{table}

%% file: content/_tables/single_tables/external_targets/external_trivia_full.tex
\begin{table}[htbp]
\centering
\caption{\textbf{External-target trivia: GT Prediction.} Each row reports mean AUROC $\pm$ standard error over three source/probe seeds.}
\label{tab:external-trivia-gt-full}
\begin{tabular*}{\textwidth}{@{\extracolsep{\fill}}llcc@{}}
\toprule
Dataset / format & Target & Target-trained & Global transfer \\
\midrule
TruthfulQA & OLMo-3-7B & 88.6 $\pm$ 1.0 & 80.7 $\pm$ 0.6 \\
TruthfulQA & GPT-OSS-20B & 93.0 $\pm$ 0.8 & 85.6 $\pm$ 1.4 \\
TruthfulQA & Gemma-3-12B & 92.4 $\pm$ 1.2 & 86.5 $\pm$ 1.2 \\
\addlinespace
MMLU & OLMo-3-7B & 93.0 $\pm$ 1.1 & 93.1 $\pm$ 1.3 \\
MMLU & GPT-OSS-20B & 97.0 $\pm$ 0.3 & 97.6 $\pm$ 0.4 \\
MMLU & Gemma-3-12B & 97.7 $\pm$ 0.3 & 97.9 $\pm$ 0.3 \\
\addlinespace
SciQ & OLMo-3-7B & 99.3 $\pm$ 0.1 & 99.1 $\pm$ 0.3 \\
SciQ & GPT-OSS-20B & 99.8 $\pm$ 0.1 & 99.7 $\pm$ 0.1 \\
SciQ & Gemma-3-12B & 99.5 $\pm$ 0.1 & 99.6 $\pm$ 0.1 \\
\addlinespace
TruthfulQA (four-choice) & OLMo-3-7B & 85.2 $\pm$ 0.9 & 81.2 $\pm$ 1.0 \\
TruthfulQA (four-choice) & GPT-OSS-20B & 89.7 $\pm$ 1.2 & 88.1 $\pm$ 0.9 \\
TruthfulQA (four-choice) & Gemma-3-12B & 88.4 $\pm$ 1.1 & 84.0 $\pm$ 1.5 \\
\addlinespace
MMLU (four-choice) & OLMo-3-7B & 92.0 $\pm$ 0.9 & 90.1 $\pm$ 1.7 \\
MMLU (four-choice) & GPT-OSS-20B & 96.5 $\pm$ 0.2 & 96.4 $\pm$ 0.4 \\
MMLU (four-choice) & Gemma-3-12B & 96.6 $\pm$ 0.8 & 95.8 $\pm$ 0.1 \\
\addlinespace
SciQ (four-choice) & OLMo-3-7B & 97.3 $\pm$ 0.6 & 97.3 $\pm$ 0.8 \\
SciQ (four-choice) & GPT-OSS-20B & 99.1 $\pm$ 0.1 & 98.8 $\pm$ 0.2 \\
SciQ (four-choice) & Gemma-3-12B & 98.6 $\pm$ 0.4 & 97.6 $\pm$ 0.7 \\
\bottomrule
\end{tabular*}
\end{table}

%% file: content/rebuttal_sections/trivia_cross_data_transfer.tex
\subsection{Related-data and joint transfer}
\label{app:trivia-cross-data-transfer}

Section~\ref{sec:exp-related-data} summarizes transfer among the trivia
datasets. Here we test every source--target dataset direction with the same
answer format under the common protocol in
Appendix~\ref{app:related-data-implementation}.
Table~\ref{tab:trivia-linear-cross-data-transfer-auroc}
reports same-model transfer. Output Prediction is saturated at least $99.1$
AUROC in every direction. For GT Prediction, Global anchored probes outperform
Standard in all displayed directions. The largest gains occur when TruthfulQA
is the source: $74.9\rightarrow88.0$ on binary SciQ,
$69.9\rightarrow81.6$ on binary MMLU, $85.9\rightarrow95.5$ on four-choice
SciQ, and $78.1\rightarrow88.6$ on four-choice MMLU. SciQ--MMLU transfer is
within half a point of target same-model performance, while TruthfulQA remains
the difficult target distribution.

\input{content/_tables/single_tables/trivia/linear/cross_data_transfer/trivia_linear_cross_data_transfer_auroc.tex}

We next change both dataset and monitored model
(Table~\ref{tab:trivia-linear-cross-data-xmodel-transfer}). Global GT
Prediction averages $83.1$ under naive transfer and $86.1$ under source
calibration, compared with $87.0$ for same-model related-data transfer.
Calibration therefore recovers most of the additional model-transfer penalty,
although directions with TruthfulQA as source or target retain the largest
gaps.

\input{content/_tables/single_tables/trivia/linear/cross_data_transfer/trivia_linear_cross_data_xmodel_transfer_single_anchor.tex}

%% file: content/_tables/single_tables/trivia/linear/cross_data_transfer/trivia_linear_cross_data_transfer_auroc.tex
\begin{table}[htbp]
\centering
\small
\caption{Same-model related-data trivia transfer (AUROC). Probes are trained on one trivia dataset, either binary or four-choice, and evaluated on another for the same monitored model. Standard and Local use relative depth 0.8; Global uses all layers. Values are mean test performance (\%) $\pm$ standard error over three seeds and six models.}
\label{tab:trivia-linear-cross-data-transfer-auroc}
\begin{adjustbox}{max width=\textwidth}
\begin{tabular}{lcccccc}
\toprule
 & \multicolumn{3}{c}{\textbf{Output Prediction}} & \multicolumn{3}{c}{\textbf{GT Prediction}} \\
\cmidrule(lr){2-4} \cmidrule(lr){5-7}
\textbf{Transfer direction} & \textbf{Standard} & \textbf{Global anchored} & \textbf{Local anchored} & \textbf{Standard} & \textbf{Global anchored} & \textbf{Local anchored} \\
\midrule
\multicolumn{7}{l}{\textit{Binary}} \\
TQA $\rightarrow$ SciQ & $100.0\pm0.0$ & $100.0\pm0.0$ & $100.0\pm0.0$ & $74.9\pm3.6$ & $88.0\pm3.4$ & $85.3\pm3.6$ \\
SciQ $\rightarrow$ TQA & $99.8\pm0.1$ & $99.8\pm0.1$ & $99.8\pm0.1$ & $73.2\pm1.9$ & $74.0\pm2.1$ & $73.5\pm2.0$ \\
TQA $\rightarrow$ MMLU & $99.8\pm0.1$ & $99.8\pm0.1$ & $99.8\pm0.1$ & $69.9\pm2.7$ & $81.6\pm3.3$ & $79.4\pm3.3$ \\
MMLU $\rightarrow$ TQA & $99.7\pm0.1$ & $99.8\pm0.1$ & $99.7\pm0.1$ & $73.0\pm2.0$ & $75.4\pm2.1$ & $75.1\pm2.1$ \\
SciQ $\rightarrow$ MMLU & $99.8\pm0.1$ & $99.8\pm0.1$ & $99.8\pm0.1$ & $95.2\pm0.4$ & $96.3\pm0.4$ & $96.2\pm0.4$ \\
MMLU $\rightarrow$ SciQ & $100.0\pm0.0$ & $100.0\pm0.0$ & $100.0\pm0.0$ & $98.8\pm0.3$ & $99.3\pm0.1$ & $99.3\pm0.2$ \\
\multicolumn{7}{l}{\textit{Four-choice}} \\
TQA (4) $\rightarrow$ SciQ (4) & $100.0\pm0.0$ & $100.0\pm0.0$ & $100.0\pm0.0$ & $85.9\pm2.2$ & $95.5\pm1.1$ & $94.6\pm1.2$ \\
SciQ (4) $\rightarrow$ TQA (4) & $99.3\pm0.2$ & $99.4\pm0.2$ & $99.1\pm0.2$ & $73.6\pm1.1$ & $74.9\pm1.4$ & $74.0\pm1.2$ \\
TQA (4) $\rightarrow$ MMLU (4) & $99.6\pm0.1$ & $99.7\pm0.1$ & $99.7\pm0.1$ & $78.1\pm2.2$ & $88.6\pm1.5$ & $87.1\pm1.6$ \\
MMLU (4) $\rightarrow$ TQA (4) & $99.2\pm0.2$ & $99.5\pm0.2$ & $99.2\pm0.2$ & $74.6\pm1.2$ & $77.3\pm1.4$ & $76.3\pm1.3$ \\
SciQ (4) $\rightarrow$ MMLU (4) & $99.7\pm0.1$ & $99.7\pm0.1$ & $99.7\pm0.1$ & $93.2\pm0.4$ & $94.3\pm0.4$ & $93.9\pm0.4$ \\
MMLU (4) $\rightarrow$ SciQ (4) & $100.0\pm0.0$ & $100.0\pm0.0$ & $100.0\pm0.0$ & $97.9\pm0.4$ & $98.4\pm0.4$ & $98.4\pm0.4$ \\
\bottomrule
\end{tabular}
\end{adjustbox}
\end{table}

%% file: content/_tables/single_tables/trivia/linear/cross_data_transfer/trivia_linear_cross_data_xmodel_transfer_single_anchor.tex
\begin{table}[htbp]
\centering
\small
\caption{Joint related-data and cross-model trivia transfer for GT Prediction. Transfer (cal) uses source-only validation. Values are mean test AUROC (\%) $\pm$ standard error over seeds and target models.}
\label{tab:trivia-linear-cross-data-xmodel-transfer}
\begin{adjustbox}{max width=\textwidth}
\begin{tabular}{lrrrr}
\toprule
Direction & \multicolumn{2}{c}{Global probe} & \multicolumn{2}{c}{Local probe} \\
\cmidrule(lr){2-3}\cmidrule(lr){4-5}
& Transfer & Transfer (cal) & Transfer & Transfer (cal) \\
\midrule
TQA $\rightarrow$ SciQ & 75.6 $\pm$ 2.3 & 82.6 $\pm$ 0.1 & 76.8 $\pm$ 2.8 & 82.6 $\pm$ 0.1 \\
SciQ $\rightarrow$ TQA & 74.5 $\pm$ 1.0 & 74.3 $\pm$ 1.3 & 74.0 $\pm$ 1.0 & 74.0 $\pm$ 1.0 \\
TQA $\rightarrow$ MMLU & 74.5 $\pm$ 2.7 & 80.6 $\pm$ 0.1 & 75.3 $\pm$ 2.9 & 80.8 $\pm$ 0.1 \\
MMLU $\rightarrow$ TQA & 74.6 $\pm$ 1.0 & 75.0 $\pm$ 0.8 & 74.2 $\pm$ 1.0 & 74.5 $\pm$ 0.7 \\
SciQ $\rightarrow$ MMLU & 95.9 $\pm$ 0.3 & 96.1 $\pm$ 0.4 & 95.8 $\pm$ 0.3 & 95.9 $\pm$ 0.3 \\
MMLU $\rightarrow$ SciQ & 99.2 $\pm$ 0.1 & 99.3 $\pm$ 0.1 & 99.2 $\pm$ 0.0 & 99.3 $\pm$ 0.1 \\
TQA (4) $\rightarrow$ SciQ (4) & 81.5 $\pm$ 3.3 & 90.9 $\pm$ 1.9 & 83.4 $\pm$ 2.9 & 94.1 $\pm$ 1.1 \\
SciQ (4) $\rightarrow$ TQA (4) & 75.5 $\pm$ 0.4 & 76.1 $\pm$ 0.4 & 74.7 $\pm$ 0.5 & 75.4 $\pm$ 0.4 \\
TQA (4) $\rightarrow$ MMLU (4) & 78.8 $\pm$ 2.6 & 88.4 $\pm$ 0.6 & 80.2 $\pm$ 2.2 & 89.4 $\pm$ 0.7 \\
MMLU (4) $\rightarrow$ TQA (4) & 76.1 $\pm$ 0.3 & 77.1 $\pm$ 0.4 & 75.3 $\pm$ 0.4 & 75.7 $\pm$ 0.4 \\
SciQ (4) $\rightarrow$ MMLU (4) & 93.8 $\pm$ 0.2 & 94.5 $\pm$ 0.3 & 93.3 $\pm$ 0.1 & 94.2 $\pm$ 0.1 \\
MMLU (4) $\rightarrow$ SciQ (4) & 97.7 $\pm$ 0.5 & 98.1 $\pm$ 0.3 & 97.7 $\pm$ 0.4 & 97.9 $\pm$ 0.3 \\
\midrule
Overall & 83.1 $\pm$ 1.7 & 86.1 $\pm$ 1.6 & 83.3 $\pm$ 1.7 & 86.1 $\pm$ 1.6 \\
\bottomrule
\end{tabular}
\end{adjustbox}
\end{table}

%% file: content/rebuttal_sections/trivia_prompt_sensitivity.tex
\subsection{Sensitivity to prompt perturbations}
\label{app:trivia-prompt-sensitivity}

\begin{figure*}[tbp]
\centering
\includegraphics[width=0.8\textwidth]{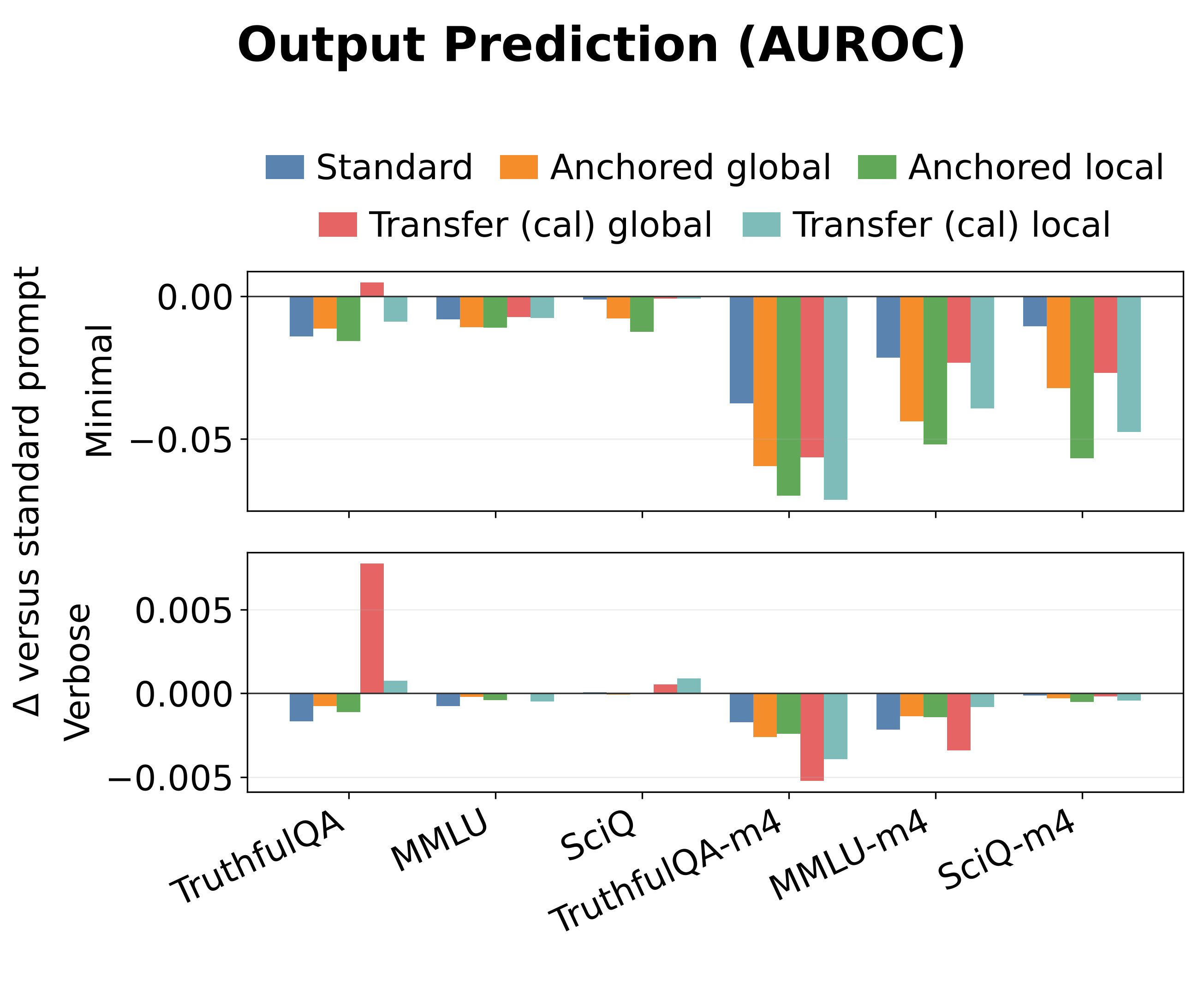}\\[4pt]
\includegraphics[width=0.8\textwidth]{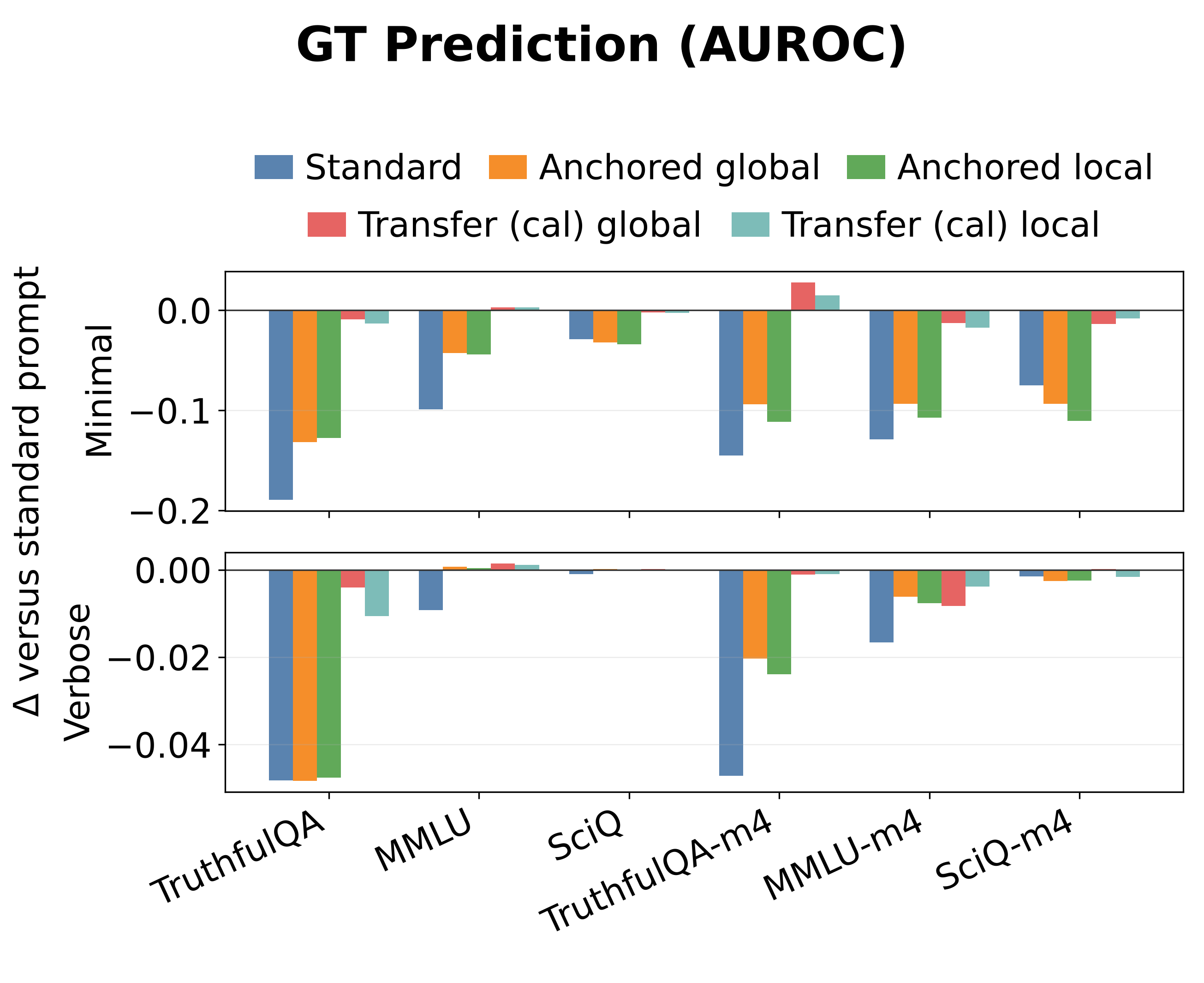}
\caption{Prompt sensitivity on the trivia datasets, measured by AUROC. Top:
Output Prediction; bottom:
GT Prediction. Bars show the shifted-prompt score minus the same method's
standard-prompt score. Each panel compares Standard, same-model Anchored
(global/local), and calibrated transfer (global/local).}
\label{fig:trivia-prompt-sensitivity-auroc}
\end{figure*}

\begin{figure*}[tbp]
\centering
\includegraphics[width=0.8\textwidth]{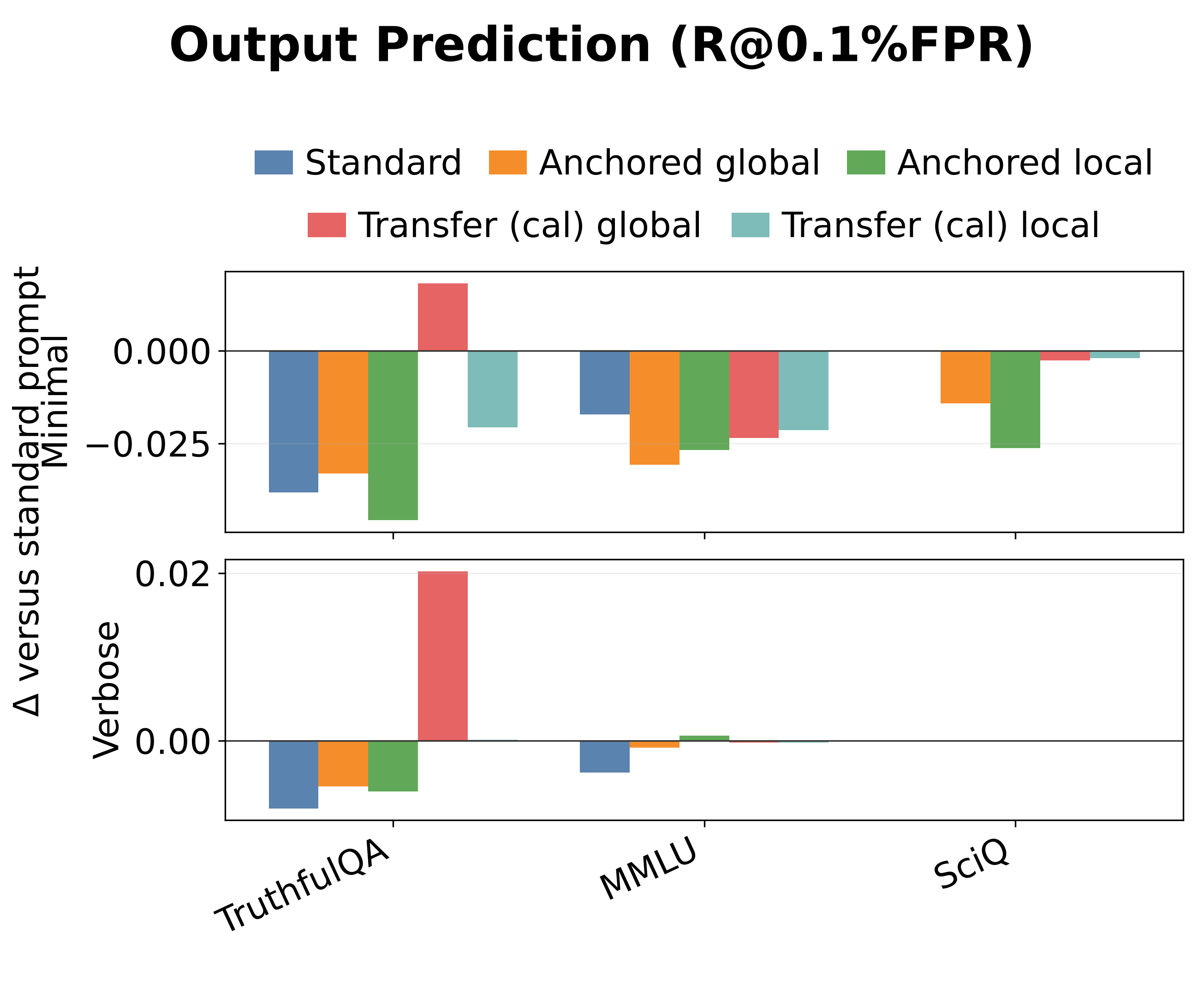}\\[4pt]
\includegraphics[width=0.8\textwidth]{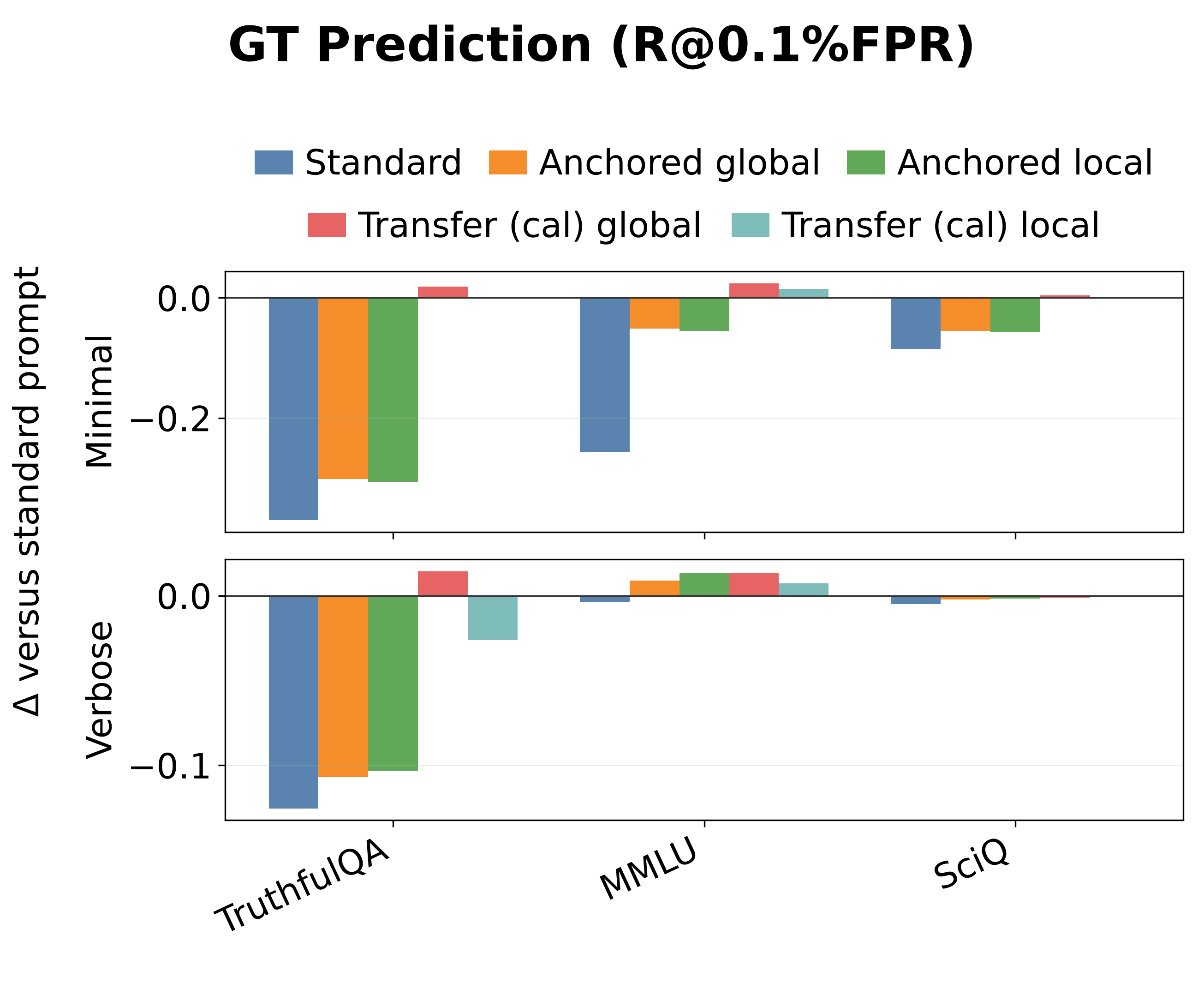}
\caption{Prompt sensitivity on the binary trivia datasets, measured by
R@0.1FPR. Top: Output Prediction; bottom: GT Prediction. Bars use the same
standard-prompt delta convention as Figure~\ref{fig:trivia-prompt-sensitivity-auroc}.}
\label{fig:trivia-prompt-sensitivity-recall}
\end{figure*}

Section~\ref{sec:exp-prompt-format} summarizes whether the trivia conclusions
depend on the exact answer prompt used to elicit monitored-model completions.
Here we use the same six trivia datasets and linear-probe setup to compare the
standard prompt against two prompt variants: a shorter minimal prompt and a
more explicit verbose prompt. We fit the probes and select source models once
using the default-prompt data, then apply them unchanged to activations elicited
by the two prompt variants; the shifted prompts do not trigger probe retraining
or new source selection. The goal is twofold: first, to test whether
anchored probes are more or less prompt-sensitive than standard probes; and
second, to test whether prompt variation compounds the cross-model transfer
penalty. Figures~\ref{fig:trivia-prompt-sensitivity-auroc}
and~\ref{fig:trivia-prompt-sensitivity-recall} plot the change in performance
for each method relative to that same method's score under the standard prompt,
so negative bars indicate degradation under the shifted prompt. The standard
prompt is the default chat-formatted multiple-choice prompt described in
Appendix~\ref{app:trivia-implementation-details}. The two prompt variants have the
following form:
\begin{quote}
\small
\textbf{Minimal prompt}\\[2pt]
\texttt{[User] \{question\}}\\
\texttt{A. \{answer A\}}\\
\texttt{B. \{answer B\}}\\
\texttt{\ldots}\\
\texttt{Answer:}
\end{quote}
\begin{quote}
\small
\textbf{Verbose prompt}\\[2pt]
\texttt{[System] You are a knowledgeable assistant. The following is a}\\
\texttt{multiple choice question. Read it carefully and select the best}\\
\texttt{answer. Respond with only the letter corresponding to your choice}\\
\texttt{(A/B/\ldots). Do not include any explanation.}\\
\texttt{[User] Question: \{question\}}\\
\texttt{Options:}\\
\texttt{A. \{answer A\}}\\
\texttt{B. \{answer B\}}\\
\texttt{\ldots}\\
\texttt{Your answer (letter only):}
\end{quote}

This stress test is motivated by prior work showing that hidden-state probes
can depend on seemingly incidental setup choices. For example,
\citet{Orgad2024-zp} find that token choice matters substantially for
white-box hallucination detection, while
\citet{Slobodkin2023-unanswerability} study prompt manipulation and decode
answerability from first-generated-token representations. More broadly, prior
work warns that white-box probes or telemetry-based detectors can be brittle
under task, context, deployment, or adaptive model-side shifts
\citep{Levinstein2023-no-lie-detector, Kramar2026-zv, Min2026-corvus}. We
therefore ask whether anchored probes and cross-model transfer show the same
sensitivity to prompt variation as same-model standard probes.

We observe that same-model probes are
quite sensitive to the minimal prompt on GT Prediction. Standard probes drop by
$11.1$ AUROC points on average across the six Trivia datasets, while same-model
Anchored probes drop by $8.1$ points for global probes and $8.9$ points for
local probes. Thus anchoring reduces prompt sensitivity slightly relative to
Standard, but does not eliminate it. The
same effect is even larger for recall: the minimal prompt reduces Standard recall by
$23.7$ points and same-model Anchored recall by roughly $13.6$--$13.8$ points.
Both AUROC and recall drops are larger than the same-model-to-transfer gap under the standard
prompt. For example, global Anchored GT
Prediction AUROC drops by $8.1$ points under the minimal prompt, whereas the
average gap between same-model global Anchored and global Transfer~(cal) under
the standard prompt is $4.1$ points.

Calibrated cross-model transfer is much less affected by the same
prompt shift. For GT Prediction, global Transfer~(cal) changes by only $1.1$
AUROC points on average under the minimal prompt, and local
Transfer~(cal) changes by $1.0$ point. Recall shows the same
pattern: global Transfer~(cal) changes by $1.4$ points on average, while local
Transfer~(cal) changes by $0.6$ points, despite the much larger same-model prompt
sensitivity. One potential
interpretation is that the strongest same-model probes exploit features that
are predictive but brittle: they work for the same model and prompt style, but
do not transfer reliably across either model identity or prompt wording. 
These features are not predictive under model transfer, which could explain why the drop when changing prompt style is much smaller.
This interpretation is tentative: one family of prompt perturbations does not
establish that transferred probes generally select more robust features.

Output Prediction is substantially more stable (see the different y-axis scale). Under the minimal prompt, most
AUROC changes are within a few points, with the largest drops concentrated in
the minimal-prompt four-choice variants.

Overall, same-model Trivia probes, including same-model anchored probes, are quite sensitive to a compressed prompt on the harder GT Prediction
target. However, calibrated transfer remains stable across the same prompt
shifts, suggesting that the transfer results are potentially more stable to prompt variations.

\FloatBarrier

\FloatBarrier

%% file: content/rebuttal_sections/trivia_linear_anchor_count.tex
\subsection{Anchor count ablations}
\label{app:trivia-linear-anchor-count}

The trivia experiments in Sections~\ref{sec:exp-same-model}
and~\ref{sec:exp-cross-model} use $300$ anchors. To check whether this
choice drives the linear-probe results, we repeat the global anchored
cross-model evaluation with $100$, $200$, $300$, and $400$ anchors while
holding the representation type, probe family, and calibration rule fixed.
Table~\ref{tab:final_trivia_linear_anchor_sensitivity_auroc} reports AUROC for calibrated
cross-model transfer (Transfer~(cal)) for the same global anchored
representations used in the main Trivia linear experiments.

The AUROC results are largely insensitive to the number of anchors. Output
Prediction is near-saturated for all six Trivia datasets; its largest change is
the 1.3-point dip at 300 anchors on binary TruthfulQA. GT Prediction is also
stable on MMLU, SciQ, and their four-choice variants. The clearest exception is
binary TruthfulQA GT Prediction, where AUROC is higher with 100--200 anchors
than with 300--400 anchors. This variation is modest relative to the main
cross-model-transfer trends and does not suggest that larger anchor sets
systematically improve transfer.

\input{content/_tables/single_tables/trivia/linear/anchor_sensitivity/anchor_sensitivity_auroc.tex}

%% file: content/_tables/single_tables/trivia/linear/anchor_sensitivity/anchor_sensitivity_auroc.tex
\begin{table}[htbp]
\centering
\caption{Trivia linear-probe AUROC cross-model transfer (Transfer (cal)) as anchor set size varies from 100 to 400. All probes use global anchored representations and sklearn linear probes with C=1.0.}
\label{tab:final_trivia_linear_anchor_sensitivity_auroc}
\begin{adjustbox}{max width=\textwidth}
\begin{tabular}{llcccc}
\toprule
Dataset & Task & 100 anchors & 200 anchors & 300 anchors & 400 anchors \\
\midrule
TQA & Output Pred. & $99.4\pm0.1$ & $99.6\pm0.1$ & $98.3\pm1.3$ & $99.6\pm0.2$ \\
MMLU & Output Pred. & $99.8\pm0.0$ & $99.8\pm0.1$ & $99.8\pm0.1$ & $99.7\pm0.1$ \\
SciQ & Output Pred. & $100.0\pm0.0$ & $100.0\pm0.0$ & $99.9\pm0.0$ & $99.9\pm0.0$ \\
TQA (4) & Output Pred. & $98.6\pm0.2$ & $98.8\pm0.2$ & $99.0\pm0.2$ & $99.0\pm0.1$ \\
MMLU (4) & Output Pred. & $99.5\pm0.1$ & $99.5\pm0.0$ & $99.5\pm0.0$ & $99.5\pm0.0$ \\
SciQ (4) & Output Pred. & $99.9\pm0.0$ & $99.9\pm0.0$ & $99.9\pm0.0$ & $99.9\pm0.0$ \\
\midrule
TQA & GT Pred. & $75.9\pm1.1$ & $75.1\pm0.4$ & $73.1\pm0.5$ & $72.7\pm0.4$ \\
MMLU & GT Pred. & $96.1\pm0.3$ & $96.3\pm0.2$ & $96.2\pm0.2$ & $96.2\pm0.2$ \\
SciQ & GT Pred. & $99.3\pm0.1$ & $99.3\pm0.0$ & $99.3\pm0.0$ & $99.3\pm0.0$ \\
TQA (4) & GT Pred. & $75.9\pm0.3$ & $77.1\pm1.0$ & $76.9\pm1.3$ & $77.1\pm0.6$ \\
MMLU (4) & GT Pred. & $93.6\pm0.4$ & $93.7\pm0.5$ & $94.3\pm0.2$ & $94.6\pm0.2$ \\
SciQ (4) & GT Pred. & $97.8\pm0.4$ & $98.0\pm0.3$ & $98.0\pm0.3$ & $98.1\pm0.3$ \\
\bottomrule
\end{tabular}
\end{adjustbox}
\end{table}

%% file: content/rebuttal_sections/scale_and_transfer_limits.tex
\subsection{Pooled-MMLU scale and transfer diagnostics}
\label{app:trivia-scale-and-transfer-limits}

\paragraph{Data and anchor scale.}

The model-scale analysis summarized in Section~\ref{sec:exp-model-scale} uses
nested 1K/3K/6K/10K subsets from a 14,042-example pooled-MMLU dataset and
repeats cross-model GT Prediction across three seeds.
We evaluate 300 and 1000 anchors except in the 1K condition, whose 600-example
training split cannot supply 1000 anchors. The scale matrix averages the
available data-size and anchor-count conditions within each seed before
averaging the three seeds; the anchor-count comparison below uses only the
common 3K/6K/10K conditions.
Table~\ref{tab:mmlu-data-anchor-scale} shows no systematic improvement with
more probe-training data: cross-model AUROC is $75.8$ at 1K examples and
$76.5$ at 10K, with a lower intermediate 3K cell. On the common
3K/6K/10K grid, 300 anchors outperform 1000 ($76.8$ versus $74.7$
cross-model AUROC), consistent with the Trivia anchor-count experiment in
Appendix~\ref{app:trivia-linear-anchor-count}.

\input{content/_tables/single_tables/trivia/linear/mmlu_scale/mmlu_data_anchor_scale.tex}

Table~\ref{tab:mmlu-scale-gt-prediction} reports transfer across the
same-family 1B/8B/70B ladder. It shows strong
upward transfer into the 70B target and weak downward transfer into the 1B
target. This scale experiment uses naive transfer and is separate from the
six-model cross-family experiment in Section~\ref{sec:exp-cross-model}. The
asymmetry repeats the weak-target pattern in
Section~\ref{sec:exp-cross-model}. One possible
interpretation is that features used by the 1B probe remain predictive in the
larger models, whereas probes trained on larger models rely on richer
distinctions that are absent or encoded differently in the 1B model.

\input{content/_tables/single_tables/trivia/linear/mmlu_scale/mmlu_scale_matrix_gt_prediction.tex}

\paragraph{Target-performance diagnostic.}
\label{app:trivia-transfer-failure-diagnostics}

Across the six main Trivia variants and six target models, target answer
accuracy correlates strongly with mean incoming off-diagonal GT-transfer AUROC
(Spearman $\rho=0.976$, Pearson $r=0.888$, $n=36$ target--dataset cells).
This extends the observed weak-target pattern from TruthfulQA to pooled MMLU.
Because target accuracy requires labels, it is a retrospective correlate rather
than a label-free pre-deployment signal. Figure~\ref{fig:tqa-gt-confusion}
illustrates the pattern on binary
TruthfulQA: transfer into Llama-3.2-3B remains near or below chance from every
off-diagonal source, even though transfer into most other targets is strong.
Within this dataset, target accuracy and mean incoming transfer have Spearman
$\rho=1.000$ (Pearson $r=0.964$, $n=6$); Llama-3.2-3B is lowest on both,
with $55.8\%$ answer accuracy and $46.4$ mean incoming AUROC.

%% file: content/_tables/single_tables/trivia/linear/mmlu_scale/mmlu_data_anchor_scale.tex
\begin{table}[htbp]
\centering
\caption{Pooled-MMLU GT Prediction scale checks. Data-size rows average all valid anchor counts at that size. Anchor-count rows use the common 3K/6K/10K cells. Values are AUROC (\%), mean $\pm$ standard error over three seeds.}
\label{tab:mmlu-data-anchor-scale}
\begin{tabular}{lrr}
\toprule
\textbf{Setting} & \textbf{All pairs} & \textbf{Cross-model} \\
\midrule
1K examples & $78.3\pm2.2$ & $75.8\pm2.9$ \\
3K examples & $77.3\pm0.3$ & $74.3\pm0.9$ \\
6K examples & $78.9\pm0.5$ & $76.3\pm0.6$ \\
10K examples & $79.1\pm0.2$ & $76.5\pm0.2$ \\
\midrule
300 anchors & $79.1\pm0.4$ & $76.8\pm0.3$ \\
1000 anchors & $77.8\pm0.3$ & $74.7\pm0.6$ \\
\bottomrule
\end{tabular}
\end{table}

%% file: content/_tables/single_tables/trivia/linear/mmlu_scale/mmlu_scale_matrix_gt_prediction.tex
\begin{table}[htbp]
\centering
\caption{Pooled-MMLU GT Prediction AUROC (\%) across the same-family 1B/8B/70B scale ladder. For each seed, we compute AUROC separately for 1K examples with 300 anchors and for 3K, 6K, and 10K examples with either 300 or 1000 anchors. Each cell averages seven AUROCs within a seed, then averages three resulting seed-level values. Values are mean $\pm$ standard error over seeds. Rows are source models and columns are target models.}
\label{tab:mmlu-scale-gt-prediction}
\begin{tabular}{lrrr}
\toprule
\textbf{Source / target} & \textbf{1B} & \textbf{8B} & \textbf{70B} \\
\midrule
1B & $68.2\pm0.7$ & $86.5\pm0.8$ & $94.8\pm0.1$ \\
8B & $51.2\pm0.5$ & $87.3\pm1.1$ & $95.3\pm0.0$ \\
70B & $50.5\pm0.8$ & $76.0\pm1.5$ & $96.0\pm0.2$ \\
\bottomrule
\end{tabular}
\end{table}

%% file: content/_tables/single_tables/trivia/linear/standard_cross_model/standard_cross_model_auroc.tex
\begin{table}[htbp]
\centering
\caption{Standard cross-model linear-probe AUROC for the trivia tasks between Llama-3.1-8B-Instruct and Mistral-7B-Instruct-v0.3. Both models have 32 layers and 4096-dimensional residual streams; probes use layer 25 ($0.8$ relative depth). Values are means $\pm$ standard errors over three seeds.}
\label{tab:trivia-linear-standard-cross-model-auroc}
\begin{adjustbox}{max width=\textwidth}
\begin{tabular}{lcccc}
\toprule
 & \multicolumn{2}{c}{Output Prediction} & \multicolumn{2}{c}{GT Prediction} \\
\cmidrule(lr){2-3} \cmidrule(lr){4-5}
Dataset & Llama$\to$Mistral & Mistral$\to$Llama & Llama$\to$Mistral & Mistral$\to$Llama \\
\midrule
TQA & $42.5\pm12.1$ & $51.5\pm3.7$ & $44.8\pm2.7$ & $50.1\pm3.2$ \\
SciQ & $69.5\pm17.2$ & $67.3\pm15.8$ & $19.6\pm4.6$ & $59.0\pm16.1$ \\
MMLU & $56.4\pm7.2$ & $65.4\pm10.3$ & $35.8\pm10.6$ & $43.6\pm11.8$ \\
TQA (4) & $51.1\pm1.3$ & $38.9\pm2.0$ & $47.2\pm2.2$ & $48.6\pm5.3$ \\
SciQ (4) & $45.5\pm4.1$ & $35.9\pm2.3$ & $32.9\pm3.5$ & $52.4\pm7.1$ \\
MMLU (4) & $50.0\pm1.7$ & $46.0\pm7.0$ & $54.3\pm6.4$ & $44.0\pm3.8$ \\
\bottomrule
\end{tabular}
\end{adjustbox}
\end{table}

%% file: content/_tables/single_tables/trivia/linear/standard_cross_model/anchored_cross_model_auroc.tex
\begin{table}[htbp]
\centering
\caption{Global anchored linear-probe AUROC for the trivia tasks between Llama-3.1-8B-Instruct and Mistral-7B-Instruct-v0.3 ($300$ anchors). Values are means $\pm$ standard errors over three seeds.}
\label{tab:trivia-linear-anchored-cross-model-auroc}
\begin{adjustbox}{max width=\textwidth}
\begin{tabular}{lcccc}
\toprule
 & \multicolumn{2}{c}{Output Prediction} & \multicolumn{2}{c}{GT Prediction} \\
\cmidrule(lr){2-3} \cmidrule(lr){4-5}
Dataset & Llama$\to$Mistral & Mistral$\to$Llama & Llama$\to$Mistral & Mistral$\to$Llama \\
\midrule
TQA & $99.4\pm0.5$ & $98.6\pm0.7$ & $81.7\pm1.7$ & $62.9\pm8.7$ \\
SciQ & $100.0\pm0.0$ & $100.0\pm0.0$ & $97.5\pm0.4$ & $99.9\pm0.0$ \\
MMLU & $99.9\pm0.0$ & $99.7\pm0.1$ & $95.1\pm1.6$ & $96.4\pm0.7$ \\
TQA (4) & $99.7\pm0.1$ & $95.9\pm1.7$ & $78.8\pm0.9$ & $73.2\pm1.5$ \\
SciQ (4) & $99.9\pm0.0$ & $99.9\pm0.0$ & $92.3\pm1.5$ & $99.1\pm0.4$ \\
MMLU (4) & $99.8\pm0.1$ & $98.7\pm0.6$ & $93.4\pm0.8$ & $94.6\pm0.9$ \\
\bottomrule
\end{tabular}
\end{adjustbox}
\end{table}

%% file: content/_tables/single_tables/trivia/linear/abs/ls0.8/abs_model-choice_auroc.tex
\begin{table}[htbp]
\centering
\caption{Linear-probe AUROC on trivia for Output Prediction: same-model standard probes at relative depth 0.8 and on concatenated all-layer activations (C=1.0, C=0.1, C=0.01). Dataset abbreviations: TQA = TruthfulQA; (4) denotes the four-choice variant.}
\label{tab:final_trivia_linear_abs_model-choice_auroc}
\begin{adjustbox}{max width=\textwidth}
\begin{tabular}{lcccc}
\toprule
Dataset & Single-layer & All-layer C=1.0 & All-layer C=0.1 & All-layer C=0.01 \\
\midrule
TQA & $99.7\pm0.0$ & $99.7\pm0.1$ & $99.6\pm0.1$ & $99.7\pm0.1$ \\
MMLU & $99.8\pm0.0$ & $99.8\pm0.0$ & $99.8\pm0.0$ & $99.8\pm0.1$ \\
SciQ & $100.0\pm0.0$ & $99.9\pm0.0$ & $100.0\pm0.0$ & $100.0\pm0.0$ \\
TQA (4) & $99.5\pm0.1$ & $99.5\pm0.1$ & $99.4\pm0.0$ & $99.4\pm0.1$ \\
MMLU (4) & $99.7\pm0.0$ & $99.7\pm0.1$ & $99.6\pm0.0$ & $99.6\pm0.0$ \\
SciQ (4) & $100.0\pm0.0$ & $100.0\pm0.0$ & $100.0\pm0.0$ & $100.0\pm0.0$ \\
\bottomrule
\end{tabular}
\end{adjustbox}
\end{table}

%% file: content/_tables/single_tables/trivia/linear/abs/ls0.8/abs_gt-prediction_auroc.tex
\begin{table}[htbp]
\centering
\caption{Linear-probe AUROC on trivia for GT Prediction: same-model standard probes at relative depth 0.8 and on concatenated all-layer activations (C=1.0, C=0.1, C=0.01). Dataset abbreviations: TQA = TruthfulQA; (4) denotes the four-choice variant.}
\label{tab:final_trivia_linear_abs_gt-prediction_auroc}
\begin{adjustbox}{max width=\textwidth}
\begin{tabular}{lcccc}
\toprule
Dataset & Single-layer & All-layer C=1.0 & All-layer C=0.1 & All-layer C=0.01 \\
\midrule
TQA & $89.4\pm1.2$ & $89.3\pm0.6$ & $89.4\pm0.5$ & $89.6\pm0.5$ \\
MMLU & $94.8\pm0.2$ & $94.5\pm0.2$ & $95.0\pm0.2$ & $95.2\pm0.3$ \\
SciQ & $99.2\pm0.1$ & $99.2\pm0.1$ & $99.2\pm0.1$ & $99.3\pm0.1$ \\
TQA (4) & $85.5\pm0.3$ & $84.3\pm0.8$ & $85.5\pm0.5$ & $85.8\pm0.5$ \\
MMLU (4) & $93.4\pm0.2$ & $93.5\pm0.3$ & $93.9\pm0.3$ & $94.2\pm0.2$ \\
SciQ (4) & $98.2\pm0.3$ & $98.1\pm0.2$ & $98.1\pm0.2$ & $98.3\pm0.2$ \\
\bottomrule
\end{tabular}
\end{adjustbox}
\end{table}

%% file: content/rebuttal_sections/guard_linear_additional_results.tex
\subsection{Full calibration progression}
\label{app:guard-calibration-progression}

Section~\ref{sec:exp-cross-model} reports \texttt{Transfer} (naive transfer) and
\texttt{Transfer~(cal)} (calibrated transfer with $N\!=\!5$) as representative
endpoints. Table~\ref{tab:final_guard_guard-safety_cross_linear_concat_auroc}
reports the complete $N\!=\!1,\dots,5$ progression alongside the Anchored
same-model column for Guard Distillation linear probes with single-token
anchors, where $N$ is the number of available source models.
Transfer reaches $81.1$ AUROC on BeaverTails and $81.6$ on Toxic-chat;
Transfer~(cal) improves both datasets, reaching $86.4$ on BeaverTails and
$83.6$ on Toxic-chat. Relative to chance, Transfer~(cal) preserves $81.1\%$ and
$89.6\%$ of same-model Anchored performance, respectively.

\input{content/_tables/single_tables/guard/linear/non-abs/agg-single/ls0.8/cross_guard-safety_auroc.tex}

\subsection{Recall at 10\% FPR}
\label{app:guard-linear-recall}

Low-FPR recall retains the AUROC ordering but has larger gaps between
same-model and transferred probes. With single-token
anchors, Table~\ref{tab:final_guard_linear_baseline_linear_concat_recall} shows
that same-model Standard and Anchored recall are close: Anchored reaches $82.0$
on BeaverTails and $65.9$ on Toxic-chat, compared with $86.0$ and $68.6$ for the
single-layer Standard probe at relative depth~$0.8$. The full calibration
progression in Table~\ref{tab:final_guard_guard-safety_cross_linear_concat_recall}
shows that naive Transfer drops to $55.4$ and $47.7$, respectively, while
Transfer~(cal) improves to $63.5$ on BeaverTails and $53.0$ on Toxic-chat. Thus
calibration remains useful at low FPR, especially on BeaverTails, but the recall
metric exposes more cross-model fragility than AUROC.

\input{content/_tables/single_tables/guard/linear/non-abs/agg-single/ls0.8/baseline_recall.tex}

\input{content/_tables/single_tables/guard/linear/non-abs/agg-single/ls0.8/cross_guard-safety_recall.tex}

\FloatBarrier

\subsection{Transfer to external model targets}
\label{app:external-target-guard}

Section~\ref{sec:exp-external-families} summarizes transfer to Guard
Distillation targets outside the original six-model pool.
Table~\ref{tab:external-guard-full} separates BeaverTails and Toxic-chat and
reports their unsafe-completion rates. Toxic-chat provides the more stable
external-target test: all seed-level AUROCs are defined and calibrated
transfer remains strongest for OLMo-3. BeaverTails has only 1.0--4.8\% unsafe
completions across the three external models. One OLMo-3 test seed contains no
unsafe example, so both its target-trained and transferred AUROC are undefined;
the reported OLMo-3 BeaverTails means use the other two seeds. This imbalance
makes BeaverTails a demanding, high-uncertainty test, while the lower GPT-OSS
and Gemma-3 point estimates show that low prevalence is not the only relevant
source of transfer error.

\input{content/_tables/single_tables/external_targets/external_guard_full.tex}

\FloatBarrier

\subsection{Related-data and joint transfer}
\label{app:guard-linear-cross-data-transfer}

Section~\ref{sec:exp-related-data} summarizes transfer between BeaverTails and
Toxic-chat. Here we report both directions, first for the same monitored model
and then while also changing model identity, using the common protocol in
Appendix~\ref{app:related-data-implementation}. Hidden-state detectors can be
brittle under such distribution shifts
\citep{Orgad2024-zp, Levinstein2023-no-lie-detector, Kramar2026-zv,
Min2026-corvus}.

Table~\ref{tab:guard-linear-cross-data-transfer-single-anchor} reports the
same-model result. Averaged over directions, Standard, Global, and Local reach
$78.3$, $77.9$, and $77.9$ AUROC. This aggregate hides a strong asymmetry:
Global reaches $70.2$ from BeaverTails to Toxic-chat but $85.5$ in the reverse
direction. Anchoring therefore roughly matches Standard on average but does not
make this distribution shift uniformly robust.

\input{content/_tables/single_tables/guard/linear/cross_data_transfer/guard_linear_cross_data_transfer_single_anchor.tex}

Changing both dataset and model further reduces performance
(Table~\ref{tab:guard-cross-data-xmodel-single}). Global Transfer reaches
$64.3$/$75.8$ by direction and $70.1$ overall; source calibration raises the
reverse direction to $79.6$ but changes the weak direction only to $65.1$,
for $72.4$ overall. Calibration therefore does not close the directional gap.

\input{content/_tables/single_tables/guard/linear/cross_data_transfer/guard_linear_cross_data_xmodel_transfer_single_anchor.tex}

\subsection{Anchor-aggregation sensitivity}
\label{app:guard-linear-mean-token-auroc}

We repeat both same-model related-data transfer and joint model-and-dataset
transfer with \emph{mean-token anchors}, which average each anchor completion's
token representations before computing anchor-relative features. The anchor
index still refers to the same prompt across models, but the corresponding
vector now depends on each model's generated completion. This comparison tests
whether prompt-indexed relational structure can remain useful even when exact
anchor tokens are not shared.

When training and evaluating on the same dataset, mean-token anchors change
same-model AUROC by at most $1.1$ points and calibrated transfer by at most
$2.7$ points. Under same-model related-data transfer, they raise the
Global/Local averages from
$77.9$/$77.9$ to $80.0$/$80.6$ (Table~\ref{tab:guard-linear-cross-data-transfer}).
When both dataset and model change, switching to mean-token anchors has mixed
effects: calibrated Global improves from $72.4$ to $75.0$ overall, while naive
Global falls from $70.1$ to $65.8$
(Table~\ref{tab:guard-cross-data-xmodel-mean}).
Mean-token anchors are therefore a useful sensitivity result, not a uniformly
better replacement for the single-token-anchor baseline.

\input{content/_tables/single_tables/guard/linear/cross_data_transfer/guard_linear_cross_data_transfer.tex}

\input{content/_tables/single_tables/guard/linear/cross_data_transfer/guard_linear_cross_data_xmodel_transfer_mean_anchor.tex}

\subsection{Anchor coverage and mixed-anchor mitigation}
\label{app:guard-mixed-anchor-mitigation}

The weak BeaverTails$\rightarrow$Toxic-chat direction also has poorer
label-free anchor coverage: its mean nearest-anchor angular distance is
$1.209$, versus $1.182$ in the reverse direction, while same-model degradation
is $17.3$ versus $9.3$ points. Across all applications this association is only
moderate; nearest-anchor distance tracks the directional difference here but is
not an established general predictor.

We test a coverage-oriented intervention using an unlabeled 50/50 anchor pool
from both datasets. In the mean-token setting, mixed global anchors narrow the
naive direction gap from $9.6$ to $1.9$ points, but leave average naive AUROC
similar and reduce calibrated average AUROC from $75.0$ to
$70.4$. They do improve low-FPR recall in the weak direction, from $17.9$ to
$28.7$ naively and from $24.9$ to $35.5$ after calibration. Mixed anchors thus
alter the failure pattern and improve weak-direction coverage; they do not
uniformly improve AUROC or solve the underlying shift
(Table~\ref{tab:guard-union-anchors-concat}).

\input{content/_tables/single_tables/guard/linear/cross_data_transfer/guard_union_anchor_concat.tex}

Local mixed-anchor results are less stable and sometimes reverse the
directional imbalance. Mixed anchors therefore do not provide a consistently
better response to the data shift
(Table~\ref{tab:guard-union-anchors-perlayer}).

\input{content/_tables/single_tables/guard/linear/cross_data_transfer/guard_union_anchor_perlayer.tex}

\subsection{Validation-selected local depths}
\label{app:guard-linear-val-plateau}

\begin{figure*}[tbp]
\centering
\includegraphics[width=\textwidth]{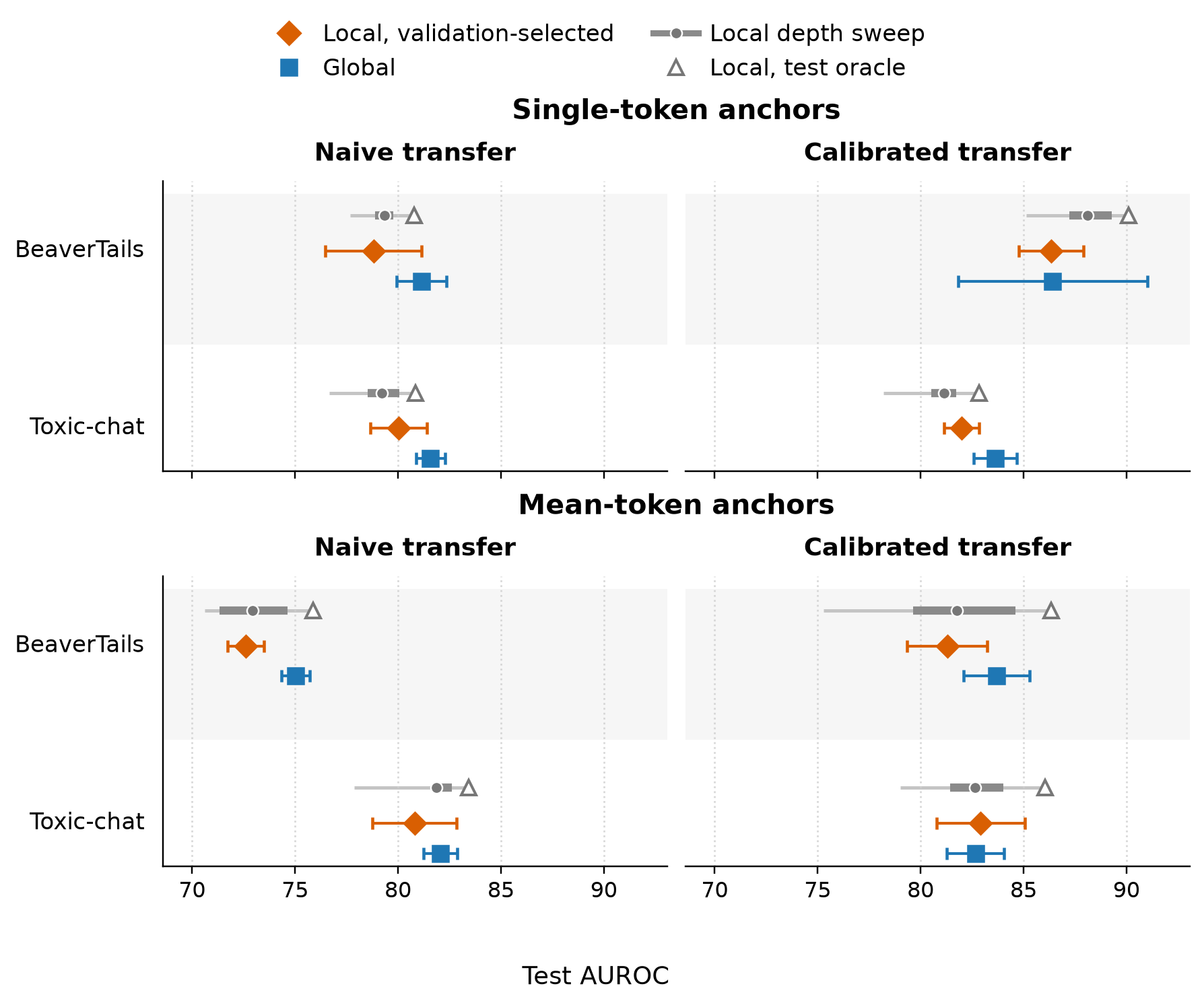}
\caption{\textbf{Guard local-depth policies with single-token and mean-token
anchors.} Orange diamonds select source and depth using source-pool validation;
blue squares are global anchored probes. Gray ranges summarize local test
AUROC across depths, and hollow triangles choose depth post hoc on test AUROC
while retaining the source policy. Mean-token anchors are an appendix
robustness analysis; Section~\ref{sec:exp-cross-model} uses single-token
anchors. Error bars are standard errors over three seeds.}
\label{fig:guard-validation-depth-policy-all}
\end{figure*}

Figure~\ref{fig:guard-validation-depth-policy-all} compares local-depth
selection with the global representation for both single-token and mean-token
anchors. For same-model evaluation, each model and seed selects a depth using
its own validation split. For naive transfer, each source selects its depth
using source-model validation. For calibrated transfer, source-pool validation
jointly selects source and depth for each in-pool target. All reported values
are then evaluated on test data; only the hollow-triangle oracle uses test
AUROC to choose depth.

For the single-token comparison summarized in
Section~\ref{sec:exp-global-vs-local}, validation-based depth selection improves
calibrated local transfer from $85.1$ to $86.4$ AUROC on BeaverTails and from
$80.4$ to $82.0$ on Toxic-chat. The selected local probe ties global on BeaverTails and trails it
on Toxic-chat ($82.0$ versus $83.6$). Across both anchor constructions, joint
source--depth validation often improves on fixed depth
$0.8$, but does not consistently realize the stronger local result visible to
the test oracle. Global anchoring is therefore the more reliable depth-free
choice under the available validation protocol.

\FloatBarrier

\FloatBarrier


\subsection{Class balancing ablation}
\label{app:guard-linear-class-balanced-linear}

We also check whether class balancing is useful for linear Guard Distillation probes.
In particular, we run mean-token-anchor global probes with class balancing.
Table~\ref{tab:guard-linear-class-balance-diagnostic}
shows that class balancing is not an across-the-board improvement. It slightly
improves BeaverTails same-model AUROC and recall, but worsens
BeaverTails naive-transfer AUROC and recall. On Toxic-chat, class balancing is
worse in both same-model and transfer comparisons for both metrics. Thus despite
the heavy imbalance in the Guard Distillation data
(Table~\ref{tab:guard-guard-label-balance}), the
linear Guard results do not suggest that class weighting is generally
beneficial.

\input{content/_tables/single_tables/guard/linear/class_balance/class_balance_diagnostic.tex}

\subsection{Sequence probing with attention probes} \label{app:guard-linear-attention-diagnostics}
Because the Guard Distillation probes aggregate completion-token representations, we
also test whether a learned attention-pooling probe can improve over fixed
mean pooling. This is plausible because attention can, in principle, focus on
especially informative generated tokens rather than averaging over the full
completion. Like the class-balancing check above, this diagnostic uses the
mean-token-anchor variant rather than the main single-token-anchor baseline. The
diagnostic attention probe uses the same global anchored feature construction and
mean-token anchors as the fixed linear/mean-token reference:
$600$ sampled anchors, angular-distance relative coordinates,
completion-token relative activations, and the same six monitored models and
Guard datasets. Architecturally, it is a single-head attention-pooling
probe: a learned linear score is applied to each completion-token feature vector,
a softmax pools tokens into one weighted feature vector, and a linear binary
classifier predicts the Guard label. We train these probes in PyTorch with
AdamW, learning rate $10^{-3}$, weight decay $10^{-2}$, batch size $64$, and up to
$20$ epochs with validation-loss early stopping with patience $3$.

In diagnostic runs, unbalanced attention pooling is consistently worse
than a matched linear probe with mean pooling, with large AUROC gaps for both
same-model Anchored evaluations and cross-model Transfer on both Guard datasets.

We also try class balancing as a diagnostic for why the unbalanced attention
probe underperforms. The motivation is that Guard labels are skewed, so an
unbalanced training objective can obtain reasonable loss or accuracy by
emphasizing the majority class and underweighting the rarer safety violations.
This concern is stronger for attention than for the linear/mean probes because
attention must also learn which tokens to pool; if minority-class evidence is
rare, the pooling rule can receive little pressure to attend to those tokens.
Class balancing helps substantially, suggesting that label skew is a real
optimization issue for learned attention pooling. This contrasts with the
fixed linear/mean probes in
Section~\ref{app:guard-linear-class-balanced-linear}, where class balancing
does not help overall and tends to hurt transfer. Even after balancing,
however, attention remains below the linear/mean reference overall: the
BeaverTails transfer AUROC gap is small, but same-model performance,
Toxic-chat transfer, and recall at $0.1$ FPR remain clearly weaker.

As an additional regularization check, we run the class-balanced attention
diagnostic with AdamW weight decay set to $10^{-1}$ instead of $10^{-2}$, while
keeping the same AUROC checkpoint selection. This stronger regularization gives
small AUROC gains and improves same-model low-FPR recall, especially on
BeaverTails, but does not close the gap to the linear/mean reference. BeaverTails
transfer AUROC is close, but same-model performance, Toxic-chat transfer, and
low-FPR recall still lag.

Completion length is a consistent diagnostic difference, but the detailed
pattern depends on the dataset and regularization setting. In BeaverTails
transfer, attention is closest to linear/mean on short completions and remains
substantially worse on long completions; class balancing and stronger weight
decay reduce this long-completion gap but do not remove it. Toxic-chat also
shows length sensitivity, though the largest remaining gap after balancing is
on middle-length completions rather than the longest completions, and stronger
regularization mostly preserves that pattern. A similar pattern applies to the 
same-model setting: on same-model BeaverTails evaluations, balanced attention is close to
linear/mean on short completions but substantially worse on longer completions,
including under stronger regularization.

In the tested configurations, attention pooling is a worse inductive bias for
Guard Distillation relative-token sequences, even after reducing the
label-imbalance issue. This is not a generic failure of the attention
architecture: for the Standard same-model probing baseline, attention is close
to linear/mean in AUROC and at low FPR. The observed underperformance is thus
specific to the combination of attention pooling and anchored features in this
experiment; determining whether data volume, regularization, or another factor
causes the gap requires further study.

%% file: content/_tables/single_tables/guard/linear/non-abs/agg-single/ls0.8/cross_guard-safety_auroc.tex
\begin{table}[htbp]
\centering
\caption{Linear-probe AUROC for Guard Distillation: same-model Anchored and cross-model Transfer as the available source pool grows. Transfer (cal-N) uses source-only validation within each N-model pool.}
\label{tab:final_guard_guard-safety_cross_linear_concat_auroc}
\begin{adjustbox}{max width=\textwidth}
\begin{tabular}{lcccccc}
\toprule
Dataset & Anchored & Transfer & Transfer (cal-2) & Transfer (cal-3) & Transfer (cal-4) & Transfer (cal) \\
\midrule
BeaverTails & $94.9\pm0.2$ & $81.1\pm1.2$ & $\mathbf{86.5}\pm1.5$ & $85.4\pm1.9$ & $85.9\pm2.9$ & $86.4\pm4.6$ \\
Toxic-chat & $87.5\pm0.7$ & $81.6\pm0.7$ & $82.0\pm0.5$ & $82.5\pm0.6$ & $83.0\pm0.8$ & $\mathbf{83.6}\pm1.0$ \\
\bottomrule
\end{tabular}
\end{adjustbox}
\end{table}

%% file: content/_tables/single_tables/guard/linear/non-abs/agg-single/ls0.8/baseline_recall.tex
\begin{table}[htbp]
\centering
\caption{Linear-probe R@0.1FPR for Guard Distillation: same-model Standard and global Anchored probes.}
\label{tab:final_guard_linear_baseline_linear_concat_recall}
\begin{adjustbox}{max width=\textwidth}
\begin{tabular}{lcc}
\toprule
 & \multicolumn{2}{c}{Guard Distillation} \\
\cmidrule(lr){2-3}
Dataset & Standard & Anchored \\
\midrule
BeaverTails & $86.0\pm1.2$ & $82.0\pm2.1$ \\
Toxic-chat & $68.6\pm2.2$ & $65.9\pm3.4$ \\
\bottomrule
\end{tabular}
\end{adjustbox}
\end{table}

%% file: content/_tables/single_tables/guard/linear/non-abs/agg-single/ls0.8/cross_guard-safety_recall.tex
\begin{table}[htbp]
\centering
\caption{Linear-probe R@0.1FPR for Guard Distillation: same-model Anchored and cross-model Transfer as the available source pool grows. Transfer (cal-N) uses source-only validation within each N-model pool.}
\label{tab:final_guard_guard-safety_cross_linear_concat_recall}
\begin{adjustbox}{max width=\textwidth}
\begin{tabular}{lcccccc}
\toprule
Dataset & Anchored & Transfer & Transfer (cal-2) & Transfer (cal-3) & Transfer (cal-4) & Transfer (cal) \\
\midrule
BeaverTails & $82.0\pm2.1$ & $55.4\pm2.7$ & $\mathbf{63.6}\pm3.9$ & $60.1\pm4.8$ & $61.9\pm7.1$ & $63.5\pm10.5$ \\
Toxic-chat & $65.9\pm3.4$ & $47.7\pm1.9$ & $48.3\pm2.1$ & $49.9\pm2.9$ & $51.5\pm3.7$ & $\mathbf{53.0}\pm4.8$ \\
\bottomrule
\end{tabular}
\end{adjustbox}
\end{table}

%% file: content/_tables/single_tables/external_targets/external_guard_full.tex
\begin{table}[htbp]
\centering
\caption{\textbf{External-target Guard results and event rates.} Single-token anchored representations are used throughout. One OLMo-3 BeaverTails seed has no unsafe test examples, so its AUROC is undefined and the corresponding AUROC means use two seeds; prevalence uses all three seeds.}
\label{tab:external-guard-full}
\begin{tabular*}{\textwidth}{@{\extracolsep{\fill}}llccc@{}}
\toprule
Target & Dataset & Target-trained & Global transfer & Unsafe prevalence \\
\midrule
OLMo-3-7B & BeaverTails & 95.1 $\pm$ 1.7 & 79.2 $\pm$ 3.0 & $(1.0 \pm 0.5)\%$ \\
OLMo-3-7B & Toxic-chat & 89.6 $\pm$ 1.8 & 87.6 $\pm$ 1.4 & $(9.0 \pm 0.9)\%$ \\
GPT-OSS-20B & BeaverTails & 85.2 $\pm$ 1.5 & 57.5 $\pm$ 12.6 & $(1.5 \pm 0.0)\%$ \\
GPT-OSS-20B & Toxic-chat & 73.1 $\pm$ 1.5 & 66.9 $\pm$ 3.8 & $(10.3 \pm 0.2)\%$ \\
Gemma-3-12B & BeaverTails & 85.6 $\pm$ 1.5 & 63.0 $\pm$ 7.5 & $(4.8 \pm 0.9)\%$ \\
Gemma-3-12B & Toxic-chat & 86.9 $\pm$ 2.1 & 72.6 $\pm$ 11.4 & $(11.8 \pm 0.3)\%$ \\
\bottomrule
\end{tabular*}
\end{table}

%% file: content/_tables/single_tables/guard/linear/cross_data_transfer/guard_linear_cross_data_transfer_single_anchor.tex
\begin{table}[htbp]
\centering
\small
\caption{Same-model related-data Guard Distillation transfer with single-token anchors. Probes are trained on one safety dataset and evaluated on the other. Global uses all layers; Standard and Local use relative depth 0.8. Values are mean test AUROC (\%) $\pm$ standard error over three seeds and six models.}
\label{tab:guard-linear-cross-data-transfer-single-anchor}
\begin{tabular}{lccc}
\toprule
\textbf{Transfer direction} & \textbf{Standard} & \textbf{Global anchored} & \textbf{Local anchored} \\
\midrule
BeaverTails $\rightarrow$ Toxic-chat & $72.7 \pm 2.0$ & $70.2 \pm 2.1$ & $70.2 \pm 1.9$ \\
Toxic-chat $\rightarrow$ BeaverTails & $83.9 \pm 1.6$ & $85.5 \pm 1.0$ & $85.6 \pm 0.9$ \\
\bottomrule
\end{tabular}
\end{table}

%% file: content/_tables/single_tables/guard/linear/cross_data_transfer/guard_linear_cross_data_xmodel_transfer_single_anchor.tex
\begin{table}[htbp]
\centering
\small
\caption{Guard Distillation related-data and cross-model AUROC (\%) with single-token anchors. Transfer (cal) uses source-only validation.}
\label{tab:guard-cross-data-xmodel-single}
\begin{tabular}{lrrrr}
\toprule
\textbf{Direction} & \textbf{Global} & \textbf{Global (cal)} & \textbf{Local} & \textbf{Local (cal)} \\
\midrule
BeaverTails $\rightarrow$ Toxic-chat & $64.3 \pm 0.5$ & $65.1 \pm 0.5$ & $64.2 \pm 0.8$ & $63.5 \pm 0.4$ \\
Toxic-chat $\rightarrow$ BeaverTails & $75.8 \pm 0.4$ & $79.6 \pm 1.3$ & $69.9 \pm 0.4$ & $83.0 \pm 1.1$ \\
\bottomrule
\end{tabular}
\end{table}

%% file: content/_tables/single_tables/guard/linear/cross_data_transfer/guard_linear_cross_data_transfer.tex
\begin{table}[htbp]
\centering
\small
\caption{Same-model related-data Guard Distillation transfer with mean-token anchors. Probes are trained on one safety dataset and evaluated on the other. Global uses all layers; Standard and Local use relative depth 0.8. Values are mean test AUROC (\%) $\pm$ standard error over three seeds and six models.}
\label{tab:guard-linear-cross-data-transfer}
\begin{tabular}{lccc}
\toprule
\textbf{Transfer direction} & \textbf{Standard} & \textbf{Global anchored} & \textbf{Local anchored} \\
\midrule
BeaverTails $\rightarrow$ Toxic-chat & $72.7 \pm 2.0$ & $71.7 \pm 2.0$ & $73.1 \pm 1.8$ \\
Toxic-chat $\rightarrow$ BeaverTails & $83.9 \pm 1.6$ & $88.3 \pm 1.0$ & $88.0 \pm 1.0$ \\
\bottomrule
\end{tabular}
\end{table}

%% file: content/_tables/single_tables/guard/linear/cross_data_transfer/guard_linear_cross_data_xmodel_transfer_mean_anchor.tex
\begin{table}[htbp]
\centering
\small
\caption{Guard Distillation related-data and cross-model AUROC (\%) with mean-token anchors. Transfer (cal) uses source-only validation.}
\label{tab:guard-cross-data-xmodel-mean}
\begin{tabular}{lrrrr}
\toprule
\textbf{Direction} & \textbf{Global} & \textbf{Global (cal)} & \textbf{Local} & \textbf{Local (cal)} \\
\midrule
BeaverTails $\rightarrow$ Toxic-chat & $61.0 \pm 0.8$ & $70.7 \pm 3.4$ & $58.6 \pm 0.6$ & $69.0 \pm 2.3$ \\
Toxic-chat $\rightarrow$ BeaverTails & $70.6 \pm 0.5$ & $79.2 \pm 5.1$ & $67.7 \pm 0.4$ & $76.0 \pm 2.6$ \\
\bottomrule
\end{tabular}
\end{table}

%% file: content/_tables/single_tables/guard/linear/cross_data_transfer/guard_union_anchor_concat.tex
\begin{table}[htbp]
\centering
\caption{Guard Distillation joint related-data and cross-model transfer with a 50/50 unlabeled BeaverTails/Toxic-chat anchor pool. Values are AUROC (\%) $\pm$ standard error over three seeds.}
\label{tab:guard-union-anchors-concat}
\begin{tabular}{lrr}
\toprule
\textbf{Direction} & \textbf{Transfer} & \textbf{Transfer (cal)} \\
\midrule
BeaverTails $\rightarrow$ Toxic-chat & $64.6 \pm 0.7$ & $70.4 \pm 2.4$ \\
Toxic-chat $\rightarrow$ BeaverTails & $66.5 \pm 0.6$ & $70.3 \pm 3.0$ \\
\bottomrule
\end{tabular}
\end{table}

%% file: content/_tables/single_tables/guard/linear/cross_data_transfer/guard_union_anchor_perlayer.tex
\begin{table}[htbp]
\centering
\caption{Guard Distillation joint related-data and cross-model transfer with a 50/50 unlabeled BeaverTails/Toxic-chat anchor pool. Values are AUROC (\%) $\pm$ standard error over three seeds.}
\label{tab:guard-union-anchors-perlayer}
\begin{tabular}{lrr}
\toprule
\textbf{Direction} & \textbf{Transfer} & \textbf{Transfer (cal)} \\
\midrule
BeaverTails $\rightarrow$ Toxic-chat & $63.8 \pm 1.1$ & $69.6 \pm 4.6$ \\
Toxic-chat $\rightarrow$ BeaverTails & $65.0 \pm 1.1$ & $59.8 \pm 2.2$ \\
\bottomrule
\end{tabular}
\end{table}

%% file: content/_tables/single_tables/guard/linear/class_balance/class_balance_diagnostic.tex
\begin{table}[htbp]
\centering
\caption{Class-balancing diagnostic for Guard Distillation linear/mean global probes. Rows compare the default unweighted linear probe with class-balanced linear training. Transfer averages off-diagonal source-target applications. Values are mean $\pm$ standard error across three seed-level aggregates.}
\label{tab:guard-linear-class-balance-diagnostic}
\begin{adjustbox}{max width=\textwidth}
\begin{tabular}{llrrrrrr}
\toprule
Dataset & Evaluation & \multicolumn{3}{c}{AUROC} & \multicolumn{3}{c}{R@0.1FPR} \\
\cmidrule(lr){3-5}\cmidrule(lr){6-8}
 & & Default & Balanced & $\Delta$ & Default & Balanced & $\Delta$ \\
\midrule
BeaverTails & Anchored & $94.6 \pm 0.3$ & $95.0 \pm 0.4$ & $+0.4 \pm 0.1$ & $79.9 \pm 1.9$ & $83.3 \pm 3.6$ & $+3.4 \pm 2.1$ \\
BeaverTails & Transfer & $75.0 \pm 0.7$ & $72.9 \pm 0.4$ & $-2.1 \pm 0.4$ & $46.8 \pm 1.8$ & $43.3 \pm 0.6$ & $-3.5 \pm 1.8$ \\
Toxic-chat & Anchored & $88.6 \pm 0.4$ & $88.4 \pm 0.6$ & $-0.2 \pm 0.2$ & $67.2 \pm 2.9$ & $66.6 \pm 3.1$ & $-0.6 \pm 0.5$ \\
Toxic-chat & Transfer & $82.1 \pm 0.8$ & $81.2 \pm 1.0$ & $-0.9 \pm 0.2$ & $49.7 \pm 2.2$ & $48.3 \pm 2.3$ & $-1.4 \pm 0.4$ \\
\bottomrule
\end{tabular}
\end{adjustbox}
\end{table}

%% file: content/rebuttal_sections/guard_relational_evidence.tex
\subsection{Anchor correspondence and relational controls}
\label{app:guard-relational-evidence}

Section~\ref{sec:exp-anchor-content} summarizes our tests of how closely anchor
representations must correspond across models. Here we ask whether different
models organize the shared anchor examples similarly
enough for their identities to be recovered from relational geometry alone. We
hide the anchor correspondence and give the matching procedure only the two
anchor-to-anchor distance matrices. Without prompt content or a shared
activation basis, the matcher can identify an anchor only through its pattern
of distances to the other anchors. Above-chance recovery therefore indicates
that these relational patterns are partly preserved across models.
Table~\ref{tab:guard-anchor-recovery-main} summarizes the aggregate result:
graph matching recovers 20.7\% of hidden identities, compared with 0.16\% for
random one-to-one mappings. Applying the source probe after this recovered
mapping yields 79.0 AUROC, versus 54.8 for random mappings and 83.2 with known
prompt correspondence. The recovered correspondence is therefore also
functionally useful. Normal monitor deployment uses known anchor identities
and does not require correspondence recovery.

\input{content/_tables/single_tables/guard/anchor_recovery/guard_anchor_recovery_main.tex}

\paragraph{Blinded recovery protocol.}
For each model pair, we construct a $600\times600$ matrix of angular distances
among the Guard anchors. The target anchor order is randomized. The matching
procedure receives only the source and reordered target matrices. It has no
access to the anchor prompts, hidden correspondence, labels, fitted probe,
validation set, or test set.

We first describe each anchor by 64 quantiles of its distances to the other
anchors. These summaries are invariant to the order of the remaining anchors.
A linear assignment gives an initial one-to-one mapping. We refine the mapping
by maximizing agreement between all pairwise distances. The refinement uses
the initial assignment, a uniform initialization, and two seeded randomized
initializations. We select the mapping with the highest geometric objective.

One BeaverTails seed-1 condition served as the development pilot for the
two-stage procedure, the 64-quantile fingerprint, and the graph-matching
initializations. These choices were then frozen before evaluation across
BeaverTails and Toxic-chat, three seed configurations, and four prespecified
model pairs. Two pairs are within the Llama or Qwen2.5 families. Two
cross-family pairs compare Llama-3.1-8B with Mistral-7B or Qwen2.5-7B. Each of
the 24 pair conditions is evaluated in both transfer directions and compared
with 20 random one-to-one mappings.

\input{content/_tables/single_tables/guard/anchor_recovery/guard_anchor_recovery_scope.tex}

Table~\ref{tab:guard-anchor-recovery-scope} shows substantial variation across
model-pair scope. Exact recovery reaches 20.6\% for same-family BeaverTails
pairs and 46.8\% for same-family Toxic-chat pairs. The corresponding
cross-family values are 10.3\% and 5.3\%. Cross-family recovery remains above
the random expectation. Its functional value depends strongly on the dataset.
Graph-mapped AUROC is 78.9 on the BeaverTails cross-family pairs and 68.8 on the
Toxic-chat cross-family pairs.

Known prompt identity is a correspondence control. Its mean AUROC is higher
than graph matching overall. Graph matching exceeds it in 12 of the 48
directions. Graph matching directly optimizes agreement between the two anchor
graphs and can occasionally find a task-favorable permutation. Its in-sample
edge agreement should therefore not be treated as an independent outcome.
Exact hidden-identity recovery and test-set probe transfer provide the
independent evaluations.

Probe-weighted recovery weights every correctly mapped anchor by the absolute
coefficient assigned to that source coordinate, then divides by the total
absolute coefficient weight. Its similarity to exact recovery indicates that
the current matcher does not preferentially recover the most heavily weighted
probe coordinates. Chance-corrected retention is
$(\mathrm{AUROC}_{\mathrm{transfer}}-0.5)/
(\mathrm{AUROC}_{\mathrm{same\text{-}model}}-0.5)$.

\paragraph{Coordinate-permutation control.}
We also test whether a frozen source probe depends on the ordering of the known
anchor coordinates. For every directed source--target pair among the six main
models, we derange 0\%, 25\%, 50\%, 75\%, or 100\% of the target coordinates.
Each nonzero fraction uses 20 deterministic permutations. Mean AUROC declines
from 0.812 to 0.524 on BeaverTails and from 0.816 to 0.582 on Toxic-chat as the
permuted fraction increases from 0\% to 100\%. A coordinated control applies
the same permutation to target features, scaler statistics, and probe
coefficients. It reproduces the original probabilities to numerical precision.

This control confirms that transfer uses matched anchor indexing. Its
interpretation is limited because any coordinate-sensitive classifier should
degrade when feature values are paired with the wrong fitted coefficients. We
therefore treat it as an implementation and indexing check.

\paragraph{Frame congruence and transfer.}
Finally, we ask whether a scalar measure of agreement between matched anchor
frames predicts functional transfer. This analysis contains 90
unordered model-pair, dataset, and seed conditions. Spearman correlation
between corresponding anchor-graph edges is 0.146 with mean bidirectional
AUROC, 0.208 with mean bidirectional chance-corrected retention, and 0.273 with
minimum bidirectional retention. Block-bootstrap 95\% intervals are
$[-0.055,0.256]$, $[0.048,0.261]$, and $[0.112,0.353]$, respectively.

The relationship varies by dataset. For mean retention, the correlation is
0.224 on BeaverTails and $-0.015$ on Toxic-chat. Alignment measures computed on
validation examples are also mixed. Matched-profile correlation and linear CKA
do not reliably predict transfer. A modest pooled RSA association is driven by
BeaverTails. The dataset-dependent and mostly null associations provide no
reliable label-free scalar predictor for an individual transfer. They motivate
the constructive recovery test and bound the interpretation of geometric
agreement.

%% file: content/_tables/single_tables/guard/anchor_recovery/guard_anchor_recovery_main.tex
\begin{table}[htbp]
\centering
\small
\caption{Blinded recovery of 600 anchor identities from relational geometry and the resulting Guard probe transfer. AUROC and chance-corrected retention average two datasets, three seeds, four model pairs, and both transfer directions; Random averages 20 mappings per pair condition. Values are mean $\pm$ standard error over three seed-level aggregates.}
\label{tab:guard-anchor-recovery-main}
\begin{adjustbox}{max width=\columnwidth}
\begin{tabular}{lrrrr}
\toprule
\textbf{Mapping} & \textbf{Exact (\%)} & \textbf{Weighted (\%)} & \textbf{AUROC (\%)} & \textbf{Retention (\%)} \\
\midrule
Fingerprint assignment & $4.89\pm0.05$ & $4.65\pm0.07$ & $70.6\pm0.6$ & $48.5\pm1.8$ \\
Graph matching & $20.72\pm0.29$ & $20.61\pm0.32$ & $79.0\pm1.9$ & $70.4\pm4.7$ \\
Known prompt identity & $100.00\pm0.00$ & $100.00\pm0.00$ & $83.2\pm0.7$ & $81.3\pm1.7$ \\
Random mapping & $0.16\pm0.00$ & $0.15\pm0.01$ & $54.8\pm1.0$ & $12.7\pm2.1$ \\
\bottomrule
\end{tabular}
\end{adjustbox}
\end{table}

%% file: content/_tables/single_tables/guard/anchor_recovery/guard_anchor_recovery_scope.tex
\begin{table}[htbp]
\centering
\small
\caption{Blinded graph-matching recovery by dataset and model-pair scope. Each row averages two prespecified model pairs, three seeds, and, for AUROC, both transfer directions. Known prompt identity is a correspondence control rather than a guaranteed performance upper bound. Values are percentages, mean $\pm$ standard error over three seed-level aggregates.}
\label{tab:guard-anchor-recovery-scope}
\begin{adjustbox}{max width=\textwidth}
\begin{tabular}{llrrrr}
\toprule
\textbf{Dataset} & \textbf{Pair scope} & \textbf{Graph recovery} & \textbf{Graph AUROC} & \textbf{Known-ID AUROC} & \textbf{Random AUROC} \\
\midrule
BeaverTails & Same family & $20.6\pm0.1$ & $85.2\pm1.6$ & $87.5\pm1.7$ & $51.2\pm1.7$ \\
BeaverTails & Cross family & $10.3\pm1.5$ & $78.9\pm4.8$ & $78.4\pm1.1$ & $52.7\pm1.2$ \\
\midrule
Toxic-chat & Same family & $46.8\pm0.4$ & $83.0\pm1.1$ & $82.3\pm0.8$ & $58.0\pm0.5$ \\
Toxic-chat & Cross family & $5.3\pm0.1$ & $68.8\pm1.3$ & $84.6\pm0.4$ & $57.4\pm1.2$ \\
\bottomrule
\end{tabular}
\end{adjustbox}
\end{table}

%% file: content/rebuttal_sections/guard_mlp_additional_results.tex

\subsection{MLP probes}
\label{app:guard-mlp-additional}

We next ask whether the global anchored representation also supports nonlinear
probes. We use an MLP with $64$ hidden units,
no dropout, weight decay $0.001$, no class weighting, and mean-token completion
aggregation. These MLP probes also use mean-token anchors, whereas the linear
Guard comparison in Section~\ref{sec:exp-cross-model} uses single-token anchors.
We optimize with Adam at learning rate $10^{-3}$ and batch size $64$ for at
most 50 epochs. Training stops after five epochs without improvement in
validation loss, and we restore the checkpoint with the lowest validation
loss.
Table~\ref{tab:guard-mlp-h64-baseline-auroc}
shows same-model performance for standard non-anchored probes and global
anchored probes. The global MLP remains below the linear global probe in
same-model AUROC, but retains strong performance: $92.2$ AUROC on BeaverTails
and $83.1$ on Toxic-chat. The standard MLP baseline is also below the standard
linear probe, so the MLP results should be read as a nonlinear stress test
rather than a replacement for the linear Guard probe.

\input{content/_tables/single_tables/guard/mlp/anchor_norm/agg-mean/anch600/params1_ls0.8/follow_up/baseline_guard-safety_auroc.tex}

With source-training normalization, however, cross-model MLP transfer is poor.
We therefore test whether the nonlinear probe is sensitive to differences in
feature scaling between source and target models.

The relevant implementation detail is feature normalization. Like most
probe-training pipelines, our probes normalize features before fitting the
classifier. In the standard same-model setting, and in most of our linear
transfer experiments, this normalization uses the feature mean and standard
deviation estimated on the source model's probe-training split. For cross-model
transfer, however, those source-model statistics need not match the target
model's anchored feature distribution. This mismatch is mild for the linear
Guard probes: in the spot check in
Table~\ref{tab:guard-linear-anchor-norm-spotcheck}, replacing source-train
normalization by target-anchor normalization changes cross-model AUROC by at
most about one point. For the MLP, source-normalized transfer produces saturated
scores and poor performance. We therefore evaluate a target-anchor normalization
strategy. At application time, we normalize the target model's
anchored features using the mean and standard deviation of that target model's
anchor-example features, rather than the source model's training examples.
This uses no target Guard labels and no target train/validation/test examples;
it only reuses the anchor prompts that are already required to construct the
anchored representation.

Cross-model transfer is more nuanced. Table~\ref{tab:guard-mlp-h64-cross-auroc}
shows that target-anchor-normalized naive single-source transfer reaches $76.5$
AUROC on BeaverTails and $78.0$ on Toxic-chat. On BeaverTails, this is slightly
above the linear global-probe naive transfer result ($75.0$ AUROC), but the
difference is small relative to run-to-run uncertainty; on Toxic-chat, it
remains below linear. Canonical source selection by validation AUROC gives only
modest additional gains for this MLP: Transfer~(cal) reaches $78.7$ AUROC on
BeaverTails and $78.5$ on Toxic-chat. Most of the improvement therefore comes
from changing the normalization rather than from source selection.

\input{content/_tables/single_tables/guard/mlp/anchor_norm/agg-mean/anch600/params1_ls0.8/follow_up/cross_guard-safety_auroc.tex}

At low false-positive rates, the target-anchor-normalized MLP still trails the
linear Guard probe. Transfer~(cal) reaches $46.0$ R@0.1FPR on BeaverTails and
$41.7$ on Toxic-chat. Thus target-anchor normalization substantially reduces
the source-normalized MLP failure, but it does not make the MLP competitive
with the linear Guard probe on recall
(Table~\ref{tab:guard-mlp-h64-cross-recall}).

\input{content/_tables/single_tables/guard/mlp/anchor_norm/agg-mean/anch600/params1_ls0.8/follow_up/cross_guard-safety_recall.tex}

\paragraph{Target-anchor normalization}
\label{app:guard-mlp-anchor-normalization}

\begin{figure*}[tbp]
\centering
\includegraphics[width=\textwidth]{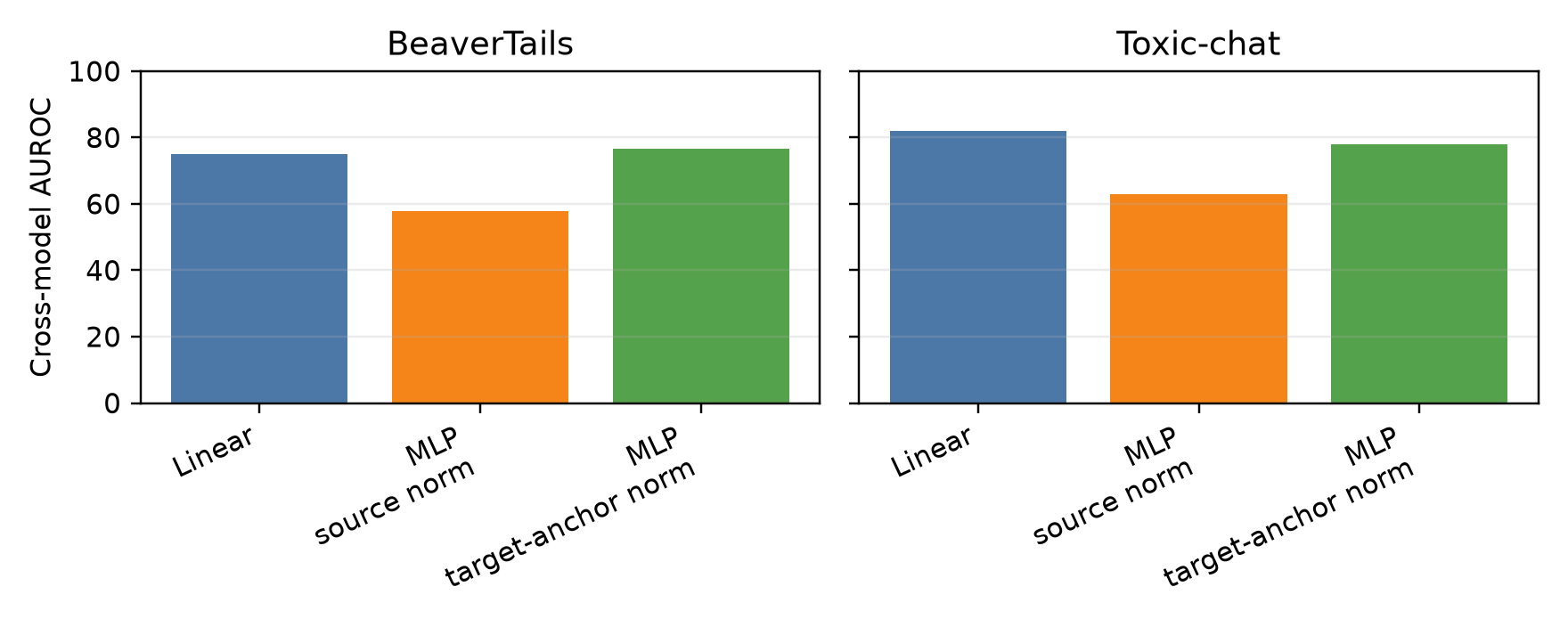}
\caption{Guard Distillation global-probe cross-model AUROC under linear and default
MLP settings with 64 hidden units. Target-anchor normalization substantially
improves the source-normalized MLP result and brings MLP transfer close to the
linear global probe.}
\label{fig:guard-mlp-anchor-calibration-auroc}
\end{figure*}

Figure~\ref{fig:guard-mlp-anchor-calibration-auroc} compares source-training and
target-anchor normalization. With source-training normalization, naive cross-model
AUROC averages only $57.7$ on BeaverTails and $62.9$ on Toxic-chat. Replacing
source normalization by target-anchor normalization raises this to $76.5$ and
$78.0$, respectively. The large gap between source-normalized and
target-anchor-normalized MLP transfer suggests that the nonlinear probe is
sensitive to cross-model differences in the scale of anchor-relative features,
rather than primarily to label calibration. Target-anchor normalization
mitigates the observed failure using only unlabeled anchor examples from the
target model.
For R@0.1FPR, target-anchor normalization also improves the MLP relative to its
source-normalized version, but only recovers it from a collapsed baseline: after
normalization, the MLP still remains below or roughly tied with the linear
global probe.

\FloatBarrier

Target-anchor normalization is much less important for linear probes in this
experiment. Table~\ref{tab:guard-linear-anchor-norm-spotcheck} shows a single-seed
Guard linear spot check: target-anchor normalization changes AUROC by at most
about one point, and it does not improve the R@0.1FPR metric in
this spot check. This supports a practical distinction: target-anchor
normalization is useful for avoiding the source-normalized MLP transfer
failure, especially in AUROC, but it is not a general recipe for improving
low-FPR recall.

\input{content/_tables/single_tables/guard/mlp/anchor_norm/agg-mean/anch600/params1_ls0.8/follow_up/linear_anchor_norm_spotcheck.tex}

\paragraph{Alternative MLP configurations}
\label{app:guard-mlp-architecture-sweep}

We sweep a small set of target-anchor-normalized MLP regularization
and architecture variants. Table~\ref{tab:guard-mlp-architecture-sweep-auroc}
includes the default MLP and four variants with weight decay $0.01$,
$32$ or $64$ hidden units, with or without dropout. For compactness, table
columns use h64/h32 for hidden size, d0.0/d0.1 for dropout probability, and
wd for weight decay. The regularized variants move naive transfer only
modestly, so we retain the default MLP with $64$ hidden units for the baseline
and normalization comparisons in this appendix. All five
target-anchor-normalized MLP variants have similar AUROC and retain the same
BeaverTails-versus-Toxic-chat ordering.

We also test class-balanced training for the $64$-hidden-unit, no-dropout
architecture.
Table~\ref{tab:guard-mlp-class-balance-diagnostic} shows that it does not help:
the default AUROC values repeat the $64$-hidden-unit, no-dropout column in
Table~\ref{tab:guard-mlp-architecture-sweep-auroc}, and class balancing does
not improve either AUROC or R@0.1FPR. We therefore use no class weighting in
every default-MLP comparison reported here.

\input{content/_tables/single_tables/guard/mlp/anchor_norm/agg-mean/anch600/params1_ls0.8/follow_up/architecture_sweep_auroc.tex}

\input{content/_tables/single_tables/guard/mlp/anchor_norm/agg-mean/anch600/params1_ls0.8/follow_up/class_balance_diagnostic.tex}

%% file: content/_tables/single_tables/guard/mlp/anchor_norm/agg-mean/anch600/params1_ls0.8/follow_up/baseline_guard-safety_auroc.tex
\begin{table}[htbp]
\centering
\caption{Same-model Guard Distillation AUROC for linear and MLP probes with mean-token anchors. The MLP has 64 hidden units, and its global variant uses target-anchor normalization.}
\label{tab:guard-mlp-h64-baseline-auroc}
\begin{adjustbox}{max width=\textwidth}
\begin{tabular}{lcccc}
\toprule
Dataset & Standard linear & Standard MLP & Global linear & Global MLP \\
\midrule
BeaverTails & $96.5\pm0.1$ & $92.5\pm2.0$ & $94.6\pm0.3$ & $92.2\pm1.2$ \\
Toxic-chat & $88.6\pm0.8$ & $83.5\pm0.4$ & $88.6\pm0.4$ & $83.1\pm0.7$ \\
\bottomrule
\end{tabular}
\end{adjustbox}
\end{table}

%% file: content/_tables/single_tables/guard/mlp/anchor_norm/agg-mean/anch600/params1_ls0.8/follow_up/cross_guard-safety_auroc.tex
\begin{table}[htbp]
\centering
\caption{Guard Distillation MLP AUROC across calibration-pool sizes, using mean-token anchors and target-anchor normalization. Transfer (cal-N) uses source-only validation within each N-model pool.}
\label{tab:guard-mlp-h64-cross-auroc}
\begin{adjustbox}{max width=\textwidth}
\begin{tabular}{lcccccc}
\toprule
Dataset & Anchored & Transfer & Transfer (cal-2) & Transfer (cal-3) & Transfer (cal-4) & Transfer (cal) \\
\midrule
BeaverTails & $\mathbf{92.2\pm1.2}$ & $76.5\pm1.5$ & $82.1\pm1.9$ & $81.2\pm2.1$ & $80.9\pm2.0$ & $78.7\pm4.1$ \\
Toxic-chat & $\mathbf{83.1\pm0.7}$ & $78.0\pm1.1$ & $78.7\pm0.5$ & $79.2\pm0.5$ & $79.1\pm0.8$ & $78.5\pm1.3$ \\
\bottomrule
\end{tabular}
\end{adjustbox}
\end{table}

%% file: content/_tables/single_tables/guard/mlp/anchor_norm/agg-mean/anch600/params1_ls0.8/follow_up/cross_guard-safety_recall.tex
\begin{table}[htbp]
\centering
\caption{Guard Distillation MLP R@0.1FPR across calibration-pool sizes, using mean-token anchors and target-anchor normalization. Transfer (cal-N) uses source-only validation within each N-model pool.}
\label{tab:guard-mlp-h64-cross-recall}
\begin{adjustbox}{max width=\textwidth}
\begin{tabular}{lcccccc}
\toprule
Dataset & Anchored & Transfer & Transfer (cal-2) & Transfer (cal-3) & Transfer (cal-4) & Transfer (cal) \\
\midrule
BeaverTails & $\mathbf{78.3\pm2.5}$ & $44.1\pm2.5$ & $50.8\pm4.2$ & $48.8\pm5.2$ & $48.5\pm5.4$ & $46.0\pm9.2$ \\
Toxic-chat & $\mathbf{51.9\pm0.5}$ & $43.4\pm2.2$ & $43.1\pm1.8$ & $43.8\pm2.3$ & $43.5\pm3.1$ & $41.7\pm4.6$ \\
\bottomrule
\end{tabular}
\end{adjustbox}
\end{table}

%% file: content/_tables/single_tables/guard/mlp/anchor_norm/agg-mean/anch600/params1_ls0.8/follow_up/linear_anchor_norm_spotcheck.tex
\begin{table}[htbp]
\centering
\caption{Single-seed Guard Distillation linear-probe comparison of source-train and target-anchor normalization under direct cross-model transfer. Each value averages 30 source-target applications.}
\label{tab:guard-linear-anchor-norm-spotcheck}
\begin{adjustbox}{max width=\textwidth}
\begin{tabular}{llcccc}
\toprule
Dataset & Anchor rep. & \multicolumn{2}{c}{AUROC} & \multicolumn{2}{c}{R@0.1FPR} \\
\cmidrule(lr){3-4}\cmidrule(lr){5-6}
 & & Source norm & Target norm & Source norm & Target norm \\
\midrule
BeaverTails & mean-token & $75.8$ & $75.6$ & $50.3$ & $46.8$ \\
BeaverTails & single-token & $83.0$ & $83.9$ & $59.2$ & $53.0$ \\
Toxic-chat & mean-token & $81.6$ & $81.9$ & $47.1$ & $47.2$ \\
Toxic-chat & single-token & $81.0$ & $81.4$ & $46.9$ & $44.2$ \\
\bottomrule
\end{tabular}
\end{adjustbox}
\end{table}

%% file: content/_tables/single_tables/guard/mlp/anchor_norm/agg-mean/anch600/params1_ls0.8/follow_up/architecture_sweep_auroc.tex
\begin{table}[htbp]
\centering
\caption{Guard Distillation cross-model AUROC for target-anchor-normalized MLP architecture and regularization variants. All rows use global anchored representations, no class weighting, and mean-token anchor aggregation. The first column is the default 64-hidden-unit MLP with no dropout; the remaining columns use weight decay 0.01 and vary hidden size and dropout.}
\label{tab:guard-mlp-architecture-sweep-auroc}
\begin{adjustbox}{max width=\textwidth}
\begin{tabular}{lccccc}
\toprule
Dataset & h64 d0.0 wd0.001 & h64 d0.0 wd0.01 & h64 d0.1 wd0.01 & h32 d0.0 wd0.01 & h32 d0.1 wd0.01 \\
\midrule
BeaverTails & $76.5\pm1.5$ & $77.9\pm1.7$ & $78.2\pm1.8$ & $78.6\pm0.6$ & $78.3\pm1.1$ \\
Toxic-chat & $78.0\pm1.1$ & $78.4\pm1.0$ & $78.5\pm1.1$ & $77.5\pm0.8$ & $78.5\pm1.1$ \\
\bottomrule
\end{tabular}
\end{adjustbox}
\end{table}

%% file: content/_tables/single_tables/guard/mlp/anchor_norm/agg-mean/anch600/params1_ls0.8/follow_up/class_balance_diagnostic.tex
\begin{table}[htbp]
\centering
\caption{Class-balancing diagnostic for the default global anchored MLP with 64 hidden units and target-anchor normalization. Transfer averages off-diagonal source--target applications.}
\label{tab:guard-mlp-class-balance-diagnostic}
\begin{adjustbox}{max width=\textwidth}
\begin{tabular}{lcccc}
\toprule
Dataset & AUROC default & AUROC class-balanced & R@0.1FPR default & R@0.1FPR class-balanced \\
\midrule
BeaverTails & $76.5\pm1.5$ & $74.0\pm1.3$ & $44.1\pm2.5$ & $39.8\pm1.5$ \\
Toxic-chat & $78.0\pm1.1$ & $77.7\pm0.6$ & $43.4\pm2.2$ & $40.9\pm0.9$ \\
\bottomrule
\end{tabular}
\end{adjustbox}
\end{table}

%% file: content/rebuttal_sections/guard_linear_standard_cross_model.tex
\subsection{Standard cross-model probe baseline}
\label{app:guard-linear-standard-cross-model-baseline}

We apply Guard Distillation linear probes directly between Llama-3.1-8B and
Mistral-7B, the two models with matching hidden size and layer count.
Table~\ref{tab:guard-linear-standard-cross-model-auroc} shows
that direct standard-probe transfer at relative depth~$0.8$ is near chance and
averages $49.7$ AUROC across the two Guard datasets, both directions, and three
seeds. Thus, even for a hidden-size- and layer-count-matched model pair,
raw-coordinate Guard Distillation probes do not reliably transfer across models.

\begin{figure*}[tbp]
\centering
\includegraphics[width=0.82\textwidth]{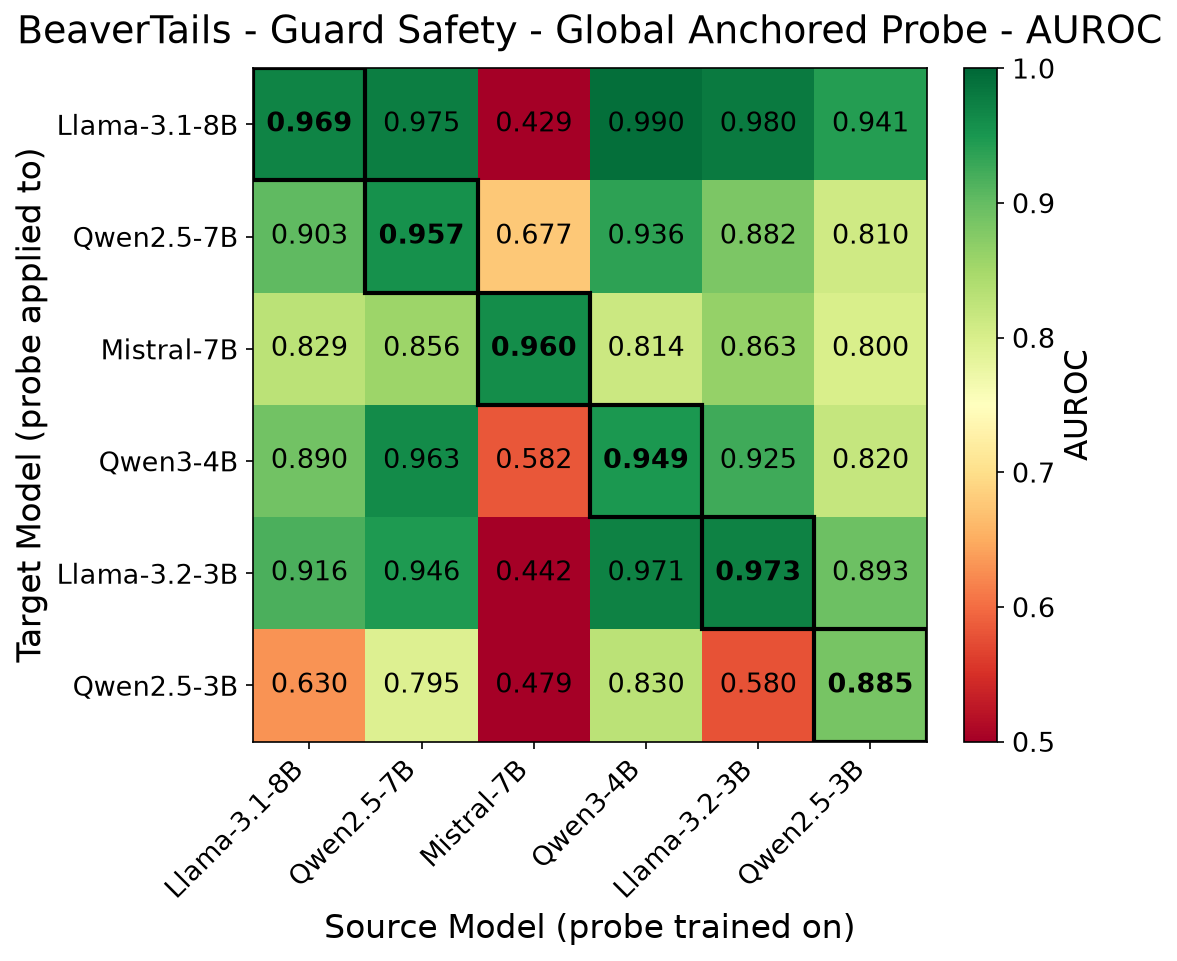}
\caption{Guard Distillation BeaverTails global anchored cross-model AUROC with
single-token anchors. Rows are target models and columns are source models.}
\label{fig:guard-linear-single-confusion-beavertails-auroc}
\end{figure*}

Table~\ref{tab:guard-linear-anchored-cross-model-auroc} reports global anchored
linear probes for the same model pair. Anchoring gives large gains on three of the four
cells, although BeaverTails Mistral$\to$Llama remains weak ($42.9$ AUROC).
Figure~\ref{fig:guard-linear-single-confusion-beavertails-auroc} shows that
this is part of a broader directed pattern: on BeaverTails, the Mistral-source
column is weak for most non-self targets, while several other source columns
transfer well. Mistral's BeaverTails completions also have the highest unsafe
rate ($12.2\%$, versus $2.0$--$6.6\%$ for other monitored models;
Table~\ref{tab:guard-guard-label-balance}). This coincident distributional
difference does not explain the AUROC gap by itself, because AUROC is
insensitive to class prevalence.

\input{content/_tables/single_tables/guard/linear/standard_cross_model/standard_cross_model_auroc.tex}

\input{content/_tables/single_tables/guard/linear/standard_cross_model/anchored_cross_model_auroc.tex}

\FloatBarrier

%% file: content/_tables/single_tables/guard/linear/standard_cross_model/standard_cross_model_auroc.tex
\begin{table}[htbp]
\centering
\caption{Standard cross-model linear-probe AUROC for Guard Distillation between Llama-3.1-8B-Instruct and Mistral-7B-Instruct-v0.3. Both models have 32 layers and 4096-dimensional residual streams; probes use layer 25 ($0.8$ relative depth). Values are means $\pm$ standard errors over three seeds.}
\label{tab:guard-linear-standard-cross-model-auroc}
\begin{adjustbox}{max width=\textwidth}
\begin{tabular}{lcc}
\toprule
Dataset & Llama$\to$Mistral & Mistral$\to$Llama \\
\midrule
BeaverTails & $55.8\pm4.4$ & $48.7\pm11.4$ \\
Toxic-chat & $49.7\pm3.6$ & $44.4\pm3.3$ \\
\bottomrule
\end{tabular}
\end{adjustbox}
\end{table}

%% file: content/_tables/single_tables/guard/linear/standard_cross_model/anchored_cross_model_auroc.tex
\begin{table}[htbp]
\centering
\caption{Global anchored linear-probe AUROC for Guard Distillation between Llama-3.1-8B-Instruct and Mistral-7B-Instruct-v0.3 ($600$ single-token anchors). Values are means $\pm$ standard errors over three seeds.}
\label{tab:guard-linear-anchored-cross-model-auroc}
\begin{adjustbox}{max width=\textwidth}
\begin{tabular}{lcc}
\toprule
Dataset & Llama$\to$Mistral & Mistral$\to$Llama \\
\midrule
BeaverTails & $82.9\pm0.6$ & $42.9\pm3.5$ \\
Toxic-chat & $89.5\pm1.6$ & $82.4\pm1.0$ \\
\bottomrule
\end{tabular}
\end{adjustbox}
\end{table}

%% file: content/_tables/single_tables/guard/linear/abs/ls0.8/abs_guard-safety_auroc.tex
\begin{table}[htbp]
\centering
\caption{Linear-probe AUROC for Guard Distillation: same-model Standard probes at relative depth 0.8 and on concatenated all-layer activations (C=0.1).}
\label{tab:final_guard_linear_abs_guard-safety_auroc}
\begin{adjustbox}{max width=\textwidth}
\begin{tabular}{lcc}
\toprule
Dataset & Standard & All-layer Standard \\
\midrule
BeaverTails & $96.5\pm0.1$ & $96.1\pm0.5$ \\
Toxic-chat & $88.6\pm0.8$ & $89.2\pm0.8$ \\
\bottomrule
\end{tabular}
\end{adjustbox}
\end{table}

%% file: content/_tables/single_tables/guard/mlp/abs/ls0.8/mlp_abs_guard-safety_auroc.tex
\begin{table}[htbp]
\centering
\caption{Guard MLP AUROC for Guard Distillation: same-model standard probes at relative depth 0.8 and on concatenated all-layer activations, with hidden sizes 64 and 256.}
\label{tab:final_guard_mlp_abs_fused_guard-safety_auroc}
\begin{adjustbox}{max width=\textwidth}
\begin{tabular}{lcc}
\toprule
Dataset & Single-layer MLP-64 & All-layer MLP-64 \\
\midrule
BeaverTails & $92.5\pm2.0$ & $84.7\pm1.0$ \\
Toxic-chat & $83.5\pm0.4$ & $72.9\pm1.5$ \\
\bottomrule
\end{tabular}
\end{adjustbox}
\end{table}

%% file: content/_tables/single_tables/guard/linear/model_family/guard_family_contrasts_auroc.tex
\begin{table}[htbp]
\centering
\caption{Guard linear model-group transfer contrasts. Values are single-source transfer AUROC in percentage points, mean $\pm$ standard error over dataset--seed aggregates. ``Same'' and ``different'' compare directed source-model to target-model pairs under the grouping in the first column.}
\label{tab:app-guard-family-contrasts}
\begin{adjustbox}{max width=0.72\textwidth}
\begin{tabular}{lccc}
\toprule
Grouping & Same & Different & $\Delta$ (Guard Distillation) \\
\midrule
Exact family & $84.9\pm1.4$ & $80.8\pm0.7$ & $4.1\pm1.6$ \\
Llama+Mistral/Qwen & $81.2\pm0.9$ & $81.5\pm0.7$ & $-0.3\pm1.1$ \\
Size bucket & $81.4\pm0.8$ & $81.4\pm0.6$ & $-0.0\pm0.5$ \\
\bottomrule
\end{tabular}
\end{adjustbox}
\end{table}

%% file: content/_tables/single_tables/guard/linear/model_family/guard_source_target_asymmetry_auroc.tex
\begin{table}[htbp]
\centering
\caption{Guard linear source-model/target-model asymmetry. Values are single-source transfer AUROC in percentage points, mean $\pm$ standard error over dataset--seed aggregates. ``As target'' is the mean AUROC of probes trained on other models and evaluated on that model. ``As source'' is the mean AUROC of that model's probe evaluated on other models. ``Gap'' is same-model Anchored AUROC minus ``As target.''}
\label{tab:app-guard-source-target-asymmetry}
\begin{adjustbox}{max width=0.72\textwidth}
\begin{tabular}{lccc}
\toprule
Model & As target & Gap & As source (Guard Distillation) \\
\midrule
Qwen2.5-3B & $71.3\pm3.7$ & $16.7\pm4.2$ & $84.5\pm1.5$ \\
Mistral-7B & $84.4\pm0.7$ & $9.3\pm1.6$ & $66.4\pm6.4$ \\
Llama-3.2-3B & $84.5\pm0.7$ & $8.8\pm2.4$ & $82.1\pm1.4$ \\
Llama-3.1-8B & $84.4\pm0.9$ & $7.4\pm1.6$ & $82.6\pm0.9$ \\
Qwen2.5-7B & $81.6\pm1.5$ & $8.7\pm1.6$ & $87.6\pm1.7$ \\
Qwen3-4B & $82.0\pm0.9$ & $8.0\pm2.0$ & $85.0\pm2.7$ \\
\bottomrule
\end{tabular}
\end{adjustbox}
\end{table}

%% file: content/appendix/math_results.tex
\section{Online mathematical correctness results}
\label{app:math-online-results}

This section reports calibration and depth-selection results for online MATH
monitoring in Section~\ref{app:math-online-depth-policy}, low-FPR recall in
Section~\ref{app:math-online-recall}, the external GPT-OSS target in
Section~\ref{app:external-target-math}, and related-data transfer to GSM8K in
Section~\ref{app:math-online-crossdata}.

\subsection{Calibration and depth selection}
\label{app:math-online-depth-policy}

Section~\ref{sec:exp-cross-model} reports the primary 2K result at 128 tokens.
Table~\ref{tab:math-online-checkpoints} extends this comparison across all
three generation checkpoints and available source-pool sizes. Global transfer
improves as the source pool grows at every checkpoint. At 128 tokens, mean
single-source transfer reaches 75.2 AUROC, calibration among two available
sources reaches 76.8, and calibration among all three reaches 78.0.

\begin{figure*}[tbp]
\centering
\includegraphics[width=\textwidth]{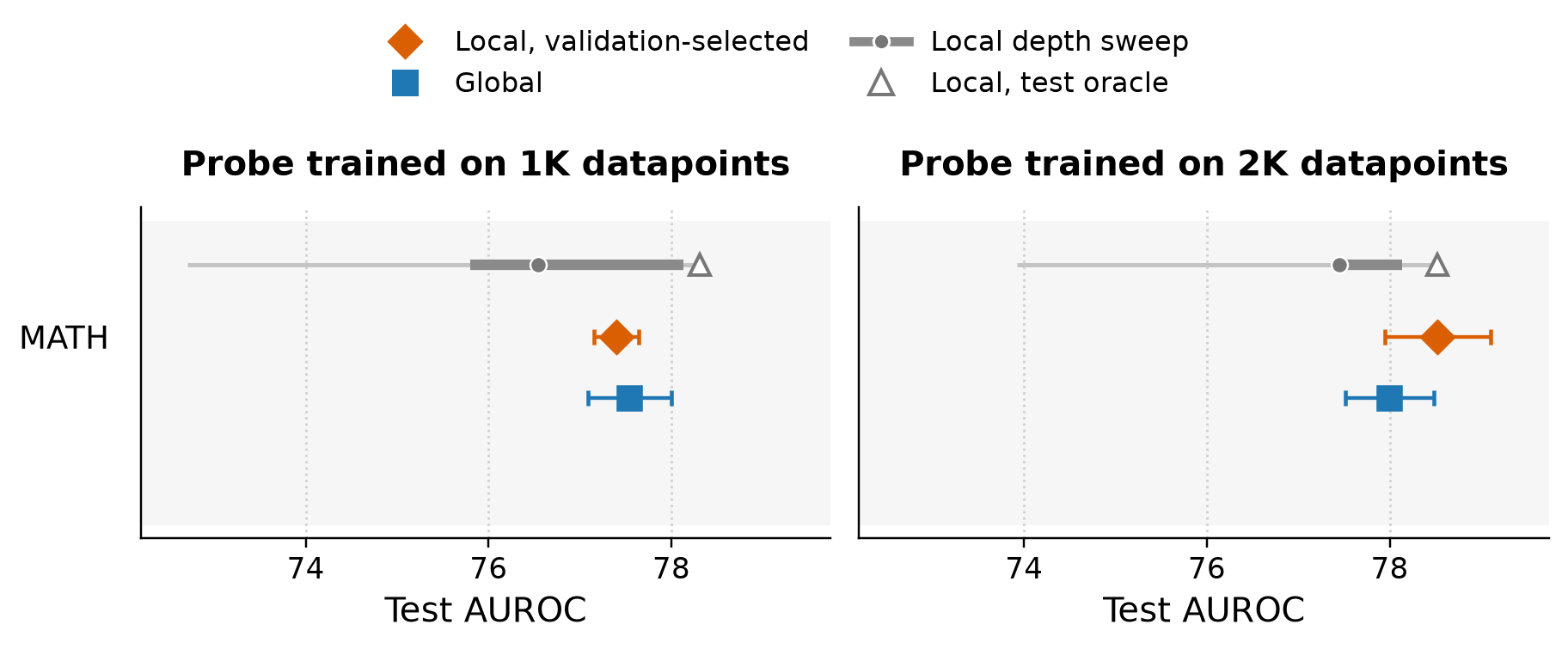}
\caption{\textbf{Online MATH correctness monitoring: calibrated global versus
local transfer.} Orange diamonds select both source and local depth using
source-only validation; blue squares show calibrated global transfer. Gray
segments summarize test AUROC across local depths (range, interquartile range,
and mean), while hollow triangles select depth post hoc using test AUROC but
retain validation-based source selection. Error bars are standard errors over
three source/probe seeds.}
\label{fig:math-online-depth-policy}
\end{figure*}

We also ask whether validation can identify a useful local depth.
Figure~\ref{fig:math-online-depth-policy} compares the global probe with local
probes on Llama-3.1-8B, Qwen3-4B, OLMo-3-7B, and Gemma-3-12B at the 128-token
checkpoint. The two panels use nested 1K and 2K source-training sets. At each
local depth, source-pool validation selects a source model; the orange diamond
jointly selects one source--depth pair using the same validation data. Neither
deployable policy uses MATH-500 test labels. The selected local and global
policies are close at both training sizes, extending the layer-selection result
in Section~\ref{sec:exp-global-vs-local}: the MATH comparison does not provide
a consistent reason to prefer local depth selection.

\input{content/_tables/single_tables/math/math_online_checkpoints.tex}

\FloatBarrier

\FloatBarrier

\subsection{Low-FPR recall}
\label{app:math-online-recall}
Table~\ref{tab:math-online-crossdata-recall} reports R@0.1FPR for the
128-token MATH-500 and GSM8K comparisons. On MATH-500, global anchoring raises
same-model recall from $40.7$ to $49.1$, while calibrated global transfer
reaches $38.7$ and validation-selected local transfer reaches $44.7$. Under the
combined model-and-data shift to GSM8K, calibrated global transfer reaches
$30.8$, compared with $18.3$ for the same-model Standard probe. Low-FPR recall
therefore follows the main AUROC pattern: transfer trails the same-model
Anchored probe but remains competitive with the Standard baseline.

\input{content/_tables/single_tables/math/math_online_crossdata_recall.tex}

\FloatBarrier

\subsection{Transfer to an external model target}
\label{app:external-target-math}

Section~\ref{sec:exp-external-families} summarizes transfer to the external
GPT-OSS MATH target. Table~\ref{tab:external-math-full} reports the
post-hoc target-trained diagnostic alongside calibrated transfer. Correctness
is linearly decodable in GPT-OSS at both source-training sizes, whereas the
source-trained decision rule remains near chance. This comparison diagnoses the
MATH transfer failure; target-trained probes do not enter source selection or
the transfer procedure.

\input{content/_tables/single_tables/external_targets/external_math_full.tex}

\FloatBarrier

\subsection{Related-data transfer}
\label{app:math-online-crossdata}
Section~\ref{sec:exp-related-data} summarizes joint transfer from MATH to
GSM8K. We evaluate the MATH-trained probes on GSM8K at the 128-token
checkpoint under the protocol in
Appendix~\ref{app:related-data-implementation}.
Table~\ref{tab:math-online-crossdata} separates the
effect of changing the data from the additional effect of changing the model.
Within the same model, global anchored probes reach 75.3 AUROC, compared with
63.0 for standard probes. Calibrated transfer across both model and dataset
reaches 70.5.

\input{content/_tables/single_tables/math/math_online_crossdata.tex}

\FloatBarrier

%% file: content/_tables/single_tables/math/math_online_checkpoints.tex
\begin{table}[htbp]
\centering
\small
\caption{\textbf{Online MATH transfer across generation checkpoints and source-pool sizes.} Probes use 2K training examples. Transfer (cal-$N$) uses source-only validation within each $N$-model pool. Values are AUROC (\%), mean $\pm$ standard error over three source/probe seeds.}
\label{tab:math-online-checkpoints}
\begin{tabular}{lccc}
\toprule
Method & 64 tokens & 128 tokens & 256 tokens \\
\midrule
Standard same-model, depth 0.8 & 78.0 $\pm$ 0.4 & 77.9 $\pm$ 0.3 & 76.6 $\pm$ 0.5 \\
Global Anchored, same-model & 81.3 $\pm$ 0.2 & 81.8 $\pm$ 0.3 & 80.9 $\pm$ 0.2 \\
Local Anchored, depth 0.8 & 81.3 $\pm$ 0.2 & 81.6 $\pm$ 0.3 & 80.7 $\pm$ 0.2 \\
\midrule
Global transfer & 73.8 $\pm$ 1.4 & 75.2 $\pm$ 0.6 & 75.0 $\pm$ 0.9 \\
Global transfer (cal-2) & 74.4 $\pm$ 0.8 & 76.8 $\pm$ 0.1 & 76.2 $\pm$ 0.6 \\
Global transfer (cal-3) & 74.9 $\pm$ 1.7 & 78.0 $\pm$ 0.5 & 76.7 $\pm$ 0.9 \\
Validation-selected local transfer & 78.3 $\pm$ 0.7 & 78.5 $\pm$ 0.6 & 78.4 $\pm$ 0.5 \\
\bottomrule
\end{tabular}
\end{table}

%% file: content/_tables/single_tables/math/math_online_crossdata_recall.tex
\begin{table}[htbp]
\centering
\small
\caption{\textbf{Low-FPR online correctness monitoring.} R@0.1FPR on MATH-500 and after transfer from MATH to GSM8K at the 128-token checkpoint. The positive class is a correct solution. Values are percentage points, mean $\pm$ standard error over three source/probe seeds.}
\label{tab:math-online-crossdata-recall}
\begin{tabular}{llc}
\toprule
Evaluation & Method & R@0.1FPR \\
\midrule
MATH-500 & Standard same-model, depth 0.8 & 40.7 $\pm$ 2.8 \\
MATH-500 & Global Anchored, same-model & 49.1 $\pm$ 1.0 \\
MATH-500 & Local Anchored, depth 0.8 & 48.7 $\pm$ 1.1 \\
MATH-500 & Global calibrated transfer & 38.7 $\pm$ 1.3 \\
MATH-500 & Validation-selected local transfer & 44.7 $\pm$ 3.5 \\
\midrule
GSM8K & Standard same-model, depth 0.8 & 18.3 $\pm$ 1.5 \\
GSM8K & Global Anchored, same-model & 36.6 $\pm$ 2.3 \\
GSM8K & Global calibrated transfer & 30.8 $\pm$ 3.3 \\
\bottomrule
\end{tabular}
\end{table}

%% file: content/_tables/single_tables/external_targets/external_math_full.tex
\begin{table}[htbp]
\centering
\caption{\textbf{External MATH target: GPT-OSS-20B.} Target-trained Anchored probes are post-hoc diagnostics and do not select transferred probes. Values are AUROC (\%), mean $\pm$ standard error over three source/probe seeds.}
\label{tab:external-math-full}
\begin{tabular}{lcc}
\toprule
Train size & Target-trained & Transfer (cal) \\
\midrule
1K & 78.5 $\pm$ 0.1 & 53.1 $\pm$ 0.7 \\
2K & 78.7 $\pm$ 0.3 & 52.4 $\pm$ 0.5 \\
\bottomrule
\end{tabular}
\end{table}

%% file: content/_tables/single_tables/math/math_online_crossdata.tex
\begin{table}[htbp]
\centering
\caption{\textbf{Related-data transfer from MATH to GSM8K.} Probes are evaluated on GSM8K at 128 generated tokens. Calibrated transfer changes both the model and dataset. Values are AUROC (\%), mean $\pm$ standard error over three source/probe seeds.}
\label{tab:math-online-crossdata}
\begin{tabular}{lc}
\toprule
Method & AUROC \\
\midrule
Standard same-model, depth 0.8 & 63.0 $\pm$ 1.0 \\
Global Anchored, same-model & 75.3 $\pm$ 0.8 \\
Global calibrated transfer & 70.5 $\pm$ 1.8 \\
\bottomrule
\end{tabular}
\end{table}

%% file: content/rebuttal_sections/xstest_staleness_main.tex
\subsection{Staleness after model updates}
\label{app:xstest-staleness-main}

Activation monitors are often trained once and then kept fixed while the model
being monitored is updated. This creates a staleness risk: the updated model may
still encode the relevant safety distinction, but the old probe may no longer
read it out reliably. We test whether anchored representations reduce this failure mode.
Following \citet{duan2026safetymonitorsstayreliable}, we study the
Qwen2.5-3B-Instruct and Qwen2.5-7B-Instruct models used in their staleness
experiments. We train standard and anchored probes to predict whether an
XSTest prompt is unsafe \citep{rottger2023xstest}; the dataset contains unsafe
prompts and safe prompts that resemble unsafe prompts. We fit the probes on
the original model checkpoints, freeze them, and evaluate them after either
quantization-only changes or small refusal-SFT updates based on BeaverTails.
The global anchored probe concatenates the six layers monitored by
\citet{duan2026safetymonitorsstayreliable} before applying the anchor transform.
Section~\ref{sec:exp-model-updates} summarizes the main result; full
implementation details are in Appendix~\ref{app:xstest-staleness-impl}.

Table~\ref{tab:xstest-indra-quantization-base-anchors} isolates quantized
inference before considering learned adapters. On Qwen2.5-7B, quantization
changes all mean AUROCs by less than half a point. On Qwen2.5-3B, standard
probes degrade more visibly, especially under NF4, but the shift is
concentrated in one monitored layer. Global anchored probes remove almost all
of this quantization sensitivity: their mean drop is at most $0.4$ points and no
global probe crosses the stale threshold.

\input{content/_tables/single_tables/staleness_experiments/xstest/indra/quantization_staleness_base_anchors.tex}

Learned updates create much larger shifts
(Table~\ref{tab:xstest-indra-learned-base-anchors}). On Qwen2.5-3B, standard
probes lose roughly $8$--$10$ AUROC points across QLoRA and LoRA updates. On
Qwen2.5-7B, QLoRA is milder but still reaches the stale threshold at 512
examples, while the 256-example full-LoRA update drops standard probes by $8.1$
points. Like \citet{duan2026safetymonitorsstayreliable}, we find that learned 
safety fine-tuning affects probe performance much more than quantization alone.

\input{content/_tables/single_tables/staleness_experiments/xstest/indra/learned_update_staleness_base_anchors.tex}

Global anchored probes provide a strong mitigation for the performance drop. For
Qwen2.5-7B, no global probe is stale in any updated setting,
and the mean drop remains between $0.1$ and $1.2$ points. For Qwen2.5-3B,
global anchored probes improve substantially over standard probes: their mean
drop stays below the $5$-point stale magnitude in every setting. Local anchored
probes also improve over standard probes on 3B, but worsen staleness on 7B.

Overall, standard probes show clear probe staleness after refusal
fine-tuning, while quantized inference alone does not produce a comparable drop.
Global anchored representations substantially reduce staleness under the tested
LoRA and QLoRA updates.

%% file: content/_tables/single_tables/staleness_experiments/xstest/indra/quantization_staleness_base_anchors.tex
\begin{table}[htbp]
\centering
\caption{XSTest probe-staleness change under quantization-only controls. Values are $\Delta$AUROC points using angular distance for anchored probes. Standard and local anchored values are first averaged over layers 4, 8, 12, 16, 20, and 24; global anchored concatenates those six layers. Rows average over three train/validation splits (mean $\pm$ SE).}
\label{tab:xstest-indra-quantization-base-anchors}
\begin{adjustbox}{max width=\textwidth}
\begin{tabular}{llccc}
\toprule
Model & Quantization & Standard & Global anchored & Local anchored \\
\midrule
Qwen2.5-3B & NF4 & $-3.6\pm0.8$ & $-0.3\pm0.1$ & $-2.6\pm0.4$ \\
Qwen2.5-3B & INT8 & $-2.3\pm0.1$ & $-0.3\pm0.0$ & $-2.3\pm0.2$ \\
Qwen2.5-7B & NF4 & $-0.2\pm0.1$ & $+0.1\pm0.0$ & $-0.4\pm0.0$ \\
Qwen2.5-7B & INT8 & $-0.3\pm0.1$ & $-0.1\pm0.0$ & $-0.2\pm0.1$ \\
\bottomrule
\end{tabular}
\end{adjustbox}
\end{table}

%% file: content/_tables/single_tables/staleness_experiments/xstest/indra/learned_update_staleness_base_anchors.tex
\begin{table}[htbp]
\centering
\caption{XSTest probe-staleness change after learned refusal-SFT updates. Values are $\Delta$AUROC points using angular distance for anchored probes. Standard and local anchored values are first averaged over layers 4, 8, 12, 16, 20, and 24; global anchored concatenates those six layers. Rows average over nine runs: three update seeds and three train/validation splits (mean $\pm$ SE).}
\label{tab:xstest-indra-learned-base-anchors}
\begin{adjustbox}{max width=\textwidth}
\begin{tabular}{lllccc}
\toprule
Model & Update & Examples & Standard & Global anchored & Local anchored \\
\midrule
Qwen2.5-3B & QLoRA & 256 & $-9.3\pm0.9$ & $-4.2\pm1.0$ & $-7.4\pm1.0$ \\
Qwen2.5-3B & QLoRA & 512 & $-9.8\pm1.0$ & $-3.3\pm0.7$ & $-7.1\pm0.9$ \\
Qwen2.5-3B & LoRA & 256 & $-9.7\pm0.5$ & $-3.3\pm0.6$ & $-6.8\pm0.9$ \\
Qwen2.5-3B & LoRA & 512 & $-8.0\pm1.5$ & $-3.6\pm0.9$ & $-4.0\pm0.5$ \\
Qwen2.5-7B & QLoRA & 256 & $-4.8\pm0.4$ & $-0.1\pm0.2$ & $-6.7\pm0.9$ \\
Qwen2.5-7B & QLoRA & 512 & $-5.8\pm1.1$ & $-1.2\pm0.3$ & $-7.5\pm0.7$ \\
Qwen2.5-7B & LoRA & 256 & $-8.1\pm1.2$ & $-0.3\pm0.2$ & $-13.0\pm1.7$ \\
Qwen2.5-7B & LoRA & 512 & $-5.1\pm0.6$ & $-0.4\pm0.2$ & $-6.5\pm1.1$ \\
\bottomrule
\end{tabular}
\end{adjustbox}
\end{table}

%% file: content/rebuttal_sections/xstest_staleness_additional.tex
\subsection{Relative-coordinate function}
\label{app:xstest-similarity-function}

The XSTest result summarized in Section~\ref{sec:exp-model-updates} uses
angular distance for anchored representations. We ablate the function used to
construct anchored features by comparing angular distance to cosine similarity.
Table~\ref{tab:xstest-baseline-similarity-ablation} compares the pre-update
probe scores. The two choices are close before any update, but
angular distance is slightly better for local anchored probes on both models and
slightly better for the 3B global anchored probe. In either case, global
anchored probes match or exceed the best layer-averaged standard result before
model updates.

\input{content/_tables/single_tables/staleness_experiments/xstest/similarity_function/baseline_similarity_ablation.tex}

\paragraph{Cosine-similarity staleness}
Cosine-similarity probes reproduce the angular-distance ordering under
quantization-only changes (Table~\ref{tab:xstest-cosine-quantization-base-anchors}):
7B changes little, 3B standard probes show a modest drop, and global
anchored probes remove nearly all of the quantization sensitivity.

\input{content/_tables/single_tables/staleness_experiments/xstest/cosine/quantization_staleness_base_anchors.tex}

Under learned updates (Table~\ref{tab:xstest-cosine-learned-base-anchors}),
cosine similarity again shows that global anchored probes mitigate probe
staleness, while local anchored probes are model-dependent. Cosine global
anchored probes are less robust than their angular-distance counterparts. On 3B,
cosine global drops remain around
$6$--$8$ points, whereas angular-distance global drops are around $3$--$4$
points. On 7B, cosine global probes remain stable relative to standard probes,
but their drops are also larger than with angular distance. Cosine similarity
therefore still mitigates global-probe staleness, but angular
distance is more robust for this task.

\input{content/_tables/single_tables/staleness_experiments/xstest/cosine/learned_update_staleness_base_anchors.tex}

\subsection{Re-encoding anchors after model updates}
\label{app:xstest-updated-anchor-encodings}

The angular-distance and cosine results above keep each original model's anchor
encodings fixed after the monitored model changes. A natural alternative is to
re-encode the same anchor prompts
with the updated model before computing anchored representations. This could
help if the model update moves anchor prompts and evaluation prompts in a
coherent way: similarities between updated evaluation prompts and updated
anchors might then resemble the original-model similarity coordinates better
than similarities between updated evaluation prompts and original-model anchors.
Under this cancellation hypothesis, updated-anchor encodings should be more
stable than original-anchor encodings.

Table~\ref{tab:xstest-indra-learned-updated-anchors} shows a mixed result for
angular distance. Updated anchors often help local anchored probes, especially on
Qwen2.5-7B, but these local probes still do not match the robustness of
global anchored probes. For global anchored probes, updated anchors are close to
original anchors in some rows, but they are worse in several 512-example
settings. The hypothesized cancellation effect is therefore partially visible for
local anchored features, but it is not the driver of the strongest mitigation.

\input{content/_tables/single_tables/staleness_experiments/xstest/indra/learned_update_staleness_updated_anchors.tex}

Cosine similarity also fails to make updated-anchor encoding consistently
beneficial (Table~\ref{tab:xstest-cosine-learned-updated-anchors}). Updated anchors
sometimes help local anchored probes, especially for 3B LoRA, but they also hurt
local probes under QLoRA and do not fix the poor 7B local results. For global
anchored probes, updated anchors are close to original anchors in some rows, but
they are substantially worse in several settings, including the 512-example 3B
updates. Across both relative-coordinate functions, re-encoding anchors after the model
update is therefore not a reliable replacement for the fixed anchor coordinate
system used by probes trained before the model update.

\input{content/_tables/single_tables/staleness_experiments/xstest/cosine/learned_update_staleness_updated_anchors.tex}

%% file: content/_tables/single_tables/staleness_experiments/xstest/similarity_function/baseline_similarity_ablation.tex
\begin{table}[htbp]
\centering
\caption{Pre-update XSTest validation AUROC for standard probes and anchored probes using two relative-coordinate functions. Unqualified anchored columns use angular distance; cosine columns use cosine similarity. Standard and local anchored columns average over layers 4, 8, 12, 16, 20, and 24; global anchored concatenates those six layers. Values are AUROC points, averaged over three train/validation splits (mean $\pm$ SE).}
\label{tab:xstest-baseline-similarity-ablation}
\begin{adjustbox}{max width=\textwidth}
\begin{tabular}{lccccc}
\toprule
Model & Standard & Global anchored & Global anchored (cosine) & Local anchored & Local anchored (cosine) \\
\midrule
Qwen2.5-3B & $94.6\pm0.4$ & $99.3\pm0.3$ & $99.1\pm0.3$ & $93.8\pm0.6$ & $92.3\pm0.7$ \\
Qwen2.5-7B & $97.4\pm0.5$ & $99.5\pm0.2$ & $99.5\pm0.3$ & $97.5\pm0.3$ & $97.1\pm0.3$ \\
\bottomrule
\end{tabular}
\end{adjustbox}
\end{table}

%% file: content/_tables/single_tables/staleness_experiments/xstest/cosine/quantization_staleness_base_anchors.tex
\begin{table}[htbp]
\centering
\caption{XSTest probe-staleness change under quantization-only controls. Values are $\Delta$AUROC points using cosine similarity for anchored probes. Standard and local anchored values are first averaged over layers 4, 8, 12, 16, 20, and 24; global anchored concatenates those six layers. Rows average over three train/validation splits (mean $\pm$ SE).}
\label{tab:xstest-cosine-quantization-base-anchors}
\begin{adjustbox}{max width=\textwidth}
\begin{tabular}{llccc}
\toprule
Model & Quantization & Standard & Global anchored & Local anchored \\
\midrule
Qwen2.5-3B & NF4 & $-3.6\pm0.8$ & $-0.2\pm0.1$ & $-1.7\pm0.1$ \\
Qwen2.5-3B & INT8 & $-2.3\pm0.1$ & $-0.4\pm0.1$ & $-1.7\pm0.2$ \\
Qwen2.5-7B & NF4 & $-0.2\pm0.1$ & $+0.1\pm0.0$ & $-0.1\pm0.1$ \\
Qwen2.5-7B & INT8 & $-0.3\pm0.1$ & $-0.1\pm0.1$ & $-0.2\pm0.1$ \\
\bottomrule
\end{tabular}
\end{adjustbox}
\end{table}

%% file: content/_tables/single_tables/staleness_experiments/xstest/cosine/learned_update_staleness_base_anchors.tex
\begin{table}[htbp]
\centering
\caption{XSTest probe-staleness change after learned refusal-SFT updates. Values are $\Delta$AUROC points using cosine similarity for anchored probes. Standard and local anchored values are first averaged over layers 4, 8, 12, 16, 20, and 24; global anchored concatenates those six layers. Rows average over nine runs: three update seeds and three train/validation splits (mean $\pm$ SE).}
\label{tab:xstest-cosine-learned-base-anchors}
\begin{adjustbox}{max width=\textwidth}
\begin{tabular}{lllccc}
\toprule
Model & Update & Examples & Standard & Global anchored & Local anchored \\
\midrule
Qwen2.5-3B & QLoRA & 256 & $-9.3\pm0.9$ & $-7.7\pm1.7$ & $-8.3\pm0.9$ \\
Qwen2.5-3B & QLoRA & 512 & $-9.8\pm1.0$ & $-6.3\pm1.3$ & $-7.7\pm0.7$ \\
Qwen2.5-3B & LoRA & 256 & $-9.7\pm0.5$ & $-6.3\pm1.3$ & $-7.4\pm0.6$ \\
Qwen2.5-3B & LoRA & 512 & $-8.0\pm1.5$ & $-6.9\pm1.5$ & $-5.2\pm0.3$ \\
Qwen2.5-7B & QLoRA & 256 & $-4.8\pm0.4$ & $-0.5\pm0.3$ & $-7.2\pm1.0$ \\
Qwen2.5-7B & QLoRA & 512 & $-5.8\pm1.1$ & $-2.4\pm0.6$ & $-7.6\pm0.5$ \\
Qwen2.5-7B & LoRA & 256 & $-8.1\pm1.2$ & $-0.7\pm0.3$ & $-13.4\pm1.8$ \\
Qwen2.5-7B & LoRA & 512 & $-5.1\pm0.6$ & $-1.4\pm0.3$ & $-8.1\pm1.3$ \\
\bottomrule
\end{tabular}
\end{adjustbox}
\end{table}

%% file: content/_tables/single_tables/staleness_experiments/xstest/indra/learned_update_staleness_updated_anchors.tex
\begin{table}[htbp]
\centering
\caption{XSTest probe-staleness change after learned refusal-SFT updates, comparing original- and updated-anchor encodings. Values are $\Delta$AUROC points using angular distance. Local anchored values are first averaged over layers 4, 8, 12, 16, 20, and 24; global anchored concatenates those six layers. Rows average over nine runs: three update seeds and three train/validation splits (mean $\pm$ SE).}
\label{tab:xstest-indra-learned-updated-anchors}
\begin{adjustbox}{max width=\textwidth}
\begin{tabular}{lllcccc}
\toprule
Model & Update & Examples & Global, original & Global, updated & Local, original & Local, updated \\
\midrule
Qwen2.5-3B & QLoRA & 256 & $-4.2\pm1.0$ & $-3.4\pm0.8$ & $-7.4\pm1.0$ & $-6.8\pm1.0$ \\
Qwen2.5-3B & QLoRA & 512 & $-3.3\pm0.7$ & $-5.8\pm1.5$ & $-7.1\pm0.9$ & $-7.6\pm1.5$ \\
Qwen2.5-3B & LoRA & 256 & $-3.3\pm0.6$ & $-3.2\pm0.9$ & $-6.8\pm0.9$ & $-4.0\pm0.6$ \\
Qwen2.5-3B & LoRA & 512 & $-3.6\pm0.9$ & $-6.5\pm2.0$ & $-4.0\pm0.5$ & $-2.6\pm0.5$ \\
Qwen2.5-7B & QLoRA & 256 & $-0.1\pm0.2$ & $-0.2\pm0.2$ & $-6.7\pm0.9$ & $-3.4\pm0.3$ \\
Qwen2.5-7B & QLoRA & 512 & $-1.2\pm0.3$ & $-1.5\pm0.6$ & $-7.5\pm0.7$ & $-4.4\pm0.7$ \\
Qwen2.5-7B & LoRA & 256 & $-0.3\pm0.2$ & $-0.4\pm0.2$ & $-13.0\pm1.7$ & $-6.7\pm1.4$ \\
Qwen2.5-7B & LoRA & 512 & $-0.4\pm0.2$ & $-0.9\pm0.4$ & $-6.5\pm1.1$ & $-3.3\pm0.7$ \\
\bottomrule
\end{tabular}
\end{adjustbox}
\end{table}

%% file: content/_tables/single_tables/staleness_experiments/xstest/cosine/learned_update_staleness_updated_anchors.tex
\begin{table}[htbp]
\centering
\caption{XSTest probe-staleness change after learned refusal-SFT updates, comparing original- and updated-anchor encodings. Values are $\Delta$AUROC points using cosine similarity. Local anchored values are first averaged over layers 4, 8, 12, 16, 20, and 24; global anchored concatenates those six layers. Rows average over nine runs: three update seeds and three train/validation splits (mean $\pm$ SE).}
\label{tab:xstest-cosine-learned-updated-anchors}
\begin{adjustbox}{max width=\textwidth}
\begin{tabular}{lllcccc}
\toprule
Model & Update & Examples & Global, original & Global, updated & Local, original & Local, updated \\
\midrule
Qwen2.5-3B & QLoRA & 256 & $-7.7\pm1.7$ & $-7.5\pm1.6$ & $-8.3\pm0.9$ & $-8.9\pm1.2$ \\
Qwen2.5-3B & QLoRA & 512 & $-6.3\pm1.3$ & $-10.7\pm2.1$ & $-7.7\pm0.7$ & $-9.4\pm1.7$ \\
Qwen2.5-3B & LoRA & 256 & $-6.3\pm1.3$ & $-6.5\pm1.6$ & $-7.4\pm0.6$ & $-6.2\pm0.7$ \\
Qwen2.5-3B & LoRA & 512 & $-6.9\pm1.5$ & $-12.6\pm2.7$ & $-5.2\pm0.3$ & $-5.1\pm0.7$ \\
Qwen2.5-7B & QLoRA & 256 & $-0.5\pm0.3$ & $-0.3\pm0.2$ & $-7.2\pm1.0$ & $-6.9\pm0.4$ \\
Qwen2.5-7B & QLoRA & 512 & $-2.4\pm0.6$ & $-4.5\pm1.8$ & $-7.6\pm0.5$ & $-8.8\pm1.5$ \\
Qwen2.5-7B & LoRA & 256 & $-0.7\pm0.3$ & $-0.7\pm0.3$ & $-13.4\pm1.8$ & $-11.8\pm2.0$ \\
Qwen2.5-7B & LoRA & 512 & $-1.4\pm0.3$ & $-3.3\pm1.5$ & $-8.1\pm1.3$ & $-7.2\pm1.5$ \\
\bottomrule
\end{tabular}
\end{adjustbox}
\end{table}

%% file: content/_tables/single_tables/guardqa/mlp/non-abs/anch300/params1_ls0.8/baseline_auroc.tex
\begin{table}[htbp]
\centering
\caption{GuardQA MLP AUROC results for Output Prediction and GT Prediction: same-model Standard and global Anchored probes. The MLP has one 64-unit hidden layer.}
\label{tab:final_guardqa_mlp_baseline_mlp_1_concat_auroc}
\begin{adjustbox}{max width=\textwidth}
\begin{tabular}{lcccc}
\toprule
 & \multicolumn{2}{c}{Output Prediction} & \multicolumn{2}{c}{GT Prediction} \\
\cmidrule(lr){2-3} \cmidrule(lr){4-5}
Dataset & Standard & Anchored & Standard & Anchored \\
\midrule
BeaverTails & $99.6\pm0.1$ & $99.8\pm0.1$ & $95.2\pm0.3$ & $94.1\pm0.5$ \\
Toxic-chat & $99.1\pm0.2$ & $99.2\pm0.1$ & $89.1\pm0.4$ & $89.3\pm0.3$ \\
\bottomrule
\end{tabular}
\end{adjustbox}
\end{table}

%% file: content/_tables/single_tables/guardqa/mlp/non-abs/anch300/params1_ls0.8/cross_model-choice_auroc.tex
\begin{table}[htbp]
\centering
\caption{GuardQA MLP AUROC results for Output Prediction: same-model Anchored and cross-model Transfer as the available source pool grows. Transfer (cal-N) uses source-only validation within each N-model pool. The MLP has one 64-unit hidden layer.}
\label{tab:final_guardqa_model-choice_cross_mlp_1_concat_auroc}
\begin{adjustbox}{max width=\textwidth}
\begin{tabular}{lcccccc}
\toprule
Dataset & Anchored & Transfer & Transfer (cal-2) & Transfer (cal-3) & Transfer (cal-4) & Transfer (cal) \\
\midrule
BeaverTails & $99.8\pm0.1$ & $72.3\pm4.6$ & $76.2\pm6.1$ & $84.1\pm4.8$ & $84.0\pm6.1$ & $\mathbf{89.0}\pm1.1$ \\
Toxic-chat & $99.2\pm0.1$ & $87.4\pm2.7$ & $95.1\pm0.6$ & $95.2\pm1.0$ & $95.7\pm1.0$ & $\mathbf{95.8}\pm1.0$ \\
\bottomrule
\end{tabular}
\end{adjustbox}
\end{table}

%% file: content/_tables/single_tables/guardqa/mlp/non-abs/anch300/params1_ls0.8/cross_gt-prediction_auroc.tex
\begin{table}[htbp]
\centering
\caption{GuardQA MLP AUROC results for GT Prediction: same-model Anchored and cross-model Transfer as the available source pool grows. Transfer (cal-N) uses source-only validation within each N-model pool. The MLP has one 64-unit hidden layer.}
\label{tab:final_guardqa_gt-prediction_cross_mlp_1_concat_auroc}
\begin{adjustbox}{max width=\textwidth}
\begin{tabular}{lcccccc}
\toprule
Dataset & Anchored & Transfer & Transfer (cal-2) & Transfer (cal-3) & Transfer (cal-4) & Transfer (cal) \\
\midrule
BeaverTails & $94.1\pm0.5$ & $78.5\pm2.1$ & $81.8\pm1.6$ & $82.6\pm1.8$ & $83.1\pm2.2$ & $\mathbf{84.0}\pm2.7$ \\
Toxic-chat & $89.3\pm0.3$ & $76.5\pm1.2$ & $80.1\pm0.7$ & $81.5\pm1.0$ & $\mathbf{82.4}\pm1.4$ & $82.2\pm1.8$ \\
\bottomrule
\end{tabular}
\end{adjustbox}
\end{table}

%% file: content/_tables/single_tables/guardqa/mlp/non-abs/anch300/params1_ls0.8/cross_summary_recall.tex
\begin{table}[htbp]
\centering
\caption{GuardQA MLP R@0.1FPR results for Output Prediction and GT Prediction: same-model Anchored, mean cross-model Transfer, and source-calibrated Transfer (cal). The MLP has one 64-unit hidden layer.}
\label{tab:final_guardqa_mlp_cross_summary_mlp_1_concat_recall}
\begin{adjustbox}{max width=\textwidth}
\begin{tabular}{lcccccc}
\toprule
 & \multicolumn{3}{c}{Output Prediction} & \multicolumn{3}{c}{GT Prediction} \\
\cmidrule(lr){2-4} \cmidrule(lr){5-7}
Dataset & Anchored & Transfer & Transfer (cal) & Anchored & Transfer & Transfer (cal) \\
\midrule
BeaverTails & $99.5\pm0.3$ & $61.6\pm5.7$ & $76.0\pm2.4$ & $82.2\pm0.3$ & $52.5\pm2.7$ & $60.5\pm4.1$ \\
Toxic-chat & $97.8\pm0.2$ & $76.8\pm2.3$ & $90.6\pm1.2$ & $68.6\pm1.1$ & $49.7\pm1.3$ & $60.4\pm0.8$ \\
\bottomrule
\end{tabular}
\end{adjustbox}
\end{table}

%% file: content/_tables/single_tables/guardqa/mlp/non-abs/anch300/params3_ls0.8/cross_summary_auroc.tex
\begin{table}[htbp]
\centering
\caption{GuardQA MLP AUROC results for Output Prediction and GT Prediction: same-model Anchored, mean cross-model Transfer, and source-calibrated Transfer (cal). The MLP has one 256-unit hidden layer.}
\label{tab:final_guardqa_mlp_cross_summary_mlp_3_concat_auroc}
\begin{adjustbox}{max width=\textwidth}
\begin{tabular}{lcccccc}
\toprule
 & \multicolumn{3}{c}{Output Prediction} & \multicolumn{3}{c}{GT Prediction} \\
\cmidrule(lr){2-4} \cmidrule(lr){5-7}
Dataset & Anchored & Transfer & Transfer (cal) & Anchored & Transfer & Transfer (cal) \\
\midrule
BeaverTails & $99.7\pm0.1$ & $73.1\pm2.5$ & $85.2\pm4.2$ & $94.1\pm0.3$ & $79.8\pm1.5$ & $84.3\pm1.8$ \\
Toxic-chat & $99.2\pm0.1$ & $89.1\pm1.3$ & $96.4\pm1.1$ & $89.1\pm0.2$ & $79.7\pm1.6$ & $83.9\pm1.5$ \\
\bottomrule
\end{tabular}
\end{adjustbox}
\end{table}

%% file: content/_tables/single_tables/guardqa/mlp/validation_depth/source_depth_selection_strategy_table.tex
\begin{table}[htbp]
\centering
\caption{GuardQA local-depth selection strategies. Values are AUROC in percentage points. Validation denotes validation-selected local depth; Optimal is post-hoc over the local-depth grid; Global uses the global anchored representation.}
\label{tab:guardqa-source-depth-selection-strategy}
\begin{adjustbox}{max width=\textwidth}
\begin{tabular}{llcccc}
\toprule
Dataset & Setting & $0.8$ & Validation & Optimal & Global \\
\midrule
\multicolumn{6}{l}{\textit{Anchored}} \\
BeaverTails & Output, h64 & $99.7\pm0.1$ & $99.8\pm0.1$ & $99.8\pm0.1$ @ $0.85$ & $99.8\pm0.1$ \\
BeaverTails & Output, h256 & $99.7\pm0.0$ & $99.7\pm0.1$ & $99.8\pm0.1$ @ $0.95$ & $99.7\pm0.1$ \\
Toxic-chat & Output, h64 & $99.2\pm0.1$ & $99.2\pm0.1$ & $99.2\pm0.1$ @ $0.95$ & $99.2\pm0.1$ \\
Toxic-chat & Output, h256 & $99.2\pm0.1$ & $99.2\pm0.1$ & $99.3\pm0.1$ @ $0.95$ & $99.2\pm0.1$ \\
\midrule
\multicolumn{6}{l}{\textit{Transfer}} \\
BeaverTails & Output, h64 & $78.7\pm0.6$ & $71.8\pm3.7$ & $79.2\pm0.7$ @ $0.85$ & $72.3\pm4.6$ \\
BeaverTails & Output, h256 & $74.3\pm0.5$ & $68.5\pm4.2$ & $78.5\pm0.3$ @ $0.85$ & $73.1\pm2.5$ \\
Toxic-chat & Output, h64 & $89.8\pm3.0$ & $88.9\pm2.1$ & $93.3\pm0.2$ @ $0.85$ & $87.4\pm2.7$ \\
Toxic-chat & Output, h256 & $91.2\pm1.6$ & $86.1\pm3.8$ & $92.8\pm0.7$ @ $0.85$ & $89.1\pm1.3$ \\
\midrule
\multicolumn{6}{l}{\textit{Transfer (cal)}} \\
BeaverTails & Output, h64 & $84.3\pm1.4$ & $81.2\pm2.6$ & $86.1\pm2.7$ @ $0.75$ & $89.0\pm1.1$ \\
BeaverTails & Output, h256 & $77.5\pm2.6$ & $83.3\pm2.0$ & $89.5\pm1.7$ @ $0.85$ & $85.2\pm4.2$ \\
Toxic-chat & Output, h64 & $97.4\pm0.6$ & $93.4\pm3.2$ & $97.4\pm0.6$ @ $0.80$ & $95.8\pm1.0$ \\
Toxic-chat & Output, h256 & $97.3\pm0.4$ & $96.3\pm1.2$ & $97.4\pm0.6$ @ $0.85$ & $96.4\pm1.1$ \\
\bottomrule
\end{tabular}
\end{adjustbox}
\end{table}

%% file: content/_tables/single_tables/guardqa/mlp/anchor_sensitivity/anchor_sensitivity_auroc.tex
\begin{table}[htbp]
\centering
\caption{GuardQA MLP AUROC cross-model transfer (Transfer (cal)) as anchor set size varies from 300 to 600. All probes use global anchored representations.}
\label{tab:final_guardqa_mlp_anchor_sensitivity_auroc}
\begin{adjustbox}{max width=\textwidth}
\begin{tabular}{lllccc}
\toprule
Dataset & Task & MLP & 300 anchors & 500 anchors & 600 anchors \\
\midrule
BeaverTails & Output Pred. & 64 & $89.0\pm1.1$ & $86.6\pm2.0$ & $73.4\pm5.4$ \\
BeaverTails & Output Pred. & 256 & $85.2\pm4.2$ & $76.0\pm5.3$ & $80.7\pm1.7$ \\
BeaverTails & GT Pred. & 64 & $84.0\pm2.7$ & $85.8\pm2.2$ & $88.1\pm0.7$ \\
BeaverTails & GT Pred. & 256 & $84.3\pm1.8$ & $84.0\pm2.1$ & $87.0\pm2.4$ \\
\midrule
Toxic-chat & Output Pred. & 64 & $95.8\pm1.0$ & $96.0\pm0.6$ & $98.0\pm0.5$ \\
Toxic-chat & Output Pred. & 256 & $96.4\pm1.1$ & $97.2\pm0.6$ & $97.7\pm0.6$ \\
Toxic-chat & GT Pred. & 64 & $82.2\pm1.8$ & $80.1\pm1.2$ & $82.3\pm1.3$ \\
Toxic-chat & GT Pred. & 256 & $83.9\pm1.5$ & $81.8\pm1.6$ & $83.5\pm0.5$ \\
\bottomrule
\end{tabular}
\end{adjustbox}
\end{table}

%% file: content/_tables/single_tables/guardqa/linear/non-abs/ls0.8/cross_summary_auroc.tex
\begin{table}[htbp]
\centering
\caption{GuardQA linear probe AUROC results for Output Prediction and GT Prediction: same-model Anchored, mean cross-model Transfer, and source-calibrated Transfer (cal).}
\label{tab:final_guardqa_linear_cross_summary_linear_concat_auroc}
\begin{adjustbox}{max width=\textwidth}
\begin{tabular}{lcccccc}
\toprule
 & \multicolumn{3}{c}{Output Prediction} & \multicolumn{3}{c}{GT Prediction} \\
\cmidrule(lr){2-4} \cmidrule(lr){5-7}
Dataset & Anchored & Transfer & Transfer (cal) & Anchored & Transfer & Transfer (cal) \\
\midrule
BeaverTails & $99.7\pm0.1$ & $78.6\pm3.4$ & $90.3\pm4.1$ & $95.0\pm0.6$ & $72.9\pm0.4$ & $80.9\pm1.4$ \\
Toxic-chat & $99.2\pm0.1$ & $91.0\pm0.6$ & $97.7\pm0.4$ & $89.9\pm0.2$ & $73.3\pm2.1$ & $80.3\pm2.4$ \\
\bottomrule
\end{tabular}
\end{adjustbox}
\end{table}

%% file: content/_tables/single_tables/guardqa/linear/non-abs/ls0.8/cross_summary_recall.tex
\begin{table}[htbp]
\centering
\caption{GuardQA linear probe R@0.1FPR results for Output Prediction and GT Prediction: same-model Anchored, mean cross-model Transfer, and source-calibrated Transfer (cal).}
\label{tab:final_guardqa_linear_cross_summary_linear_concat_recall}
\begin{adjustbox}{max width=\textwidth}
\begin{tabular}{lcccccc}
\toprule
 & \multicolumn{3}{c}{Output Prediction} & \multicolumn{3}{c}{GT Prediction} \\
\cmidrule(lr){2-4} \cmidrule(lr){5-7}
Dataset & Anchored & Transfer & Transfer (cal) & Anchored & Transfer & Transfer (cal) \\
\midrule
BeaverTails & $99.2\pm0.3$ & $69.9\pm4.8$ & $82.5\pm4.9$ & $84.0\pm2.0$ & $41.3\pm1.5$ & $52.2\pm4.2$ \\
Toxic-chat & $97.4\pm0.3$ & $81.4\pm0.5$ & $91.5\pm2.3$ & $70.5\pm2.8$ & $41.5\pm1.9$ & $51.0\pm1.7$ \\
\bottomrule
\end{tabular}
\end{adjustbox}
\end{table}

%% file: content/_tables/single_tables/guardqa/linear/standard_cross_model/standard_cross_model_auroc.tex
\begin{table}[htbp]
\centering
\caption{Standard cross-model linear-probe AUROC for GuardQA between Llama-3.1-8B-Instruct and Mistral-7B-Instruct-v0.3. Both models have 32 layers and 4096-dimensional residual streams; probes use layer 25 ($0.8$ relative depth). Values are means $\pm$ standard errors over three seeds.}
\label{tab:guardqa-linear-standard-cross-model-auroc}
\begin{adjustbox}{max width=\textwidth}
\begin{tabular}{lcccc}
\toprule
 & \multicolumn{2}{c}{Output Prediction} & \multicolumn{2}{c}{GT Prediction} \\
\cmidrule(lr){2-3} \cmidrule(lr){4-5}
Dataset & Llama$\to$Mistral & Mistral$\to$Llama & Llama$\to$Mistral & Mistral$\to$Llama \\
\midrule
BeaverTails QA & $73.7\pm13.0$ & $72.2\pm2.6$ & $47.7\pm8.6$ & $54.7\pm10.5$ \\
Toxic-chat QA & $79.0\pm13.9$ & $69.0\pm3.0$ & $54.2\pm11.3$ & $60.6\pm2.9$ \\
\bottomrule
\end{tabular}
\end{adjustbox}
\end{table}

%% file: content/_tables/single_tables/guardqa/linear/standard_cross_model/anchored_cross_model_auroc.tex
\begin{table}[htbp]
\centering
\caption{Global anchored linear-probe AUROC for GuardQA between Llama-3.1-8B-Instruct and Mistral-7B-Instruct-v0.3 ($300$ anchors). Values are means $\pm$ standard errors over three seeds.}
\label{tab:guardqa-linear-anchored-cross-model-auroc}
\begin{adjustbox}{max width=\textwidth}
\begin{tabular}{lcccc}
\toprule
 & \multicolumn{2}{c}{Output Prediction} & \multicolumn{2}{c}{GT Prediction} \\
\cmidrule(lr){2-3} \cmidrule(lr){4-5}
Dataset & Llama$\to$Mistral & Mistral$\to$Llama & Llama$\to$Mistral & Mistral$\to$Llama \\
\midrule
BeaverTails QA & $81.8\pm17.0$ & $96.1\pm3.1$ & $68.2\pm2.9$ & $78.3\pm4.3$ \\
Toxic-chat QA & $98.6\pm1.4$ & $97.7\pm0.4$ & $81.0\pm2.1$ & $72.4\pm4.4$ \\
\bottomrule
\end{tabular}
\end{adjustbox}
\end{table}